%% file: arvix-main.tex
\documentclass[letterpaper,twocolumn,10pt]{article}
\usepackage{usenix2019_v3}

\usepackage{tikz}
\usepackage{filecontents}
\usepackage{cite}
\usepackage{amsmath,amssymb,amsfonts}
\usepackage{mathrsfs}
\usepackage{graphicx}
\usepackage{enumitem}
\usepackage{xcolor}
\def\BibTeX{{\rm B\kern-.05em{\sc i\kern-.025em b}\kern-.08em
    T\kern-.1667em\lower.7ex\hbox{E}\kern-.125emX}}
\usepackage{hyperref}
\usepackage{diagbox}
\usepackage{algorithm}
\usepackage{algorithmicx}
\usepackage{algpseudocode}
\usepackage{comment}
\usepackage{xspace, amsthm, bbm, bm,theoremref}
\usepackage{subfigure}
\usepackage{multirow}

\newtheorem{theorem}{Theorem}

\usepackage{filecontents}
\usepackage[title]{appendix}
\usepackage{multicol} 
\usepackage{tabularx} 
\begin{document}

\date{}

\title{Lurking in the shadows: Unveiling Stealthy Backdoor Attacks \\ against Personalized Federated Learning}

\author{
{\rm Xiaoting Lyu}\\
Beijing Jiaotong University
\and
{\rm Yufei Han}\\
INRIA
\and
{\rm Wei Wang}\thanks{Corresponding author: Wei Wang and Guangquan Xu.}\\
Beijing Jiaotong University
\and
{\rm Jingkai Liu}\\
Beijing Jiaotong University
\and
{\rm Yongsheng Zhu}\\
Beijing Jiaotong University
\and
{\rm Guangquan Xu\textcolor{green!80!black}{\textsuperscript{*}}}\\
Tianjin University
\and
{\rm Jiqiang Liu}\\
Beijing Jiaotong University
\and
{\rm Xiangliang Zhang}\\
University of Notre Dame
}

\maketitle

\begin{abstract}
Federated Learning (FL) is a collaborative machine learning technique where multiple clients work together with a central server to train a global model without sharing their private data. However, the distribution shift across non-IID datasets of clients poses a challenge to this one-model-fits-all method hindering the ability of the global model to effectively adapt to each client's unique local data. To echo this challenge, personalized FL (PFL) is designed to allow each client to create personalized local models tailored to their private data.

While extensive research has scrutinized backdoor risks in FL, it has remained underexplored in PFL applications. In this study, we delve deep into the vulnerabilities of PFL to backdoor attacks. Our analysis showcases a tale of two cities. On the one hand, the personalization process in PFL can dilute the backdoor poisoning effects injected into the personalized local models. Furthermore, PFL systems can also deploy both server-end and client-end defense mechanisms to strengthen the barrier against backdoor attacks. On the other hand, our study shows that PFL fortified with these defense methods may offer a false sense of security. We propose \textit{PFedBA}, a stealthy and effective backdoor attack strategy applicable to PFL systems. \textit{PFedBA} ingeniously aligns the backdoor learning task with the main learning task of PFL by optimizing the trigger generation process. Our comprehensive experiments demonstrate the effectiveness of \textit{PFedBA} in seamlessly embedding triggers into personalized local models. \textit{PFedBA} yields outstanding attack performance across 10 state-of-the-art PFL algorithms, defeating the existing 6 defense mechanisms. Our study sheds light on the subtle yet potent backdoor threats to PFL systems, urging the community to bolster defenses against emerging backdoor challenges.

\end{abstract}

\section{Introduction}

Conventional federated learning (FL) techniques encounter the challenge of addressing data heterogeneity issues arising from non-IID data distributions among local devices in real-world scenarios\cite{towardsPFL}. For example, when using FL to develop a language model for mobile keyboards, clients from diverse demographic groups may demonstrate distinct usage patterns due to cultural, linguistic, and generational disparities. To address this challenge, Personalized Federated Learning (PFL) has emerged as a recent technological advancement. Unlike conventional FL, which relies on a single global model, PFL introduces personalized local models that effectively adapt to the unique data distribution of each client, thereby enhancing classification accuracy for individual clients.

Various PFL strategies have been explored \cite{kddpflbackdoor}, which can be categorized into full model-sharing methods\cite{localfinetuning-1,localfinetuning-2,per-fedavg,pfedme,ditto,fedala,scaffold} and partial model-sharing methods\cite{fedbn,fedrep}. Full model-sharing methods involve globally aggregated models that clients adapt through techniques such as fine-tuning with local data \cite{localfinetuning-1,localfinetuning-2}, meta-learning \cite{pfedme,per-fedavg}, parallel global and local model training \cite{ditto}, global model regularization during local training \cite{scaffold,fedprox}, or the weighted average of global and local models for local training initialization \cite{fedala}. In partial model-sharing methods, only part of the classifier model is shared among different clients. \cite{fedrep} shares the feature encoder while keeping the classification head private for each client. Similarly, \cite{fedbn} globally aggregates model parameters except for batch normalization layers for each client. 
As reported in these works, PFL outperforms traditional FL by creating personalized local models with significantly higher classification accuracy on each client's local data in non-IID scenarios.

{Customizing personalized local models for clients is both a blessing and a curse. While PFL methods enhance classification accuracy for clients, they concurrently open the door to backdoor attacks. Adversaries can seamlessly transfer backdoor triggers to personalized local models by optimizing the backdoor poisoning effect in the globally shared model. Previous works \cite{DBA,howtobackdoor,tails,alittle,durable} highlight traditional FL's vulnerability to backdoor attacks orchestrated by adversary-controlled clients. These malicious clients submit carefully crafted local models to establish trigger-target mappings in the global model, thereby manipulating it to output the adversary's desired label for backdoor data.
However, compared to traditional FL, PFL introduces additional barriers against backdoor attacks due to the personalization process adopted by benign clients, which fine-tunes the global model to better fit individual client data. Even if backdoors are successfully injected into the global model, personalization techniques can mitigate the backdoor poisoning effects in the personalized local models using backdoor-free data. 
For instance, local fine-tuning \cite{localfinetuning-1,localfinetuning-2} is one such personalization technique that utilizes the client's backdoor-free data to tune the global model. Fine-tuning global models with backdoor-free data can disrupt the presence of the backdoor trigger in personalized local models, leading to catastrophic forgetting of the trigger \cite{french1999catastrophic}, thereby mitigating backdoor poisoning effects on these models.
Therefore, executing successful backdoor attacks in PFL requires not only maintaining stealthiness to evade defenses deployed at the central server or local clients but also ensuring the persistence of backdoor poisoning effects throughout the personalization process. However, the specific threat posed by backdoor attacks within the personalization context of PFL remains relatively underexplored.}

{Our study aims to fill this gap by exploring the potential threat of backdoor attacks in PFL systems and demonstrating their feasibility to mainstream PFL techniques, covering both state-of-the-art full and partial model sharing PFL methods \cite{localfinetuning-1,localfinetuning-2,per-fedavg,pfedme,ditto,fedala,fedprox,scaffold,fedbn,fedrep}. Traditional backdoor attacks on FL encounter three primary bottlenecks when targeting PFL.}
\textbf{First}, in full model-sharing PFL methods, fine-tuning the global model with clean local training data can weaken the backdoor poisoning effect. This phenomenon is akin to catastrophic forgetting\cite{french1999catastrophic}, where the local model forgets the trigger-target mapping, thereby reducing the impact of backdoor poisoning effects.
\textbf{Second}, in partial model-sharing PFL methods, only the globally shared layers of a neural network are aggregated at the server, while each client customizes the remaining layers using their clean local training data. Adversaries can only inject backdoor poisoning effects into the globally shared layers. It remains uncertain whether these backdoor poisoning effects can persist in the personalized local model, which consists of the poisoned globally shared layers and privately owned local layers. 
Empirical observations in Section \ref{sec:5.2} confirm PFL systems' resilience to backdoor attacks on clients' personalized local models. For instance, Table \ref{tab:nodefense} shows that the existing backdoor attacks cannot succeed in inserting the backdoor trigger into personalized local models, whether from partial or full model-sharing PFL algorithms on the Fashion-MNIST dataset. 

\textbf{Furthermore}, PFL systems can deploy both server-end and client-end countermeasures against backdoor attacks to enhance their resilience to model integrity threats. The central server can leverage robust aggregation algorithms \cite{mult-krum,trimmed_mean,ndss2021,flame} to counteract the poisoning effects introduced by malicious clients. Simultaneously, local clients can utilize backdoor mitigation methods \cite{NAD,wang2019neural} on their local devices to remove potential backdoor triggers.

Despite these challenges, our research reveals that the flourishing of PFL approaches fortified with various defense mechanisms is a false sense of comfort in the face of the threat of backdoor poisoning attacks. To substantiate this backdoor threat, we introduce a novel backdoor attack method, \textit{PFedBA}, and evaluate its effectiveness against mainstream PFL approaches in different scenarios.  Experimental results show that \textit{PFedBA} can perform successful attacks against mainstream PFL approaches, circumventing state-of-the-art server-end and client-end defense methods.

Our design of \textit{PFedBA} is inspired by the recent advancements in understanding the training dynamics of deep neural networks.
First of all, \cite{Domingos2020EveryML} has shown that deep neural networks trained using gradient descent can be mathematically equivalent to a Gaussian process equipped with neural tangent kernels (NTKs). According to the theory of NTK, the similarity between the decision boundaries of two neural network models can be quantified by measuring the degree of alignment between the gradient vectors of the two models. 
Besides, \cite{ma2018teacher,ainsworth2023git} reveal that the loss landscape of a neural network training process nearly contains one single basin at convergence. By strategically manipulating the training data, It is possible to guide two neural network models to converge towards the same loss basin, ultimately leading to similar classification boundaries, a.k.a Machine Teaching \cite{ma2018teacher}. 
With these concepts in mind, our intuition is: if we align the gradient of the backdoor task with that of the main learning task, and align the classification loss of the two tasks into the same loss basin, we will obtain a backdoor-poisoned model whose classification boundary closely resembles that of the backdoor-free model trained with clean data. Maintaining consistency in classification boundaries enables the backdoor poisoning effects to persist during the personalization process and evade the defense mechanisms deployed at both local clients and the central server.

More specifically, we formulate the backdoor attack method against PFL as a joint optimization problem, where the backdoor trigger and poisoned model parameters are simultaneously optimized during the global training phase. Our goal is twofold: minimizing the classification loss on the backdoor input by alternately optimizing the trigger and model parameters while minimizing the Euclidean distance between the gradient of the backdoor task and the gradient of the main task in each attack iteration. In this way, we align the gradients and losses of the main and backdoor tasks, making the decision boundary of the backdoor poisoned model very similar to that of the clean model. In Section \ref{sec4}, we elaborate on how \textit{PFedBA} persists within PFL algorithms, and this persistence  holds true even in the presence of defense mechanisms.

We conduct a comprehensive empirical study to demonstrate the effectiveness of our proposed \textit{PFedBA} attack against PFL systems on 4 large-scale datasets: Fashion-MNIST, CIFAR-10, CIFAR-100, and N-BaIoT. \textit{PFedBA} successfully injects backdoor triggers into personalized local models generated with 10 popular PFL algorithms, including 8 full model-sharing methods \cite{localfinetuning-1,localfinetuning-2,pfedme,per-fedavg,ditto,fedala,scaffold} and 2 partial model-sharing methods\cite{fedbn,fedrep}. In comparison to \textit{PFedBA}, we evaluate 5 prevalent backdoor attack methods—\textit{Sybil attack}\cite{foolsgold}, \textit{Model replacement attack}\cite{howtobackdoor}, \textit{PGD attack}\cite{tails},  \textit{Neurotoxin}\cite{durable}, \textit{CerP}\cite{LyuHWLWL023}—as attack baselines. To gauge attack effectiveness against countermeasures, we consider 6 state-of-the-art defense methods, including 4 server-end defenses\cite{mult-krum,trimmed_mean,flame,ndss2021} in FL and 2 client-end backdoor mitigation techniques \cite{NAD,wang2019neural}. 
Our experimental results reveal that \textit{PFedBA} achieves high attack success rates on personalized local models while maintaining accuracy in the main task. Even in scenarios where robust FL methods and backdoor mitigation techniques are implemented, \textit{PFedBA} outperforms the attack baselines. For instance, on Fashion-MNIST, the attack success rate of the baselines drops to at least 70\% and 58\% of \textit{PFedBA} without and with defense methods, respectively. These findings underscore the attack effectiveness of \textit{PFedBA} based on gradient alignment and loss alignment.

The key contributions in our study can be summarized in the following perspectives:
\begin{itemize}
\item Our research exposes the backdoor poisoning vulnerabilities in PFL systems by introducing an effective distributed backdoor attack targeting PFL systems. We highlight a critical security concern regarding model integrity and reliability in practical PFL applications. This issue, while rarely explored in mainstream PFL solutions, demands attention from the security community. 


\item We formulate the \textit{PFedBA}-driven attack as a joint optimization problem to simultaneously generate the backdoor trigger and optimize the poisoned models to align the main and backdoor tasks. The goal is to maintain the backdoor poisoning effect during the personalization stage while maximizing the attack performance against the server-end and client-end defense mechanisms. 

\item Our study organizes a large-scale and comprehensive evaluation of the backdoor threat in PFL systems on 4 benchmark datasets, utilizing 10 mainstream PFL methods, 6 backdoor attack methods (including \textit{PFedBA}), and 6 defense mechanisms. Our results provide a convincing argument for the attack effectiveness of \textit{PFedBA}. 

  
\end{itemize}

\section{Related Work}

\subsection{Personalized Federated Learning}
Our work focuses on the state-of-the-art PFL methods, which includes full model-sharing \cite{localfinetuning-1,localfinetuning-2,ditto,fedala,per-fedavg,pfedme,scaffold} and partial model-sharing\cite{fedbn,fedrep} methods.

\noindent
\textbf{In full model-sharing PFL methods}, local clients jointly optimize the global model and use it as a reference to adapt their personalized local models.
\textit{Local fine-tuning} \cite{localfinetuning-1,localfinetuning-2} is the most straightforward PFL technique of this kind.  It first updates the global model by aggregating clients' local model updates globally.  After that, each client uses the global model as an initialization for local training and fine-tunes their local model with their own data.
Beyond the simple fine-tuning strategy, \textit{SCAFFOLD}\cite{scaffold} regularizes the difference between the global model update and individual local model updates, which aims to reach a balance of generalization and diversity of the locally adapted models. 
Furthermore, PFL methods based on meta-learning \cite{per-fedavg,pfedme} train a meta global model to facilitate few-shot fine-tuning of each local model. \textit{Per-FedAvg} \cite{per-fedavg} is a direct extension of the vanilla \textit{FedAvg}, incorporating Model-Agnostic Meta-Learning (MAML)\cite{finn2017model} into the global model training stage. 
\textit{pFedMe} \cite{pfedme} further extends the learning framework by integrating Moreau envelopes and the L2-norm regularization to tune the trade-off between local model adaptation to the local training data and generalization performance on data beyond the local training set. 

In parallel, \textit{Ditto} \cite{ditto} introduces a multi-task learning framework for personalized local models, considering two tasks: global model learning and local model learning. To relate these tasks, \textit{Ditto} incorporates a regularization term that encourages the personalized local models to stay close to the optimal global model.
Other PFL methods \cite{fedala,apple} propose to adapt the local models by customizing the initialized model parameters for local training. For example, instead of using the shared global model directly for initializing local models, \textit{FedALA}\cite{fedala} proposes to learn a client-specific weighted average of the global and local model derived from the past global training iteration as each client's initialized local model. 

\noindent \textbf{In partial model-sharing PFL methods}, each personalized local model is composed of the globally shared layers and privately owned local layers. Though both sets of layers are updated via local training at local clients, only the globally shared layers are submitted to the central server for model aggregation. Different clients utilize the globally shared layers while adapting the privately owned local layers to their data.
In \textit{FedBN}\cite{fedbn}, clients keep the Batch Normalization layers of the learned model as the privately owned local layers, while the remaining layers of the model are globally updated following the standard \textit{FedAvg} protocol. Similarly, \textit{FedRep}\cite{fedrep} divides the model into a globally shared feature encoder and a local classification head. Clients jointly learn the feature encoder and share the same feature embedding space, while tuning their classification heads to better fit their local data.

\subsection{Distributed Backdoor Attacks to FL} \label{sec:2.3}
Several  studies\cite{DBA,howtobackdoor,tails,foolsgold,alittle,backdoormeta,durable} have highlighted the vulnerability of FL to distributed backdoor attacks launched by malicious clients in FL systems. 
Pioneering efforts in this domain, such as the \textit{Sybil attack} \cite{foolsgold}, \textit{Model Replacement attack} \cite{howtobackdoor}, and \textit{DBA} \cite{DBA}, organize distributed backdoor attacks by having malicious clients independently train and submit the poisoned local model. 
\textit{Sybil attack} \cite{foolsgold} involves malicious clients submitting backdoor-poisoned local models directly targeting FedAvg-based aggregation rules that aggregate model parameters from both benign and malicious clients.
\textit{Model Replacement attack} \cite{howtobackdoor} strengthens significantly the attack strength by scaling up the poisoned model updates submitted by malicious clients, effectively achieving its attack objective in very few global aggregation iterations. 
Since the model parameters submitted by malicious clients are distinguished from those submitted by benign clients, both attacks can be easily mitigated by Byzantine-robust FL methods.
Besides, \textit{DBA}\cite{DBA} divides a global trigger into multiple local triggers and distributes them to all malicious clients. However, it is challenging to embed the complete trigger into the global model since partial-participation FL cannot ensure the selection of all malicious clients for model aggregation. 

Other distributed backdoor attack methods focus on improving the attack performance against robust FL aggregation methods. 
\cite{alittle} attempts to evade the robust FL aggregation rules by constraining the parameters of poisoned local models within the parameter value range observed in benign local models. 
\textit{PGD attack}\cite{tails} utilizes the projected gradient descent to update the poisoned local models, aiming to keep their local models relatively close to the global model. 
Similarly, \textit{Neurotoxin} \cite{durable}, to increase the durability of the poisoning effect, adopts a coordinate-wise project gradient descent approach, focusing on the parameters that are rarely updated during benign local training, to learn the backdoor model.
\textit{CerP} \cite{LyuHWLWL023} slightly fine-tunes the backdoor trigger with an L2-ball constraint and explicitly controls malicious model bias to improve the attack effectiveness and stealthiness.

Effective as they are, these distributed backdoor attack methods are designed to poison only the globally shared model in FL systems. It remains an open question whether embedded backdoor effects persist in personalized local models. Nonetheless, from these studies, we can gain insights from two perspectives. First, optimizing the learning process of backdoor poisoned local models by constraining or regularizing gradient updates can narrow the gap between the backdoor and clean models, aiding in evading the robust FL methods without compromising backdoor attack performance.
Second, as suggested by \cite{LyuHWLWL023}, it is feasible for the adversary to adapt the trigger pattern to improve attack performance, yet guarantee the high utility of the backdoor model.

Our study is guided by these insights learned from previous works. We increase the flexibility of our attack by generating carefully crafted triggers to force gradient alignment and loss alignment of the main and backdoor tasks. This design facilitates the seamless transfer of backdoor effects from the globally shared model to locally personalized models while keeping the attack undetected facing robust FL methods. 

{It is worth noting that our work focuses on exploring the vulnerabilities to backdoor attacks in PFL systems that incorporate both global training and local personalization, encompassing both partial and full model-sharing PFL methods. In contrast, \cite{DBLP:journals/corr/abs-2307-15971} focuses solely on attacks targeting partial model-sharing PFL systems, where only the classification head is fine-tuned locally, while the feature encoder is globally shared among all clients. 
Furthermore, \cite{DBLP:journals/corr/abs-2201-07063} focuses on the HyperNetFL system, which employs a neural network hosted by the server to generate local model parameters for each client using the client-specific descriptor and parameters of a special shared model as inputs. Backdooring this HyperNetFL model resembles backdooring traditional FL systems, where a specially crafted model embedded with backdoors is shared among all clients. This differs from the PFL system examined in our study, which incorporates a local personalization process.}

\subsection{Backdoor Defense Methods}
\noindent\textbf{Server-end Backdoor Defenses.} State-of-the-art robust FL methods leverage a variety of anomaly detection techniques to filter out abnormal local model updates or introduce random noise to the model, thus mitigating the impact of backdoor attacks. \textit{Trimmed mean} \cite{trimmed_mean}, a coordinate-wise aggregation rule, sorts $n$ local model updates for parameters in each dimension and excludes the largest and smallest $\beta$ values. The remaining $n-2\beta$ values are then averaged as the corresponding coordinate of the global model update. \textit{Multi-Krum} \cite{mult-krum} selects multiple local model updates with the closest Euclidean distance to the nearest $n-m-2$ local model updates (where $m$ is the number of malicious clients in FL) and computes their average as the global model update. \textit{DnC} \cite{ndss2021} first computes the principal component of local model updates and then calculates the dot product of each local model update with the principal component. It removes the local model updates that deviate the most from the principal component. \textit{FLAME} \cite{flame} utilizes the model clustering method HDBSCAN \cite{HDBSCAN} to identify potentially poisoned local models. It then clips the parameter values of the remaining local models and introduces the random Gaussian noise to the aggregated global model. 
{\textit{FLTrust} \cite{DBLP:conf/ndss/CaoF0G21} allows the central server to maintain a small, poison-free dataset sampled from a distribution similar to the entire training dataset. The server utilizes this dataset to train a global reference model and computes a linear combination of local models weighted by the score, measured by the cosine similarity between the global reference model and local models.
\textit{FedRecover} \cite{DBLP:conf/sp/CaoJZG23} is a model recovery method capable of leveraging historical information to restore clean global models from poisoning attacks.
\textit{FLCert} \cite{DBLP:journals/tifs/CaoZJG22} constructs an ensemble of global models by randomly selecting $k$ out of $n$ clients to produce $N$ global models. The label for a testing input is predicted based on the majority vote among these derived global models.}

\noindent\textbf{Client-end Backdoor Defenses.} Current strategies addressing backdoor attacks in centralized machine learning can be classified into three main categories: pre-training defense, in-training defense, and post-training defense, as outlined in \cite{wu2022backdoorbench}. Pre-training defense methods, like \textit{Februus} \cite{doan2020februus}, focus on identifying and removing poisoned data samples before the model training phase. However, these methods are not practical in FL systems due to privacy-preserving protocols. In-training defense mechanisms, exemplified by \textit{anti-backdoor learning techniques}\cite{li2021anti}, aim to mitigate the impact of backdoor poisoning during the model training process. Nevertheless, this approach faces challenges in FL because the distributed nature of FL restricts the defender from overseeing the local training processes of individual clients. Post-training defense methods concentrate on eliminating or mitigating the backdoor poisoning effects in already-trained models. For instance, \textit{Neural Cleanse} (NC) \cite{wang2019neural} functions by reversing the trigger on clean data instances and subsequently removing the backdoor through unlearning-based deep neural network patching or neuron pruning. Another approach, \textit{Neural Attention Distillation} (NAD) \cite{NAD}, achieves backdoor mitigation by pruning neurons highly correlated with the backdoor trigger.

Our study explores whether these state-of-the-art server-end and client-end defense methods can enhance the resilience of the PFL system against the proposed backdoor attack when adopted by both the server and benign clients.
Our key insight lies in the alignment of losses and gradients between backdoor and main tasks, offering a two-fold advantage in evading these defense methods. Firstly, it reduces model bias between backdoor-poisoned and clean models, facilitating bypass the server-end defenses. Secondly, aligning the main task with the backdoor task narrows the gap between the decision boundaries of the backdoor model and the clean model, enabling the circumvention of client-end defenses.

\section{Problem Definition and Threat Model}
\subsection{Problem Definition}
\label{sec:3.1}

Suppose we have $N$ clients in the PFL system. Each client possesses a local dataset ${D_i} = \left\{ {\{ {x_{i,j}}\in{R^{m}},{y_{i,j}}\} _{j = 1}^{{d_i}}} \right\}$, where ${d_i} = \left| {{D_i}} \right|$ and ${\{ {x_{i,j}},{y_{i,j}}\} }$ represents the feature vector and class label of each data instance hosted by the client $i$. 
We highlight that the local datasets of different clients are non-IID. 
The PFL system aims to collaboratively use the private datasets ${D_1}, \cdots ,{D_N}$ to find the optimal set of the personalized local models $[{{{\omega}}_1}, \cdots ,{{{\omega}}_N}]$ that better fit these local datasets. The PFL problem can be formulated as:
\begin{equation}
\small
\mathop {\min }\limits_{{W=[{{{\omega}} _1},{{{\omega}} _2}, \cdots ,{{{\omega}} _N}]}} {F_P}({W}) = \mathop {\min }\limits_{{{{\omega}} _i},i \in [N]} \sum\limits_{i = 1}^N {{p_i}{f_i}({{{\omega}} _i})} 
\end{equation}
where ${F_P}( \cdot )$ is the global objective of PFL, and $p_i \ge 0$ is the aggregated weight of client $i$. 

In this study, we focus on both full and partial model-sharing based PFL methods. All of these PFL methods involve two stages: a global aggregation stage and a local model adaptation stage, which runs sequentially or jointly. More specifically, the global aggregation stage follows the standard FL protocol. During each global aggregation iteration, the server aggregates local model updates from the selected clients to update the globally shared model, which is then shared back with all clients. While or after the global aggregation stage, local clients utilize the globally shared model to fine-tune their personalized local models ${{\omega}}_i$, thereby capturing the data distribution of each client.

\subsection{Threat Model}
\label{sec:threat model}
\noindent \textbf{Attack Goal.}
The objective of backdoor attacks in PFL differs from those in traditional FL\cite{alittle,DBA,howtobackdoor,tails,durable}. While backdoor attacks against traditional FL aim to induce misclassification in the global model for trigger-embedded inputs, adversaries targeting PFL aim to introduce backdoor poisoning effects into personalized local models. The poisoned personalized local models should produce the adversary-desired output for trigger-embedded inputs while functioning normally for inputs without triggers. Our focus further extends to stealthy backdoor attacks, designed to bypass both server-end and client-end defenses while ensuring the success of the attacks.

\noindent \textbf{Attack Capability.}
We assume the adversary has control $C$ out of $N$ clients ($C\ll N$) in the PFL system. {The adversary can access and arbitrarily manipulate the local data, labels, training status, and the local model training process hosted by malicious clients to launch dirty label attacks.} The adversary can leverage the local datasets from malicious clients to create a dataset denoted as $D^{mal}$ for generating the optimized trigger. It is important to note that the adversary cannot collect any auxiliary data for use in the attack or compromise the central server.
Furthermore, the adversary cannot access or tamper with the local datasets, the local model training process, and the personalization process of benign clients.

\noindent\textbf{Attack Knowledge.}
The adversary has complete access to the local datasets, local training, and personalization processes of malicious clients to conduct backdoor attacks. Nevertheless, the adversary remains entirely unaware of any information concerning the benign clients and the central server. Furthermore, the adversary is unaware of the defense techniques implemented on the server or clients in the PFL system. 

In this study, we aim to unveil the feasibility of backdoor attacks in practical PFL systems, where the adversary can only manipulate compromised local clients and remains agnostic to the learning strategy adopted by the central server and other benign clients. As such, our work is based on a logical premise within our threat model: if an adversary can successfully execute backdoor attacks with such limited knowledge and attack capabilities, the potential for even more severe threats to model integrity increases when the adversary possesses greater attack capabilities and richer knowledge of the target PFL system. Our findings provide insights into the vulnerability of real-world PFL applications to backdoor attacks.



\section{The PFedBA-based attack method}
\label{sec4}
\begin{figure}[t] 
\centering  
\includegraphics[height=4.8cm,width=8.3cm]{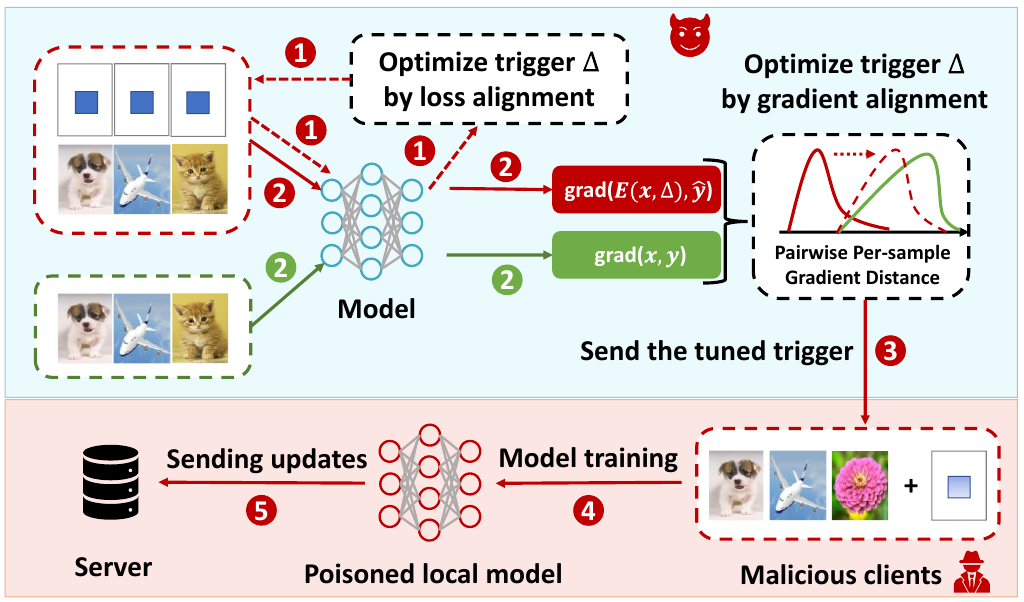}
\caption{Illustration of our \textit{PFedBA} attack.} 
\label{fig:overview}
\end{figure}
{The workflow and the pseudo codes of the proposed \textit{PFedBA} attack are given in Figure \ref{fig:overview} and Algorithm \ref{alg:attack}.
{In the $t$-th poisoned federated training iteration, }the pipeline of \textit{PFedBA} consists of two stages: alignment-optimizing driven trigger generation and backdoor poisoned local training. In the trigger generation stage, the adversary uses the training data $D^{mal}$ owned by the adversary-controlled malicious clients to generate the backdoor trigger $\Delta^{t}$ by enforcing the loss alignment and gradient alignment between the backdoor and the main task. After obtaining the optimized trigger, the adversary sends the trigger $\Delta^{t}$ to the malicious clients. In the backdoor poisoned local training stage, the malicious clients embed the optimized trigger $\Delta^{t}$ into their local training data and train the backdoor local models. Finally, the selected malicious clients send the poisoned model updates to the central server. Both the backdoor and clean local model updates are aggregated at the server to update the global model. }

\begin{algorithm}[htb]
\caption{ PFedBA}
\label{alg:attack}
\begin{algorithmic}[1] 
   \Require
      The number of the selected clients $N_p$,
      the first attack iteration $T_a$,
      {the poisoned federated training iteration $t$,}
      the initial trigger $\Delta^0$,
      the mask $m$,
      the dataset $D^{mal}$.
           
  \Ensure 
      The optimized backdoor trigger ${\Delta ^t}$, the backdoor local model update $\Delta{{\omega}}_i^t$.
      
  \For{each attack iteration $t$}
       \State Server sends ${{\omega}}^{t-1}$ to all clients and randomly select $N_p$ clients to participate in model aggregation.
       \If{ malicious clients are selected by the server}{}

           \Procedure{ \textbf{1: Trigger generation}} {}
           \State $\Delta^{t} \leftarrow \Delta^{t-1}$;
            \If{ $t==T_a$}{}
                \State The adversary uses $D^{mal}$ and ${{\omega}}^{t-1}$ to generate $\Delta^{t}$ by computing Eq.\ref{eq:trigger-tuning} with {$\lambda=0$};
            \EndIf

           \State The adversary generates $\Delta^{t}$ by computing Eq.\ref{eq:trigger-tuning} with {$\lambda=1$} based on $D^{mal}$ and ${{\omega}}^{t-1}$;
           \State The server send the optimized trigger $\Delta^{t}$ to malicious clients.
           \EndProcedure
           
           \Procedure{\textbf{2: backdoor Poisoned Local training}}{}
           \For {each selected malicious clients $i$}
           \State Patch $x, x \in D_{i}^{mal}$ with ${\Delta ^t}$ by Eq.\ref{eq:trigger_embed};
           \State Train the backdoor poisoned local model ${{\omega}} _i^t$ through Eq.\ref{eq:ourattack};
           \State Send the poisoned local model update $\Delta{{\omega}}_i^t={{\omega}} _i^t-{{\omega}}^{t-1}$ to the server.
           \EndFor
           \EndProcedure  
       \EndIf
  \EndFor
\end{algorithmic}
\end{algorithm}

\subsection{The Design of PFedBA}\label{sec:4.1}

Our \textit{PFedBA} attack is formulated as a distributed optimization problem with the objective function defined in Eq.\ref{eq:ourattack}.
In the $t$-th training iteration, the adversary launches the attack via jointly optimizing Eq.\ref{eq:ourattack} with the malicious clients selected to join the global model aggregation:
\begin{equation}
\small
\label{eq:ourattack}
\begin{split}
&{\omega _i^{*,t}} = \mathop {\arg \min }\limits_{{\omega _i^t}}   \sum\limits_{j \in D_i^{mal}} {{{\cal L} _{{\omega _i^t}}}} (E({x_{i,j}},{\Delta ^{*,t}}),\hat y) + \sum\limits_{j \in D_i^{nor}} {{{\cal L} _{{\omega _i^t}}}} ({x_{i,j}},{y_{i,j}}) \\
&{\rm{s}}{\rm{.t}}{\rm{. }} \,\,{\Delta ^{*,t}} = \mathop {\arg \min }\limits_{{\Delta ^{t}}} AL({\Delta ^{t}},{{\omega ^{t-1}}, D^{mal} })
\end{split}
\end{equation}
{where ${\cal L}_{{\omega _i^t}}(x,y)$ denotes the classification loss function given the data instance $(x,y)$ and the backdoored local model ${\omega_{i}^t}$ of the malicious client $i$. ${\omega ^{t - 1}}$ represents the aggregated model obtained from  the previous federated training iteration $t-1$, which the adversary uses to optimize the task-specific backdoor trigger. Specifically, in full model-sharing PFL methods, ${\omega ^{t - 1}}$ represents the global model aggregated at the previous FL training iteration $t-1$, which serves as the initialized local model for local clients at iteration $t$. In partial model-sharing PFL methods, ${\omega ^{t - 1}}$ refers to the initialized local model of the malicious client specified by the adversary at iteration $t$, comprising both the globally shared layers and the privately owned local layers. Notably, if the adversary chooses to inject backdoors from the $t$-th federated training iteration onwards, $\omega^{t-1}$ remains backdoor-free, while $\omega_i^{t}$ is injected with the backdoor trigger. As the attack progresses, the backdoor is embedded in the model $\omega^{t}$.}
$D_i^{{mal}}$ and $D_i^{{nor}}$ represent the backdoor and the clean training data hosted by the malicious client $i$, respectively. $\hat y$ indicates the adversary-desired classification output. 
$D^{mal} = \{D_i^{{mal}}\}$ denotes the union of all the backdoor local training datasets owned by malicious clients. \textit{In our study, we assume the adversary can manipulate all the training samples hosted by the malicious clients. However, she cannot collect any auxiliary data.} 
 
{The function $AL(\cdot)$ in Eq.\ref{eq:ourattack} of \textit{PFedBA}, as defined in Eq.\ref{eq:trigger-tuning}, represents the module responsible for generating the task-specific backdoor trigger $\Delta^t$ using the dataset $D^{mal}$ and the model ${\omega ^{t - 1}}$. The parameter $\lambda$, which can take values of either 1 or 0, determines the optimization method: when set to 1, optimization prioritizes gradient alignment; when set to 0, it emphasizes loss alignment. In \textit{PFedBA}, the trigger is optimized using gradient alignment and loss alignment by adjusting $\lambda$ accordingly. Specifically, in the first attack iteration, the adversary initially optimizes the backdoor trigger by minimizing Eq.3 with $\lambda=0$. Then, the adversary tunes the optimized trigger by minimizing Eq.3 with $\lambda=1$. For all subsequent attack iterations, the adversary solely minimizes Eq.3 with $\lambda=1$ to further optimize the trigger.}
\begin{equation}
\label{eq:trigger-tuning}
\small
 AL =  \left\{ \begin{array}{l}
{\lambda} \sum\limits_{l \in {D^{mal}}}  \left\| {{{\nabla _{{{\omega ^{t - 1}}}}}{\cal L} } (E({x_l},{\Delta ^t}),\hat y) - {\nabla _{{{\omega ^{t - 1}}}}}{\cal L} ({x_l},{y_l})} \right\|_2\\ \\
 + (1 - {\lambda} )\sum\limits_{l \in {D^{mal}}} {{{\cal L} _{{{\omega ^{t - 1}}}}}(E({x_l},{\Delta ^t}),\hat y)} 
\end{array} \right.
\end{equation}
$E$(·) defined in Eq.\ref{eq:trigger_embed} represents the function that embeds a trigger $\Delta$ into the original data sample $x$. $m$ is a matrix called the mask, deciding how much the trigger can overwrite the original data sample. Values in the mask are 0 or 1. $m_{i,j}=0$ means the feature is not modified at all, and $m_{i,j}=1$ means the trigger completely overwrites the feature ($x_{i,j}={\Delta}_{i,j}$). In our work, the mask $m$ designed by the adversary is fixed.  {The symbol $\odot$ represents element-wise multiplication.}
\begin{equation}
\small
\label{eq:trigger_embed}
    E(x,\Delta ) = x \odot (1 - m) + \Delta  \odot m
\end{equation}
Next, we describe the details of the proposed \textit{PFedBA} attack, which consists of two steps: {\textit{alignment-optimizing driven trigger generation}} and {\textit{backdoor poisoned local training}}.

\subsubsection{Alignment-Optimizing driven Trigger Generation}
The common way of backdoor attack is to train the model using the trigger-embedded data, in order to force the model to memorize the association between the trigger-embedded input with the adversary-desired classification output. However, the backdoor poisoning effect embedded in the globally shared model can be diluted during the personalization process. In this process, the globally shared model is further fine-tuned with clean training data hosted by local clients. Due to the catastrophic forgetting phenomenon unveiled in neural networks, the poisoning effect becomes significantly weaker with more rounds of local model tuning. Besides, injecting an arbitrary trigger may cause excessive malicious model bias {\cite{LyuHWLWL023}}. The backdoor poisoned local models may thus be easily detected by the server-end defense mechanisms during the global aggregation stage of PFL. At the personalization stage, benign clients can also deploy client-end backdoor mitigation methods to detect and mitigate the backdoor trigger in the local model. 
To address these bottlenecks, we consider both the trigger and the global model parameters as optimization variables in \textit{PFedBA}. We propose to generate task-specific triggers to align the backdoor task with the main task. The proposed trigger generation module, given in Eq.\ref{eq:trigger-tuning}, is designed to achieve \textit{gradient alignment} and \textit{loss alignment} simultaneously between the main learning task and the backdoor task. These two alignment objectives are discussed below.

\noindent
\textbf{\textit{Objective of Gradient Alignment.}} This alignment term (the first term of Eq.\ref{eq:trigger-tuning}) is designed to adapt the backdoor trigger $\Delta$ such that it can directly align the gradients of the backdoor and main tasks along the same direction, i.e. we minimize the Euclidean distance between the two gradient vectors. Both of these gradient vectors mentioned above are computed by the adversary using the model ${\omega}^{t-1}$, the poison-free data $D^{mal}$ and the corresponding backdoor training samples ${\hat D}^{mal}$.
Intuitively, minimizing the Euclidean distance serves to bring the gradient of the backdoor task closer to that of the main task. As benign clients adapt the model using clean training data through gradient descent, the memory of the mapping between the trigger-embedded input and the adversary-desired class label in the personalized local model is strengthened due to the consistent gradient direction of the main task and the backdoor task. Therefore, this gradient alignment term prevents the erasure of triggers during the personalization stage.

Furthermore, the idea of introducing gradient alignment into our attack is also inspired by theoretical breakthroughs in Neural tangent kernels (NTKs) \cite{Domingos2020EveryML}. Originally introduced to explain the evolution of deep neural networks using gradient descent during training, NTK theory allows us to measure the similarity between the decision boundaries of two neural network models based on the inner product between the gradients of the two neural network models for given inputs. The greater the inner product (reflecting closer alignment between the gradients), the more similar the decision boundaries of these two models are. Building upon this theory, we enforce alignment between the gradients of the backdoor task and the main task, with the goal of making the decision boundary of the backdoor model as similar as possible to the decision boundary of the backdoor-free model. The enhanced closeness between the decision boundaries of the backdoor model and the backdoor-free model helps the trigger  evade the client-end defense methods\cite{NAD,wang2019neural}.

The enhanced closeness between the decision boundaries of the backdoor model and the backdoor-free model also helps our attack evade the server-end defense methods\cite{mult-krum,flame,ndss2021,trimmed_mean} that are designed to detect Byzantine clients following the spirit of anomaly detection. Performing the gradient alignment reduces the model bias between the backdoor local models and the backdoor-free local models, which makes the server-end defense methods unable to differentiate the backdoor local models from the backdoor-free ones. 

\noindent
\textbf{\textit{Objective of Loss Alignment.}} {The loss alignment between the main and backdoor tasks is formulated as a bi-level optimization problem, described in Eq.2. In the outer-layer optimization, our aim is to minimize classification losses for both tasks by adjusting local model parameters while keeping the trigger fixed. In the inner-layer optimization (the second term of Eq.\ref{eq:trigger-tuning}), we freeze the model and optimize the trigger to minimize the classification loss of the backdoor task on the model ${\omega}^{t-1}$, while ensuring the classification loss of the main task remains unchanged.
Intuitively, if a model well-trained on backdoor-free data can achieve a low classification loss on backdoored data, the attack objective is achieved without significantly altering the model parameters, thereby mitigating malicious model bias caused by trigger injections. Therefore, the objective of loss alignment is to align the classification loss for the backdoor task with that of the main task through trigger optimization, bringing them into the same loss landscape on the model without changing the model parameters.}

The idea of trigger generation using the loss alignment can find its origin in the concept of Machine Teaching\cite{ma2018teacher}: given a testing set, fine-tuning the training data can calibrate the decision boundary of a classifier and help the classifier produce the desired classification outputs for the test data. We follow the spirit of Machine Teaching and minimize the backdoor classification loss, thus achieving loss alignment between the main task and the backdoor task on the model. This alignment helps minimize the divergence in decision boundaries between the backdoor and backdoor-free models. Therefore, the client-end backdoor mitigation methods \cite{wang2019neural} may be ineffective in removing adversary-specified backdoor triggers, since these methods cannot distinguish between the decision boundaries in backdoor and backdoor-free models. 

Additionally, according to \cite{ainsworth2023git}, the loss landscape of deep neural network training typically exhibits a few equivalent basins with similar model parameters at convergence. Aligning the loss landscape of the backdoor task with that of the main task forces the backdoor model to converge to the same basin area of the backdoor-free model, which can effectively reduce the model bias between backdoor models and backdoor-free models, thereby allowing the trigger to evade the server-end defense methods.

\textbf{\textit{In summary}}, we generate the optimized trigger in the way that the classification boundary of the backdoor model is aligned as much as possible to that of the backdoor-free model. Deploying the optimized trigger in attack not only preserves the backdoor effects in the personalization process, but also circumvents the server-end and client-end defense methods.

\subsubsection{Backdoor Poisoned Local Training} 
By optimizing Eq.\ref{eq:trigger-tuning} using the training data owned by malicious clients, the adversary obtains an optimized trigger $\Delta^t$ for attacks, and sends $\Delta^t$ to all malicious clients. The malicious clients embed the optimized trigger $\Delta^t$ into the poisoned local training dataset $D_i^{mal}$ via Eq.\ref{eq:trigger_embed}. Then, the malicious clients train local models based on $D_i^{{mal}}$ and $D_i^{{nor}}$ and send their poisoned local model updates to the central server.

\input{table/table-nodefense}
\section{Empirical Evaluation}
\subsection{Experimental Settings}
\noindent
\textbf{Datasets, Triggers, and Model Structures.}
\textit{CIFAR-10}\cite{cifar10} and \textit{CIFAR-100}\cite{cifar10} are color image classification datasets, each consisting of 50,000 training instances and 10,000 testing instances, all with a size of $32\times32$. We use ResNet\cite{resnet} as the model architecture. The trigger is an $8\times8$ pixel pattern placed at the center of the image. For \textit{CIFAR-10}, the target class label for the poisoned input instances is set as \textit{airplane}, and for \textit{CIFAR-100}, it is designated as \textit{apple}.
\textit{Fashion-MNIST}\cite{fmnist} is a 10-class classification dataset with 60,000 training instances and 10,000 testing instances, each of size $28\times28$. We use LeNet\cite{lenet} as the model structure. The trigger, a $10\times10$ pixel pattern, is positioned at the center of the image. The target class label for the poisoned input instance is set as \textit{T-shirt}.
The \textit{N-BaIoT}\cite{iot} dataset is a botnet dataset that provides authentic traffic data gathered from 9 commercial IoT devices. The dataset consists of 7 million data instances, each with 115 features, supporting an 11-class classification task. We create a trigger using 23 features, with the target class label for the poisoned input instance set as \textit{gafgyt\_combo}. The model structure we employ is a 4-layer fully connected network.
{We show the \textit{PFedBA}-driven optimized triggers applied to the CIFAR-10, CIFAR-100, Fashion-MNIST, and N-BaIoT datasets in Appendix \ref{app:trigger}.  }

For each dataset, we employ a non-IID data distribution. Specifically, we utilize the Dirichlet distribution\cite{DBA} to sample the data instances allocated to each client, thereby introducing complexity and diversity into the data distribution across clients. For the Fashion-MNIST and N-BaIoT datasets, we set the $\alpha$ value of the Dirichlet distribution to 0.1, while for CIFAR-10 and CIFAR-100, the $\alpha$ value is set to 0.5.
{We provide a visual representation of the class label distribution among clients with various $\alpha$ values in Appendix \ref{app:labeldis}.}

\noindent
\textbf{The PFL System Settings.}
In this work, we focus on the following 10 PFL algorithms: \textit{FedAvg-FT}\cite{localfinetuning-1}, \textit{FedProx-FT}\cite{fedprox}, \textit{SCAFFOLD}\cite{scaffold}, \textit{Per-Fedavg(FO)}\cite{per-fedavg}, \textit{Per-Fedavg(HF)}\cite{per-fedavg},  \textit{pFedMe}\cite{pfedme}, \textit{Ditto}\cite{ditto}, \textit{FedBN}\cite{fedbn}, \textit{FedRep}\cite{fedrep}, and \textit{FedALA}\cite{fedala}.  
For all datasets, we assume a total of $N=100$ clients in the PFL system. In each global aggregation iteration, the central server randomly selects $N_p=10$ clients to participate in the global aggregation.
During the personalization stage, for \textit{FedAvg-FT}, \textit{FedProx-FT}, \textit{Per-FedAvg(FO)}, and \textit{Per-FedAvg(HF)}, we adapt the global model generated by these methods with 1 epoch (Fashion-MNIST and N-BaIoT) or 5 epochs (CIFAR-10 and CIFAR-100) of stochastic gradient descent using the local training data to produce personalized local models. For {the rest 6 PFL methods}, we create personalized local models for clients based on the hyperparameters suggested by the original research papers. See the Appendix \ref{app:pflsetting} for more detailed PFL system settings.


\noindent
\textbf{Baseline Attacks and Attack Settings.}
To organize a comparative study, we involve the following 5 baseline backdoor attacks. 
\textit{ModelRe}\cite{howtobackdoor} scales up the poisoned local model updates with a factor $\lambda$ before sending them to the central server. We set $\lambda$ to 20 for all the datasets.
\textit{PGD attack}\cite{tails} employs projected gradient descent to train the poisoned local models. It periodically project the parameters of the local poisoned models on a small ball centered around the global model of the previous iteration. This method thus poses the constraint that the poisoned local models stay relatively close to the global model to bypass Byzantine-resilient FL aggregation.
\textit{Sybil attack}\cite{foolsgold} controls multiple malicious clients to independently train their poisoned local models.
\textit{Neurotoxin}\cite{durable} targets model parameters with minimal magnitude changes during benign local training to insert durable backdoor triggers into the global model. We maintain a 1\% ratio for masked gradients, following the original paper.
\textit{CerP}\cite{LyuHWLWL023} enhances the stealthiness of backdoor attacks by optimizing triggers and explicitly controlling trigger-induced model deviations.
By default, we assume that 10\% of clients are malicious for all the backdoor attack methods. 
{ For all datasets, we set the poisoning rate to be 1/4.}

\input{table/table-fmnist}

\noindent
\textbf{Defense Methods.}
We study the attack performance of various backdoor attacks against different defense techniques.
We evaluate the effectiveness of backdoor attack methods against a range of defense techniques implemented on the central server, which include \textit{Multi-Krum}\cite{mult-krum}, \textit{Trimmed mean}\cite{trimmed_mean}, \textit{DnC}\cite{ndss2021}, and \textit{FLAME}\cite{flame}. Specifically, following \cite{flame}, we set $\lambda=0.001$ (representing the amount of adaptive noise used in \textit{FLAME}) for CIFAR-10, CIFAR-100, and Fashion-MNIST, and $\lambda=0.01$ for B-NaIoT. Additionally, we evaluate the performance of backdoor attacks when clients utilize the widely adopted backdoor mitigation methods \textit{NC}\cite{wang2019neural} and \textit{NAD}\cite{NAD}.

\noindent
\textbf{Evaluation Metrics.}
We consider two metrics to evaluate jointly the effectiveness of backdoor attacks against PFL. \textit{ACC} denotes the average clean test accuracy of benign clients' personalized local models on their respective clean test data. \textit{ASR} measures the average backdoor accuracy of benign clients' personalized local models on their trigger-embedded test data. {A successful backdoor attack against PFL is expected to produce as high as possible ACC and ASR on the personalized local models of the target benign clients. In this sense, given similar ACC levels, the attack methods reaching higher ASR are considered more effective, and vice versa.}  

\subsection{Experimental Results}
\label{sec:5.2}
\begin{figure}[t] 
\centering
\includegraphics[width=8.3cm, height=5.5cm]{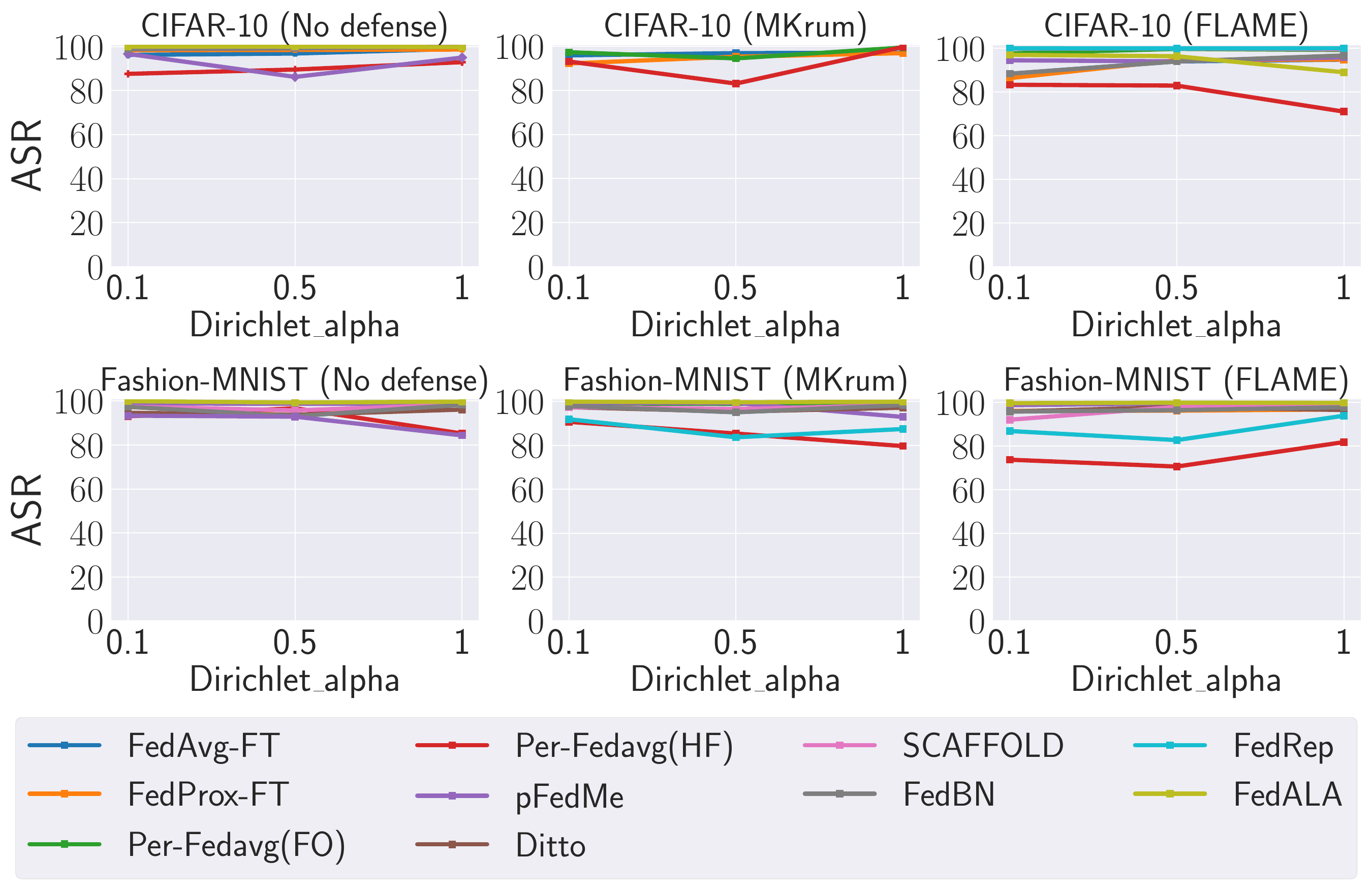}
\caption{ ASR for PFedBA against 10 PFL algorithms with the various degrees of non-IID data distribution on CIFAR-10 and Fashion-MNIST. }
\label{fig:noniid}
\end{figure}

\begin{figure}[t] 
\centering
\includegraphics[width=8.3cm, height=5.5cm]{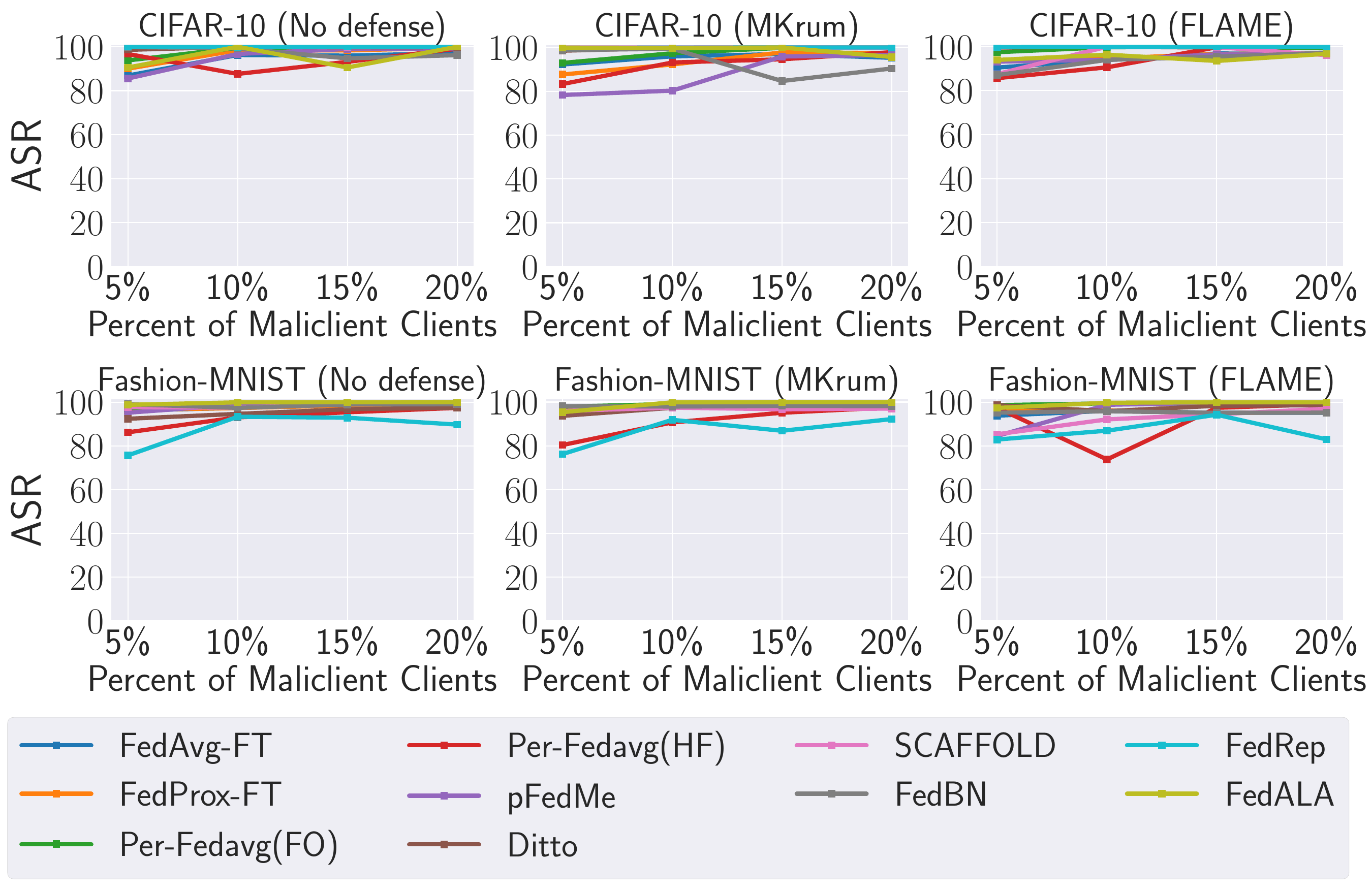}
\caption{ ASR for PFedBA against 10 PFL algorithms with various percentages of malicious clients on CIFAR-10 and Fashion-MNIST. }
\label{fig:malclient}
\end{figure}

We provide a comprehensive analysis of the threat posed by backdoor attacks on the PFL system across 4 datasets. We present ACC and ASR for 6 backdoor attacks and a no-attack scenario under various conditions.
More specifically, we present the attack performance of all the backdoor attacks and the no-attack scenario under 10 PFL algorithms without any defenses on N-BaIoT, Fashion-MNIST, CIFAR-10, and CIFAR-100 in Table \ref{tab:nodefense}.
Tables \ref{tab:fmnist}-- \ref{tab:cifar}, and \ref{tab:cifar100} illustrate the attack performance of all the backdoor attacks and the no-attack scenario for 10 PFL algorithms equipped with various server-end defense methods on Fashion-MNIST, N-BaIoT, CIFAR-10, and CIFAR-100, respectively. Due to space constraints and similar results derived on different datasets, we present the results for Fashion-MNIST, CIFAR-10, and N-BaIoT in this section, with results for CIFAR-100 available in Appendix \ref{app:attackresult}.
{Moreover, Table \ref{tab:client-defense} presents the ACCs and ASRs of various backdoor attacks against client-end defense methods across 10 PFL methods.}
In each table, we use bold fonts to highlight the highest ASR value achieved among all the attack methods (including \textit{PFedBA}), facing different defensive mechanisms. 
{To further demonstrate the effectiveness of \textit{PFedBA} on benign clients within the PFL system, we provide the ACC and ASR values of \textit{PFedBA} on benign clients across various datasets in Figure \ref{fig:accandasr} of Appendix \ref{app:resultforclient}.}
We also evaluate the sensitivity of the \textit{PFedBA}'s attack performance with respect to different alignment terms on Fashion-MNIST and CIFAR-10, shown in Tables \ref{tab:ablation-fmnist} and \ref{tab:ablation-cifar}. {The impacts of several parameters within PFL on \textit{PFedBA} are shown in Figures \ref{fig:noniid}, \ref{fig:malclient}, \ref{fig:numclient}-- \ref{fig:targetclass}. }

The experimental results show that \textit{PFedBA} outperforms the other baseline attack methods in terms of attack effectiveness across all the discussed scenarios. Further detailed observations and discussions are provided below.

\input{table/table-iot}

\noindent \textbf{(1) Attack performance without any defenses.}
As indicated in Table \ref{tab:nodefense}, across all 10 PFL algorithms without any defense mechanisms, our \textit{PFedBA} exhibits outstanding attack performance on personalized local models, while still maintaining the model's performance on the main learning task. In contrast, the other baseline attacks face challenges when attempting to inject triggers into personalized local models, even in scenarios where PFL systems do not employ defenses.

For instance, \textit{PFedBA} consistently achieves ASR values exceeding 80\% for all 10 PFL algorithms across the 4 datasets, while maintaining ACC values similar to the non-attacked model (NoAttack).
However, the attack baselines often struggle to successfully inject backdoor triggers into personalized local models. For Fashion-MNIST, the ASR values for all the attack baselines are below 47\% against \textit{Per-Fedavg(FO)}, \textit{pFedMe}, and \textit{FedRep}, while \textit{PFedBA} achieves ASR values exceeding 93\%.
On CIFAR-10 and N-BaIoT, under \textit{Per-FedAvg(HF)}, none of the baseline attacks can achieve ASR values exceeding 60\%, whereas \textit{PFedBA} achieves ASR values exceeding 82\%.
Similarly, the ASR values of the baseline attacks are lower than 51\% against \textit{pFedMe} on CIFAR-100, while \textit{PFedBA} obtains a 98.9\% ASR value.

The personalization process forces the personalized local models to forget the trigger-target mappings by adapting the local models with backdoor-free local training data. These attack baselines are designed without considering the impact of the personalization process with clean data on the backdoor effects. Therefore, embedding triggers into personalized local models within PFL systems becomes challenging for these baseline attacks.
\textit{PFedBA}, on the contrary, proactively aligns the gradients of the backdoor and main tasks by optimizing the trigger. This design enables the model to exploit as much trigger-related information as possible from the gradient of the main task. Moreover, by aligning the main task with the backdoor task, the backdoor model exhibits the decision boundary that closely resembles that of the clean model, so that the trigger can persist in personalized local models.

\input{table/table-cifar10}

\noindent \textbf{(2) Attack performance against server-end defenses.}
As shown in Tables \ref{tab:fmnist}--\ref{tab:cifar} in this section and Table \ref{tab:cifar100} in Appendix \ref{app:attackresult}, \textit{PFedBA} achieves significantly high ASR values against the server-end defenses under 10 PFL algorithms on 4 datasets without harming the accuracy of the main task.

For instance, on Fashion-MNIST, \textit{PFedBA} consistently achieves ASR values exceeding 87\% against all defenses across all PFL algorithms, except \textit{Per-Fedavg(HF)}. Even against \textit{Per-Fedavg(HF)}, \textit{PFedBA} still obtains more than 73\% of ASR values against all the defenses. Notably, the ACC values of the main task on the personalized local models attacked by \textit{PFedBA} remain very close to those of unattacked models. However, none of the attack baselines can achieve an ASR higher than 58\%.  
For N-BaIoT, \textit{PFedBA} achieves ASR values greater than 93\% against \textit{Multi-Krum} and \textit{FLAME} under \textit{Per-Fedavg(FO)} and \textit{pFedMe}, while the attack baselines fail to attain ASR values higher than 37\%.
Similarly, on CIFAR-10, \textit{PFedBA} achieves ASR values higher than 94\% against \textit{Trimmed mean} under \textit{FedAvg-FT} and \textit{FedProx-FT}, outperforming other attack baselines with ASR values lower than 40\%.
For CIFAR-100, on \textit{pFedMe}, \textit{PFedBA} obtains a 98.1\% ASR value against \textit{FLAME}, which is at least 34\% higher than the ASR values achieved by all the other attack baselines.

{
We further discuss the attack performance of \textit{PFedBA} against more sophisticated defense methods: \textit{FLTrust} \cite{DBLP:conf/ndss/CaoF0G21}, \textit{FedRecover} \cite{DBLP:conf/sp/CaoJZG23}, and \textit{FLCert} \cite{DBLP:journals/tifs/CaoZJG22}. The experimental results consistently demonstrate that \textit{PFedBA} achieves ASR values above 70\% against these advanced defense methods. For detailed discussions, please refer to Appendix \ref{app:fltrust}.}

These attack baselines fail to minimize the trigger-induced model bias, which makes them difficult to bypass defense methods deployed at the central server. Although \textit{CerP} can adapt a pre-specified trigger to reduce malicious model biases, which help bypass the server-end defenses. However, this method is restricted from two perspectives. Firstly, it can only slightly change the given trigger with a strict L2-ball constraint posed to the trigger optimization. In contrast, \textit{PFedBA} offers significantly greater tuning flexibility by allowing the adversary to generate and optimise the task-specific trigger. Second, \textit{PFedBA} enforces both gradient and loss alignment objectives to explicitly narrow down the gap between the gradient descent trajectories of the backdoor and clean models. Furthermore, the loss alignment enables the models to learn the trigger without significantly changing the model parameters. Consequently, \textit{PFedBA} outperforms the other attacks by producing higher ASR facing the deployed server-end defense methods. For example, on Fashion-MNIST, the ASR value of \textit{PFedBA} is at least 40\% higher than that of other attack baselines. The results confirm that \textit{PFedBA} performs well in evading the server-end defense methods and retaining more backdoor poisoning effects than the other baselines.

\noindent \textbf{(3) Attack performance against client-end defenses.} 
{Table \ref{tab:client-defense} presents the ACCs and ASRs of various attack methods against \textit{NC} and \textit{NAD}. }
Our findings reveal that backdoor mitigation techniques applied by benign clients are unable to remove the trigger embedded in personalized local models by \textit{PFedBA}. ASR of \textit{PFedBA} consistently exceeds 70\% against \textit{NC} and \textit{NAD} across all 10 PFL methods on Fashion-MNIST and B-NaIoT. Conversely, backdoor triggers introduced by the other attack baselines are identified and removed by \textit{NC} and \textit{NAD}.
For example, on Fashion-MNIST and B-NaIoT, all the attack baselines against \textit{NC} achieve a maximum ASR value of only 39.9\%. However, \textit{PFedBA} can achieve an ASR value of at least 70\%, successfully embedding the trigger into the personalized local models. Similarly, against \textit{NAD} on Fashion-MNIST, all the attack baselines achieve an ASR value of no more than 61\%.

{The explanation behind can be summarized in the followings. On one hand, the baseline attacks cause significant differences in the decision boundaries between backdoor and backdoor-free models. The client-end backdoor mitigation methods can exploit the gap and thus effectively eliminate the backdoor poisoning effect.
On the other hand, the gradient and loss alignment optimization adopted in \textit{PFedBA} brings the decision boundary of the backdoor model as close as possible to that of the backdoor-free model. Therefore, \textit{NC} and \textit{NAD} cannot distinguish the backdoor model from the backdoor-free models, allowing \textit{PFedBA} to bypass these client-end defense mechanisms.}

\input{table/table-client-defense}
\noindent
\textbf{(4) Ablation Study of PFedBA. }
{\textit{PFedBA} optimizes the trigger through both gradient alignment and loss alignment to enhance its attack effectiveness against PFL systems.
As detailed in Section \ref{sec:4.1}, the parameter $\lambda$ determines the method of trigger optimization. 
In \textit{PFedBA}, we consider only two cases for $\lambda$: $\lambda=0$ or $\lambda=1$, representing gradient alignment or loss alignment for trigger optimization, respectively. 
In this section, we evaluate the effectiveness of both alignment modules utilized in \textit{PFedBA} on CIFAR-10 and Fashion-MNIST. }
Due to space limitations, we show the results on Fashion-MNIST and CIFAR-10 in Tables \ref{tab:ablation-fmnist} and \ref{tab:ablation-cifar} in Appendix \ref{app:ablation}.

By observing the results, we find that both gradient alignment and loss alignment modules are necessary to ensure the success of \textit{PFedBA}. Removing any of these modules significantly decreases the ASR value of \textit{PFedBA}. 
For example, on Fashion-MNIST, \textit{PFedBA} achieves ASR values exceeding 87\% against \textit{DnC} and \textit{FLAME} under \textit{FedRep}. In contrast, the variant of \textit{PFedBA} without gradient alignment (referred to as No-Gradient) yields an ASR value of less than 27\%.
Besides, on Fashion-MNIST, \textit{PFedBA} attains ASR values over 89\% against \textit{Multi-Krum} and \textit{DnC} under \textit{Per-Fedavg(HF)}. However, the variant of \textit{PFedBA} without loss alignment (referred to as No-Loss) fails to conduct a successful attack, with ASR values below 11\%. 
Similarly, on CIFAR-10, the absence of any alignment module reduces the ASR values of \textit{PFedBA}, especially under \textit{pFedMe} and \textit{FedRep}.
These results underline the necessity of both the gradient alignment and loss alignment modules in the \textit{PFedBA} attack framework.

In summary, \textit{PFedBA} consistently and effectively injects backdoor triggers into personalized local models across various PFL scenarios. {Optimizing triggers to align the backdoor task with the main task is crucial for successful backdoor attacks against PFL systems. We further assess the alignment-driven regularization by computing Euclidean distances between backdoor local models and the backdoor-free global model in Appendix \ref{app:i2distance}. As shown in Table \ref{tab:l2-distance}, \textit{PFedBA} demonstrates the smallest distance among all the attack methods, indicating effective reduction of trigger-induced model bias.}

\noindent
\textbf{(5) Impact of Personalized FL Parameters. }
We study the impact of several parameters in PFL on \textit{PFedBA}, such as the degree of non-IID data distribution, the percentage of malicious clients, the total number of clients, {the trigger size, and the target class}.


\noindent
\textbf{Impact of the Degree of Non-IID Data Distribution.}
Figure \ref{fig:noniid} illustrates the impact of varying degrees of non-IID data distribution on the attack performance of \textit{PFedBA} against PFL on CIFAR-10 and Fashion-MNIST. Smaller $\alpha$ values indicate a higher degree of non-IIDness in the data distribution. We find that \textit{PFedBA} can obtain high ASR values under different degrees of non-IID data distribution. 
For instance, on both datasets, the ASR values of \textit{PFedBA} against all the PFL algorithms with varying degrees of non-IID data distribution are above 80\%. Furthermore, even though the server deploys multiple defense methods, our attack is still able to successfully embed backdoor triggers into the personalized local model, achieving ASR values above 70\% across all the 10 PFL algorithms under different non-IID data distributions on both two datasets. 
These results indicate that \textit{PFedBA} raises challenges to the PFL system with different non-IID data distributions.

\noindent
\textbf{Impact of the Percentage of Malicious Clients.} Figure \ref{fig:malclient} depicts the change in ASR of \textit{PFedBA} as the percentage of malicious clients increases from 5\% to 20\% on CIFAR-10 and Fashion-MNIST. We observe that \textit{PFedBA} can successfully execute backdoor attacks with just a small fraction of malicious clients. On both the Fashion-MNIST and CIFAR-10 datasets, even with only 5\% of malicious clients, \textit{PFedBA} can achieve an ASR value of over 70\%, regardless of whether the server deploys defensive measures.  
These results illustrate that \textit{PFedBA} can maintain strong attack performance even with a limited number of malicious clients.

\noindent
\textbf{Impact of the Total Number of Clients.} 
Figure \ref{fig:numclient} in Appendix \ref{app:numberclient} shows the \textit{PFedBA}'s ASR changes as the total number of clients ranges from 50 to 200 on N-BaIoT and Fashion-MNIST, with 10\% of clients being malicious. The results demonstrate that \textit{PFedBA} achieves ASR values exceeding 70\% across 10 PFL algorithms, regardless of the total number of clients or the presence of the defense mechanisms on the server. More details are available in Appendix \ref{app:numberclient}.

\noindent
{\textbf{Impact of the Trigger Size.} In Appendix \ref{app:triggersize}, Figure \ref{fig:triggersize} illustrates the attack performance of \textit{PFedBA} using various trigger sizes on CIFAR-10 and N-BaIoT datasets. The results indicate that \textit{PFedBA} consistently achieves ASR values exceeding 70\% across 10 PFL algorithms, regardless of the trigger sizes used. Further details are provided in Appendix \ref{app:triggersize}.}

\noindent
{\textbf{Impact of the Target Class.} Figure \ref{fig:targetclass} in Appendix \ref{app:targetclass} shows the attack performance of \textit{PFedBA} on the CIFAR-10 and N-BaIoT datasets with different target classes. The results demonstrate that \textit{PFedBA} consistently achieves ASR values above 80\% across various target classes on 10 PFL algorithms. For further details, please refer to Appendix \ref{app:targetclass}.}



\section{Takeaways}
As revealed in our study, we can summarize the vulnerability of PFL facing backdoor attacks from two perspectives: 

\noindent\textbf{Compared to conventional FL, PFL gains more resilience to backdoor attacks by the personalization process and defense methods deployed on both the central server and local clients.}
First, the personalized local model forgets the knowledge of triggers during the personalization process by learning information from clean local datasets.  Second, defense methods employed at the central server can minimize the impact of malicious local model updates on the global model during the global aggregation stage. Finally, backdoor mitigation methods deployed on benign clients can eliminate the presence of backdoor triggers in the personalized local model. Consequently, the integration of the personalization process and defense mechanisms enhances PFL's resilience to backdoor attacks.

\noindent\textbf{Aligning the backdoor task with the main task is the key-to-success in embedding the triggers into the personalized local models in PFL.} 
Despite the strong resilience of PFL against backdoor attacks, we can still carry out successful and stealthy backdoor attacks by enforcing the loss alignment and the gradient alignment between the main task and the backdoor task. On the one hand, enforcing both alignments can decrease the expected backdoor classification loss in the local model of benign clients during the personalization process, thus preventing the personalized local model from forgetting the embedded backdoor triggers. On the other hand, aligning the main task and the backdoor task narrows the gap between the decision boundary of the backdoor model and that of the backdoor-free model. This alignment enables the adversary to circumvent the defense methods implemented on the server or local clients and embed the backdoor into the personalized local model. This backdoor attack raises a severe security concern regarding model integrity in practical PFL applications. Therefore, we appeal to the security community to devise mitigation strategies that can effectively counter these risks. 

\section{Conclusion}
This study presents a comprehensive analysis of the threat posed by backdoor attacks on the PFL system. Our findings illustrate that the PFL system can mitigate the backdoor poisoning effect by using its unique personalization process. Additionally, deploying defense mechanisms on both the central server and local clients can further enhance the PFL system's resistance to backdoor attacks. Nonetheless, we introduce a new backdoor attack method called \textit{PFedBA}, which utilizes the alignment of loss and gradient between the main learning task and the backdoor learning task to highlight the vulnerabilities of robust-enhanced PFL systems. Our extensive experiments demonstrate that \textit{PFedBA} can effectively inject backdoor triggers into personalized local models, circumventing defense methods within the PFL system. 
{In future work, our aim is to develop a defense method inspired by the concept of randomized smoothing. This method will adaptively optimize the magnitude of added noise to mitigate backdoor poisoning effects while preserving the performance of the main task to the greatest extent possible.}

\section{Acknowledgment}
We thank the anonymous shepherd and reviewers for their valuable feedback. This work is supported by the National Key Research and Development Program of China under grant No. 2022YFB3102100, the Beijing Natural Science Foundation under grant No. L221014, the Systematic Major Project of China State Railway Group Corporation Limited under grant No. P2023W002, and the French National Research Agency with the reference ANR-23-IAS4-0001 (CKRISP).


\bibliographystyle{plain}
\bibliography{ref.bib}

\begin{figure*}[t] 
\centering
\includegraphics[width=14cm, height=4cm]{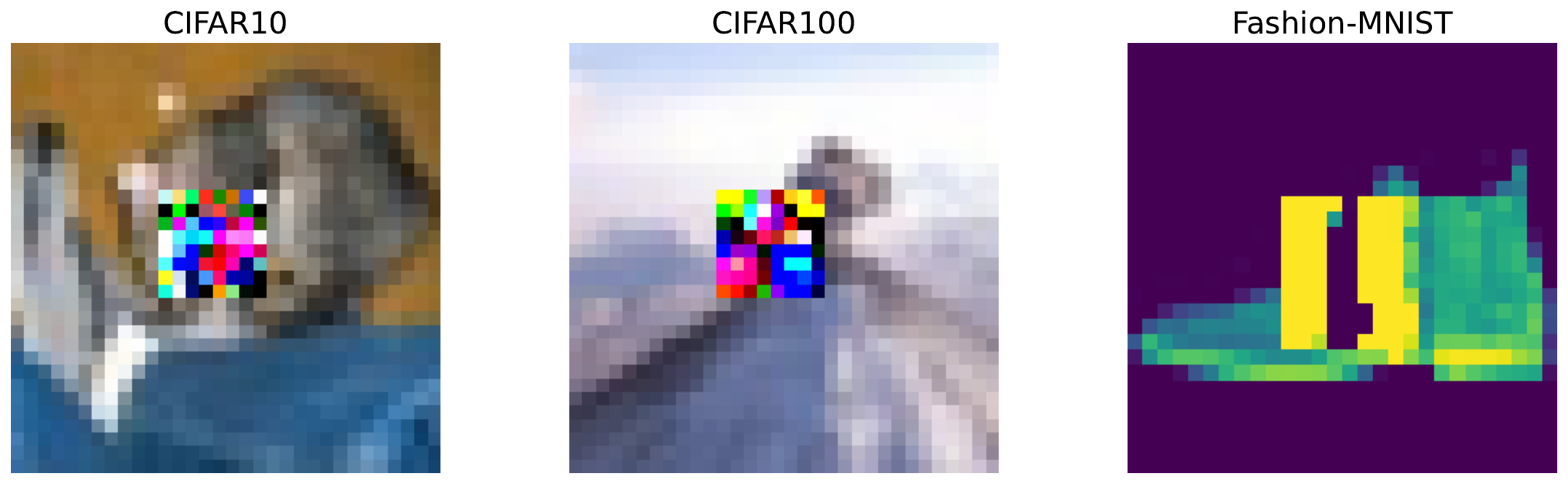}
\caption{{Visualization of the \textit{PFedBA}-driven optimized triggers for CIFAR-10, CIFAR-100, and Fashion-MNIST.}}
\label{fig:trigger}
\end{figure*}

\begin{figure*}[t] 
\centering
\includegraphics[width=14cm, height=5cm]{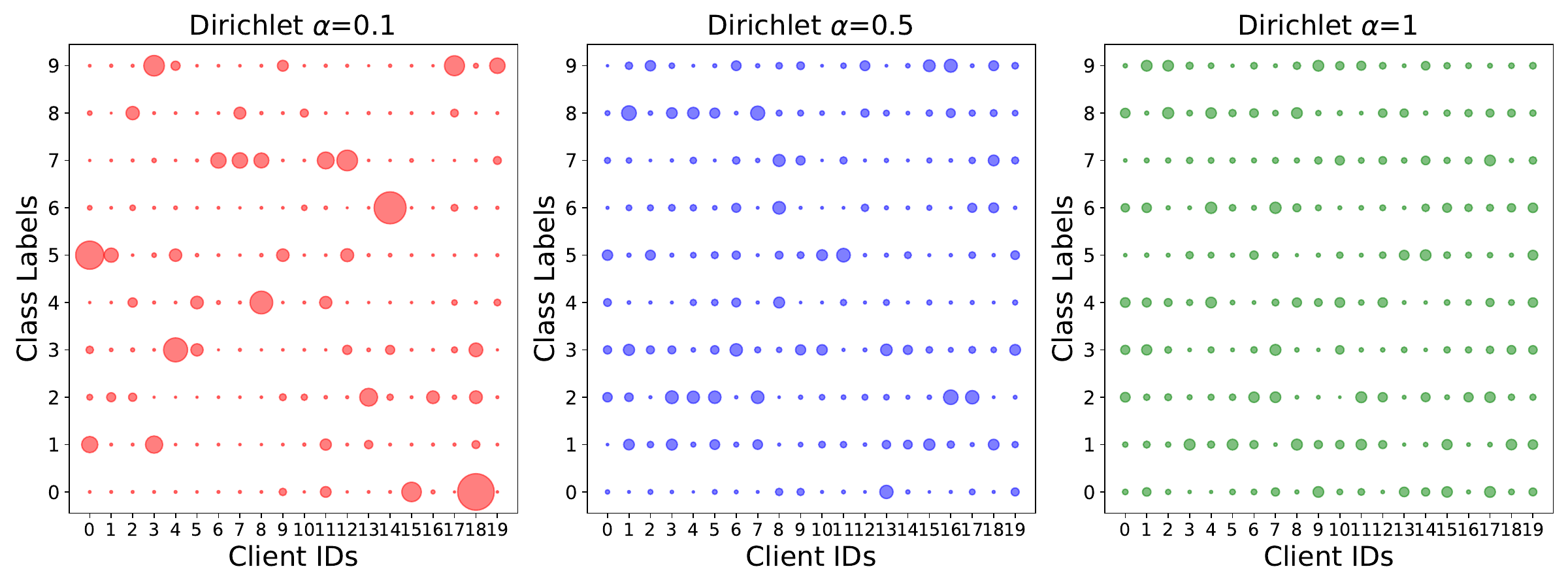}
\caption{{Visualization of the number of samples per class allocated to each client (indicated by dot sizes) on Fashion-MNIST for various values of Dirichlet distribution $\alpha$.}}
\label{fig:label}
\end{figure*}

\begin{appendices}
{\section{Visualization of Triggers}\label{app:trigger}
In \textit{PFedBA}, we optimize the backdoor trigger through gradient alignment and loss alignment between the main and backdoor tasks. The trigger generated at the convergence of the backdoor-poisoned federated training process serves as the final trigger for the attack during testing, achieving minimal classification loss and high ASRs for the backdoor task.
To illustrate the backdoor trigger optimized by \textit{PFedBA}, we visualize the optimized triggers applied to the CIFAR-10, CIFAR-100, Fashion-MNIST, and B-NaIoT datasets. In Figure \ref{fig:trigger}, we show the \textit{PFedBA}-driven optimized triggers based on FedAvg-FT for the CIFAR-10, CIFAR-100, and Fashion-MNIST datasets. Additionally, for the B-NaIoT tabular dataset, the adversary selects a subset of features from the feature space of data samples to serve as the trigger, assigning specific values predefined by the adversary to these features. Table \ref{tab:feature_values} shows the \textit{PFedBA}-driven optimized feature values based on FedAvg-FT for B-NaIoT.}

\begin{table}[ht]
\centering
\small
\caption{{The \textit{PFedBA}-driven optimized feature values for the N-BaIoT dataset.}}
\label{tab:feature_values}
\begin{tabular}{l|l|l}
\hline
Feature ID: Value & Feature ID: Value & Feature ID: Value \\
\hline
11: 5.4612        & 12: -4.3340       & 13: -2.4734       \\
14: 10.2924       & 15: -7.5629       & 16: 7.5796        \\
17: 4.0859        & 18: -2.3738       & 19: 7.2500        \\
20: 6.6069        & 21: 2.8452        & 22: 18.7352       \\
23: 1.4203        & 24: 2.1966        & 25: 26.7083       \\
26: 8.1407        & 27: 1.0658        & 28: 3.4069        \\
29: 0.8732        & 30: 3.2248        & 31: 16.3804       \\
32: -2.2518       & 33: 5.3170        &                    \\
\hline
\end{tabular}
\end{table}

{\section{Visualization of the Class Label Distribution} \label{app:labeldis}
{Figure \ref{fig:label} shows the distribution of the number of samples per class among 20 clients with varying values of the Dirichlet distribution parameter $\alpha$ on the Fashion-MNIST dataset. As the Dirichlet distribution parameter $\alpha$ decreases, the discrepancies in local data distribution among clients become more noticeable. Particularly, when $\alpha$ is set to 0.1, an extremely non-IID situation emerges in the local data distribution among clients.}}

\section{The PFL System Settings}\label{app:pflsetting}
\input{table/table-cifar100}

In this work, we focus on the following 10 PFL algorithms: \textit{FedAvg-FT}\cite{localfinetuning-1}, \textit{FedProx-FT}\cite{fedprox}, \textit{SCAFFOLD}\cite{scaffold}, \textit{Per-Fedavg(FO)}\cite{per-fedavg}, \textit{Per-Fedavg(HF)}\cite{per-fedavg},  \textit{pFedMe}\cite{pfedme}, \textit{Ditto}\cite{ditto}, \textit{FedBN}\cite{fedbn}, \textit{FedRep}\cite{fedrep}, and \textit{FedALA}\cite{fedala}.

For all datasets, we assume a total of $N=100$ clients in the PFL system. In each global aggregation iteration, the central server randomly selects $N_p=10$ clients to participate in the global aggregation.
During the personalization stage, for \textit{FedAvg-FT}, \textit{FedProx-FT}, \textit{Per-FedAvg(FO)}, and \textit{Per-FedAvg(HF)}, we adapt the global model generated by these methods with 1 epoch (Fashion-MNIST and N-BaIoT) or 5 epochs (CIFAR-10 and CIFAR-100) of stochastic gradient descent using the local training data to produce personalized local models. For {the rest 6 PFL methods}, we create personalized local models for clients based on the hyperparameters suggested by the original research papers. 

\input{table/paramters}
Additionally, for CIFAR-100 and CIFAR-10, the local learning rate is set to 0.1 for all the PFL algorithms. For Fashion-MNIST and N-BaIoT, the local learning rate is set to 0.1 for all the PFL algorithms except \textit{Per-FedAvg(FO)}, \textit{Per-FedAvg(HF)}, and \textit{pFedMe}, for which the local learning rate is set to 0.01. The batch size is consistently set at 64 for all datasets, except for N-BaIoT, where a batch size of 128 is employed. For N-BaIoT, Fashion-MNIST, CIFAR-10, and CIFAR-100, the total number of global training iterations is set to 100, 150, 200, and 300, respectively.

\section{Server-end Defenses on CIFAR-100}\label{app:attackresult}

Table \ref{tab:cifar100} displays the ACC and ASR values for various backdoor attack methods against the 10 PFL methods equipped with distinct defense mechanisms implemented at the central server on CIFAR-100. 
The results demonstrate the superior attack performance of \textit{PFedBA}. On the CIFAR-100 dataset, for \textit{pFedMe} with diverse defense mechanisms, \textit{PFedBA} achieves ASR values exceeding 97\%, which are at least 22\% higher than the ASR values of the other baseline attack methods. 
Similarly, the ASR values of the other baseline attacks against \textit{FLAME} on \textit{FedAvg-FT} are below 73\%, while our \textit{PFedBA} attack achieves an ASR value exceeding 96\%.

Our gradient and loss alignment-based \textit{PFedBA} method can successfully embed backdoor triggers into personalized local models, which is difficult for the other attack baseline methods.

{\section{Evaluation on Advanced Defenses}\label{app:fltrust}
\input{table/table-fltrust}
\input{table/table-fedrecover}
In this section, we evaluate the effectiveness of \textit{PFedBA} against \textit{FLTrust} \cite{DBLP:conf/ndss/CaoF0G21}, \textit{FedRecover} \cite{DBLP:conf/sp/CaoJZG23}, and \textit{FLCert} \cite{DBLP:journals/tifs/CaoZJG22}.}

{\textit{FLTrust} \cite{DBLP:conf/ndss/CaoF0G21} allows the central server to maintain a small, poison-free dataset sampled from a distribution similar to the entire training dataset. Using this dataset, the server trains a global reference model and computes a linear combination of local models weighted by the score, measured by the cosine similarity between the global reference model and local models. We evaluate the attack performance of \textit{PFedBA} against \textit{FLTrust} on the Fashion-MNIST, CIFAR-10, and B-NaIoT datasets.  As shown in Table \ref{tab:fltrust}, \textit{PFedBA} consistently achieves ASR values exceeding 71\% across 10 PFL algorithms.}

{
\textit{FedRecover} \cite{DBLP:conf/sp/CaoJZG23} is a model recovery method capable of restoring clean global models from poisoning attacks using stored historical information, with minimal computation and communication costs for clients. Table \ref{tab:fedrecover} presents the results of \textit{PFedBA} against \textit{FedRecover} on the Fashion-MNIST, CIFAR-10, and B-NaIoT datasets. Across all 10 PFL algorithms evaluated, \textit{PFedBA} consistently achieves ASR values exceeding 72\% against \textit{FedRecover}.}

{
\textit{FLCert} \cite{DBLP:journals/tifs/CaoZJG22} constructs an ensemble of global models and employs a majority vote among these global models to make prediction decisions. Specifically, it randomly selects $k$ clients out of $n$ to train global models, resulting in a total of $N$ global models. A majority vote among these global models is then used to predict the label for a testing input.
However, \textit{FLCert} faces limitations when applied to PFL systems due to several reasons. First, \textit{FLCert} randomly selects clients to train global models and then employs voting among these global models for final predictions. However, in partial model-sharing PFL systems where only the feature encoder is globally trained, omitting the classification head, conducting a vote among multiple global models for prediction results becomes infeasible.
Second, according to the setting in \cite{DBLP:journals/tifs/CaoZJG22}, the FL system consists of 1,000 local clients, training a total of $N=500$ global models to form the ensemble. Therefore, each client in PFL systems would need to fine-tune these 500 global models during the personalization process. This results in the need to fine-tune $1,000 \times 500=500,000$ personalized local models in the PFL system, leading to excessive computational overhead.}

{To assess the attack performance of \textit{PFedBA} against \textit{FLCert}, we conduct experiments on the Fashion-MNIST dataset using \textit{FedAvg-FT} and \textit{Per-FedAvg(FO)}, comprising 20 local clients, with 20\% of them being malicious clients. We randomly select 4 out of the 20 clients to train the global model, resulting in a total of 500 global models being trained. Consequently, the PFL systems maintain $20 \times 500=10,000 $ personalized local models.
In these tested cases, the ACC value of \textit{PFedBA} exceeds 84.5\%, and the ASR value surpasses 90.0\% for both PFL methods.}

\begin{figure*}[t] 
\centering
\includegraphics[width=17cm, height=12.0cm]{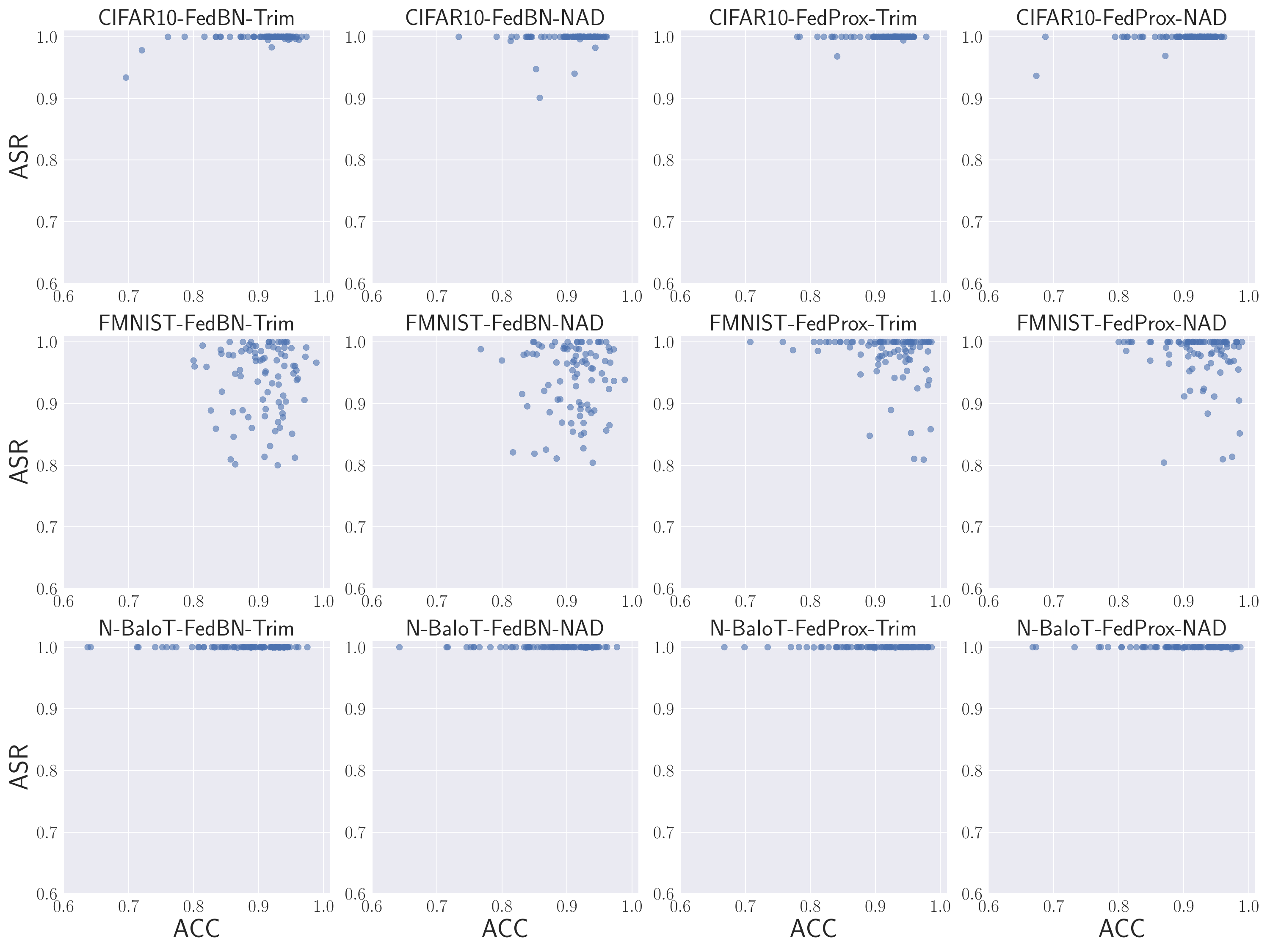}
\caption{{ Visualization of ACCs and ASRs of PFedBA on benign clients.} }
\label{fig:accandasr}
\end{figure*}
{\section{Attack Performance on Benign Clients} \label{app:resultforclient}
Figure \ref{fig:accandasr} illustrates the attack performance of \textit{PFedBA} on benign clients, displaying the ACC and ASR values of \textit{PFedBA} on all benign clients across different datasets in both partial and full model-sharing PFL systems, against various defense methods. Our results demonstrate that \textit{PFedBA} consistently achieves ASR values exceeding 80\% across all benign clients, irrespective of the datasets, PFL systems, or defense methods considered.  This consistent success of the \textit{PFedBA} attack across all benign clients underscores its effectiveness against PFL systems.}

\section{The Euclidean Distance}\label{app:i2distance}
\input{table/l2-distance}
Table \ref{tab:l2-distance} shows the Euclidean distance between the backdoor local model generated by various backdoor attacks and the backdoor-free global model obtained in the previous global model training iteration in the full model-sharing PFL methods. These results are derived from the Fashion-MNIST dataset.
We observe that, under the meta learning-based PFL methods, including \textit{Per-Fedavg(FO)}, \textit{Per-Fedavg(HF)}, and \textit{pFedMe}, all the backdoor attacks result in relatively small model biases. This is because these PFL methods aim to find an initial global model that can quickly adapt to the client's local data, resulting in less model biases in the backdoor local models.

More importantly, we find that our \textit{PFedBA} method produces the least Euclidean distance between the backdoor local model and the backdoor-free global model aggregated at the previous iteration. This observation confirms the design of the gradient alignment and loss alignment in \textit{PFedBA}. The two alignment-driven regularization terms reduce the gap between the backdoor model and the backdoor-free model on the optimization path and decision boundary. Notably, the gradient alignment term explicitly reduces the difference between the gradient descent trajectories between the backdoor and backdoor-free models. Furthermore, enforcing the loss alignment term helps the backdoor model to better learn the trigger without significantly changing the model parameters. In summary, our \textit{PFedBA} attack method, based on gradient alignment and loss alignment, effectively mitigates the impact of the personalization process on the backdoor poisoning effect and enables the triggers to bypass both server-end and client-end defense mechanisms.

\input{table/ablation-fmnist-onlyl2}
\input{table/ablation-cifar-onlyl2}
\section{Ablation Study of PFedBA on CIFAR-10}\label{app:ablation}

The results in Tables \ref{tab:ablation-fmnist} and \ref{tab:ablation-cifar} show the sensitivity of the \textit{PFedBA}'s attack performance with respect to different alignment terms on Fashion-MNIST and CIFAR-10. 

The results indicate that integrating the gradient alignment module into the \textit{PFedBA} attack significantly improves its attack performance compared to the variant of \textit{PFedBA} without this module (noted as No-Gradient). 
Specifically, \textit{PFedBA} with gradient alignment consistently achieves ASR values exceeding 75\% against \textit{Trimmed mean}, \textit{DnC}, and \textit{FLAME} under \textit{Per-Fedavg(HF)} on CIFAR-10. In contrast, the variant of \textit{PFedBA} without gradient alignment fails to conduct a successful attack, resulting in ASR values below 40\%.
Moreover, the results also emphasize the importance of the loss alignment module for the success of \textit{PFedBA}. For instance, in CIFAR-10, the ASR values of the \textit{PFedBA} variant without the loss alignment module (referred to as No-Loss) falls below 25\% against \textit{DnC} when using \textit{Per-Fedavg(FO)} and \textit{Per-Fedavg(HF)}.
Similarly, in Fashion-MNIST, the absence of any alignment module reduces
the ASR values of \textit{PFedBA}, especially under \textit{Per-Fedavg(HF)}, \textit{pFedMe} and \textit{FedRep}.

In conclusion, these results validate the indispensability of both gradient alignment and loss alignment modules within the \textit{PFedBA} attack framework.

\section{Impact of the Total Number of Clients}\label{app:numberclient}
\begin{figure}[t] 
\centering
\includegraphics[width=8.3cm, height=5.5cm]{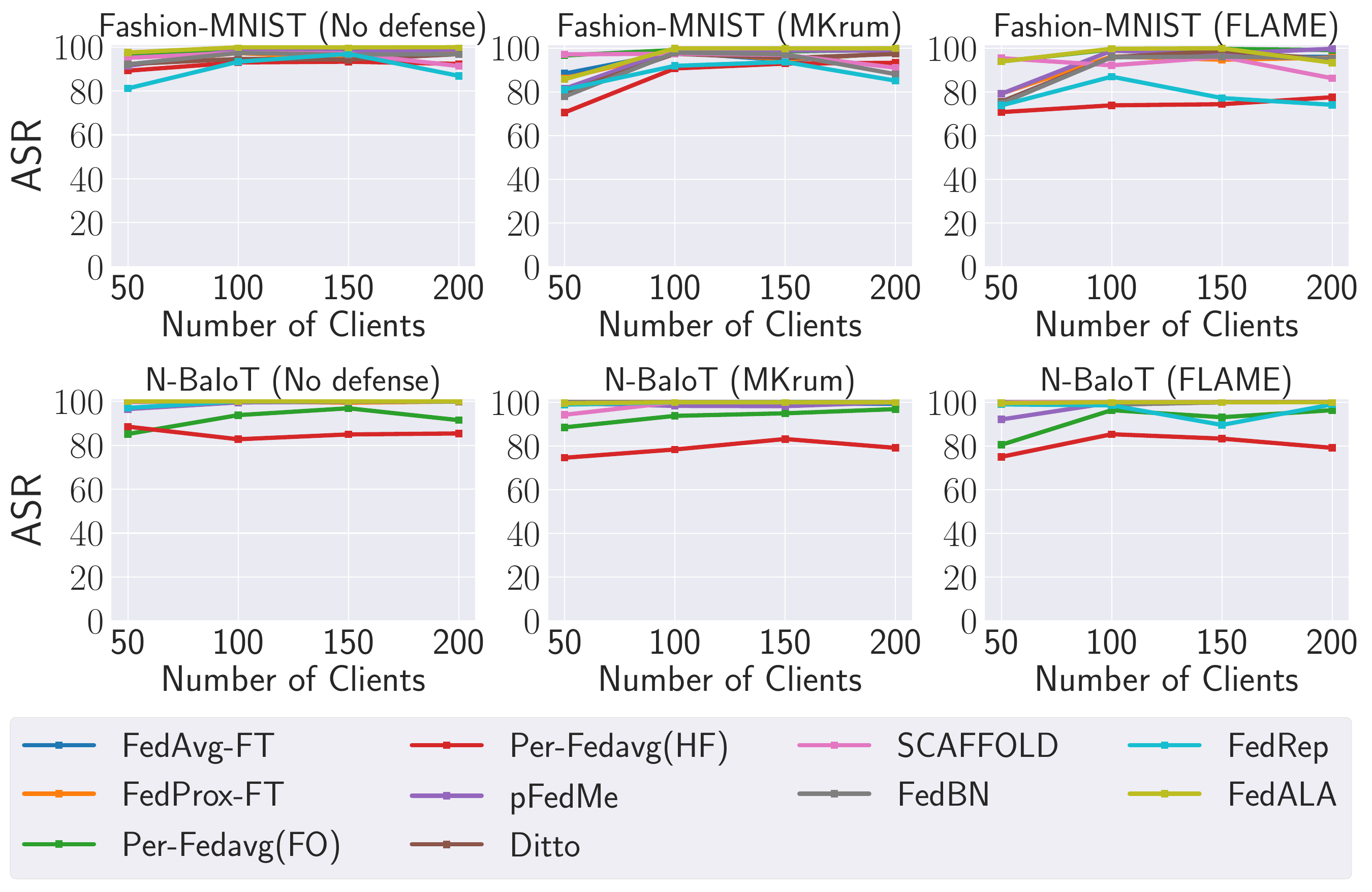}
\caption{ ASR for PFedBA against 10 PFL algorithms with the various total numbers of clients on N-BaIoT and Fashion-MNIST. }
\label{fig:numclient}
\end{figure}
Figure \ref{fig:numclient} depicts the change of \textit{PFedBA}'s ASR as the total number of clients varies from 50 to 200 on the N-BaIoT and Fashion-MNIST datasets, with 10\% of clients being malicious. The results indicate that the attack performance of \textit{PFedBA} remains consistent even when the number of clients increases. 
Specifically, \textit{PFedBA} can achieve more than 80\% ASR values under all the PFL algorithms on N-BaIoT and Fashion-MNIST, regardless of the total number of clients. Additionally, even with defense mechanisms deployed at the central server, \textit{PFedBA} can still attain an ASR value greater than 70\% against all the PFL methods on both datasets. These experimental results validate the effectiveness of \textit{PFedBA}.

{\section{Impact of the Trigger Size}\label{app:triggersize}
\begin{figure}[t] 
\centering
\includegraphics[width=8.3cm, height=5.5cm]{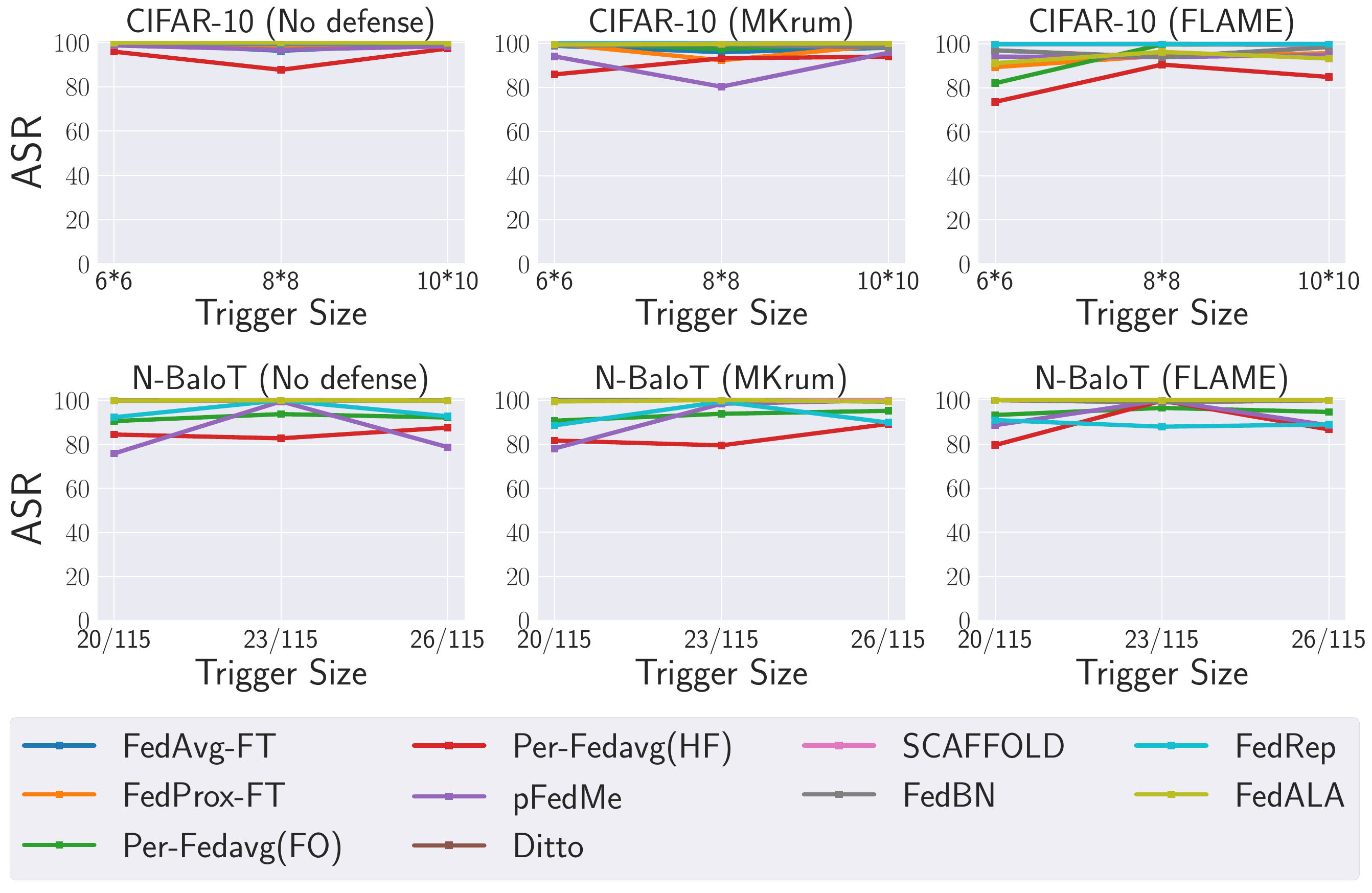}
\caption{ {ASR for PFedBA against 10 PFL algorithms with various trigger sizes on N-BaIoT and CIFAR-10.} }
\label{fig:triggersize}
\end{figure}
Figure \ref{fig:triggersize} illustrates the variation in ASR values for \textit{PFedBA} across different trigger sizes in the N-BaIoT and CIFAR-10 datasets. The results consistently show that \textit{PFedBA} achieves ASR values exceeding 70\% across all 10 PFL algorithms in both datasets, irrespective of the trigger size. This confirms the effectiveness of \textit{PFedBA}. Moreover, there is a slight decrease in the ASR values in certain tested scenarios as the trigger size decreases. This decline may be attributed to the smaller trigger size resulting in less flexibility in optimizing the trigger injected into the training data, thereby increasing the difficulty of achieving the desired attack objectives.}

{\section{Impact of the Target Class}\label{app:targetclass}
\begin{figure}[t] 
\centering
\includegraphics[width=8.3cm, height=5.5cm]{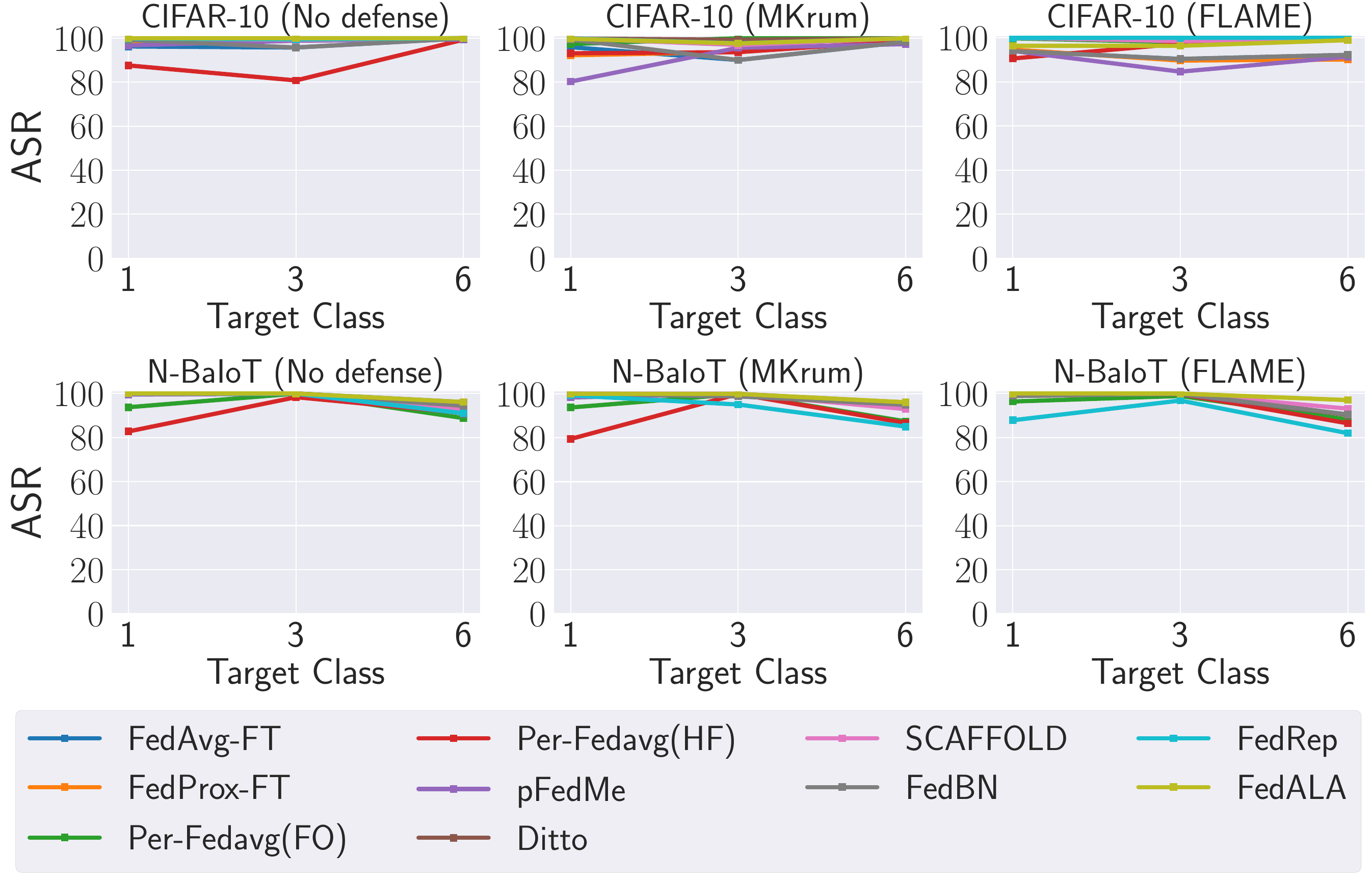}
\caption{{ASR for PFedBA against 10 PFL algorithms with various target classes on N-BaIoT and CIFAR-10. } }
\label{fig:targetclass}
\end{figure}
Figure \ref{fig:targetclass} shows how the ASR value of \textit{PFedBA} varies when using different target classes in the N-BaIoT and CIFAR-10 datasets.  The results indicate that \textit{PFedBA} maintains consistent attack performance across various target classes. Across all 10 PFL algorithms in the N-BaIoT and CIFAR-10 datasets, \textit{PFedBA} achieves ASR values exceeding 80\%, regardless of the target class.}

\end{appendices}

\end{document}

%% file: table/table-nodefense.tex
\begin{table*}[]
\centering
\caption{The ACC/ ASR of various backdoor attack methods against PFL without defenses.(\%) {Standard Deviation Range: 0.6\%-5\%.}}
\label{tab:nodefense}
\resizebox{\linewidth}{!}{
\begin{tabular}{|c|c|c|c|c|c|c|c|c|c|c|c|}
\hline
Datasets                       & Attacks    & FedAvg-FT   & FedProx-FT  & SCAFFOLD    & FO          & HF         & pFedMe     & Ditto       & FedBN       & FedRep      & FedALA      \\ \hline
\multirow{7}{*}{\begin{tabular}[c]{@{}c@{}}Fashion\\ MNIST\end{tabular}} & NoAttack   & 92.3/ 10.4  & 92.3/ 10.4  & 91.8/ 8.8   & 92.2/ 10.2  & 87.5/ 9.7  & 89.8/ 10.4 & 91.9/ 10.5  & 92.3/ 10.4  & 89.7/ 10.2  & 90.4/ 8.7   \\
                               & Sybil      & 91.9/ 49.6  & 91.9/ 49.3  & 91.5/ 51.8  & 92.2/ 33.2  & 87.6/ 14.9 & 89.7/ 19.0 & 91.9/ 38.6  & 91.9/ 49.5  & 90.0/ 29.1  & 89.8/ 53.4  \\
                               & ModelRe    & 92.2/ 10.6  & 92.2/ 10.7  & 92.7/ 10.8  & 92.2/ 10.6  & 87.5/ 9.8  & 89.9/ 10.5 & 91.8/ 10.8  & 92.4/ 10.7  & 89.8/ 10.3  & 90.4/ 9.0   \\
                               & PGD        & 92.0/ 25.9  & 92.0/ 25.9  & 91.5/ 26.4  & 92.0/ 22.5  & 87.8/ 20.4 & 89.6/ 28.6 & 91.4/ 51.6  & 92.0/ 25.8  & 89.9/ 17.8  & 89.9/ 55.1  \\
                               & Neurotoxin & 91.8/ 48.2  & 92.0/ 47.6  & 91.6/ 46.9  & 92.0/ 15.4  & 87.3/ 18.2 & 89.5/ 19.3 & 91.7/ 38.3  & 91.8/ 48.2  & 90.0/ 16.5  & 89.7/ 52.1  \\
                               & CerP & 92.2/ 62.9 & 92.1/ 63.4 & 92.5/ 69.5 & 92.2/ 46.7 & 87.6/ 18.9 & 89.8/ 24.6 & 91.7/ 46.5 & 92.2/ 62.9 & 90.6/ 37.9 & 89.8/ 70.2 \\
                               &  PFedBA   & 92.2/ \textbf{97.4}  & 92.2/\textbf{ 97.3}  & 91.7/ \textbf{97.4}  & 92.1/ \textbf{99.3}  & 87.5/ \textbf{93.0} & 89.7/ \textbf{98.8} & 91.7/ \textbf{94.6}  & 92.1/ \textbf{97.4}  & 89.9/ \textbf{93.3}  & 90.2/ \textbf{99.9}  \\ \hline
\multirow{7}{*}{CIFAR-10}      & NoAttack   & 90.3/ 6.6   & 90.4/ 7.5   & 92.1/ 7.2   & 86.7/ 7.1   & 78.2/ 6.5  & 87.1/ 8.0  & 91.9/ 8.3   & 90.3/ 6.6   & 92.2/ 7.2   & 80.9/ 11.5  \\
                               & Sybil      & 89.4/ 43.7  & 89.7/ 47.5  & 92.1/ 43.1  & 86.7/ 89.5  & 78.8/ 51.1 & 87.1/ 83.2 & 91.8/ 60.2  & 89.4/ 43.7  & 92.1/ 59.8  & 78.9/ 93.3  \\
                               & ModelRe    & 84.4/ 44.4  & 84.0/ 46.3  & 91.4/ 46.8  & 87.0/ 19.4  & 77.6/ 8.5  & 89.1/ 31.9 & 87.5/ 54.3  & 84.4/ 44.4  & 92.1/ 11.3  & 80.0/ 15.8  \\
                               & PGD        & 90.0/ 44.5  & 90.5/ 45.4  & 92.1/ 38.3  & 86.5/ 63.2  & 77.5/ 15.4 & 88.1/ 73.3 & 91.7/ 72.1  & 90.0/ 44.5  & 92.0/ 51.3  & 77.2/ 74.5  \\
                               & Neurotoxin & 90.1/ 47.3  & 90.0/ 56.6  & 91.9/ 37.4  & 86.8/ 80.5  & 77.2/ 38.9 & 87.6/ 83.5 & 91.4/ 60.0  & 89.4/ 52.7  & 92.2/ 55.9  & 79.1/ 58.9  \\
                               & CerP & 90.1/ 64.3 & 90.6/ 62.3 & 92.0/ 56.1 & 87.3/ 97.7 & 78.1/ 58.5 & 87.0/ 85.9 & 91.4/ 71.6 & 89.9/ 61.8 & 92.4/ 73.7 & 79.0/ 77.0 \\
                               &  PFedBA   & 90.0/ \textbf{98.7}  & 90.3/ \textbf{98.1}  & 92.2/ \textbf{99.6}  & 86.7/ \textbf{99.9}  & 78.2/ \textbf{83.7} & 87.7/ \textbf{99.4} & 91.7/ \textbf{99.7}  & 90.0/ \textbf{98.7}  & 92.3/ \textbf{99.4}  & 77.5/ \textbf{99.4}  \\ \hline
\multirow{7}{*}{CIFAR-100}     & NoAttack   & 77.7/ 1.1   & 77.7/ 1.3   & 77.1/ 1.0   & 64.9/ 1.5   & 64.3/ 1.8  & 62.2/ 0.9  & 78.0/ 1.2   & 77.7/ 1.1   & 78.2/ 1.4   & 77.9/ 1.0   \\
                               & Sybil      & 76.1/ 82.0  & 76.3/ 81.0  & 76.5/ 58.8  & 64.3/ 93.0  & 63.2/ 92.7 & 61.7/ 45.4 & 77.3/ 87.5  & 76.1/ 82.0  & 77.9/ 84.9  & 76.6/ 95.8  \\
                               & ModelRe    & 63.5/ 34.7  & 66.2/ 32.8  & 76.2/ 17.0  & 65.2/ 27.4  & 63.7/ 5.6  & 61.9/ 14.4 & 70.6/ 41.8  & 63.5/ 34.7  & 78.5/ 3.3   & 77.7/ 2.3   \\
                               & PGD        & 77.2/ 78.3  & 76.8/ 77.3  & 76.6/ 45.3  & 64.5/ 81.6  & 62.2/ 42.5 & 63.9/ 48.7 & 76.9/ 96.9  & 77.2/ 78.3  & 78.1/ 79.5  & 70.2/ 98.7  \\
                               & Neurotoxin & 76.3/ 79.2  & 75.3/ 79.7  & 76.5/ 38.6  & 64.7/ 84.0  & 61.7/ 40.8 & 62.1/ 48.1 & 77.5/ 84.9  & 76.3/ 79.2  & 77.3/ 82.0  & 76.8/ 96.5  \\
                               & CerP & 75.2/ 81.7 & 75.6/ 81.0 & 76.4/ 64.5 & 64.1/ 96.0 & 65.5/ 97.1 & 62.6/ 50.5 & 77.3/ 91.2 & 75.7/ 83.0 & 77.7/ 88.3 & 6.5/ 96.7 \\
                               &  PFedBA   & 76.9/ \textbf{100} & 76.8/ \textbf{100} & 76.3/ \textbf{99.9}  & 64.4/ \textbf{100} & 62.4/ \textbf{99.8} & 62.5/ \textbf{98.9} & 77.7/ \textbf{100} & 76.9/ \textbf{100} & 77.8/ \textbf{99.9}  & 77.2/ \textbf{100} \\ \hline
\multirow{7}{*}{ N-BaIoT}           & NoAttack   & 90.2/ 5.2   & 90.2/ 4.4   & 88.1/ 1.6   & 87.7/ 12.1  & 85.6/ 14.0 & 89.8/ 12.7 & 90.7/ 5.2   & 88.3/ 2.3   & 88.6/ 10.0  & 90.2/ 2.6   \\
                               & Sybil      & 90.1/ 92.7  & 90.1/ 94.2  & 87.9/ 98.2  & 87.9/ 75.7  & 86.7/ 53.3 & 89.8/ 74.2 & 90.7/ 95.9  & 87.9/ 93.0  & 88.7/ 61.0  & 89.9/ 89.9  \\
                               & ModelRe    & 90.4/ 31.5  & 90.2/ 32.7  & 88.2/ 15.9  & 87.0/ 29.6  & 85.5/ 26.4 & 89.7/ 19.3 & 90.7/ 32.8  & 88.0/ 19.6  & 88.8/ 16.2  & 90.2/ 17.5  \\
                               & PGD        & 90.0/ 86.1  & 90.1/ 85.9  & 87.9/ 80.3  & 87.8/ 75.2  & 85.3/ 60.2 & 89.8/ 99.3 & 90.6/ 99.9  & 88.2/ 86.3  & 89.0/ 94.8  & 89.7/ 97.4  \\
                               & Neurotoxin & 88.2/ 93.0  & 88.2/ 92.8  & 87.8/ 90.0  & 87.7/ 40.5  & 81.5/ 25.6 & 87.2/ 84.8 & 88.4/ 92.3  & 88.2/ 93.0  & 88.2/ 35.8  & 89.7/ 93.8  \\
                               & CerP & 90.1/ 95.4 & 90.1/ 94.3 & 90.7/ 92.9 & 87.9/ 80.4 & 86.1/ 56.5 & 89.8/ 78.5 & 90.8/ 98.7 & 90.1/ 95.4 & 91.6/ 89.6 & 91.7/ 95.1 \\
                               &  PFedBA   & 90.1/ \textbf{100} & 90.1/ \textbf{100} & 87.9/ \textbf{100} & 87.6/ \textbf{93.7}  & 85.7/ \textbf{82.7} & 89.7/ \textbf{99.5} & 90.8/ \textbf{100} & 88.3/ \textbf{100} & 88.6/ \textbf{100} & 90.0/ \textbf{100} \\ \hline
\end{tabular}
}
\end{table*}

%% file: table/table-fmnist.tex
\begin{table*}[]
\centering
\caption{The ACC/ ASR of various backdoor attacks against PFL with server-end defenses on Fashion-MNIST.(\%) {Standard Deviation Range: 0.3\%-4\%.}}
\label{tab:fmnist}
\resizebox{\linewidth}{!}{
\begin{tabular}{|c|c|c|c|c|c|c|c|c|c|c|c|}
\hline
Defenses               & Attacks    & FedAvg-FT  & FedProx-FT & SCAFFOLD   & FO         & HF         & pFedMe     & Ditto      & FedBN      & FedRep     & FedALA     \\ \hline
\multirow{7}{*}{MKrum} & NoAttack   & 91.9/ 10.5 & 91.9/ 10.5 & 91.5/ 8.7  & 92.2/ 10.5 & 87.7/ 9.7  & 89.8/ 10.4 & 91.8/ 10.5 & 92.0/ 10.5 & 89.0/ 10.5 & 89.4/ 8.7  \\
                       & Sybil      & 92.1/ 10.5 & 92.1/ 10.6 & 91.6/ 29.5 & 92.2/ 10.5 & 87.8/ 9.8  & 89.9/ 10.4 & 92.0/ 10.6 & 92.1/ 10.6 & 89.5/ 23.2 & 89.6/ 8.9  \\
                       & ModelRe    & 92.0/ 10.5 & 92.0/ 10.6 & 92.4/ 10.4 & 92.1/ 10.5 & 87.7/ 9.7  & 89.9/ 10.4 & 91.8/ 10.5 & 92.0/ 10.5 & 89.0/ 10.2 & 89.5/ 8.7  \\
                       & PGD        & 91.9/ 25.6 & 91.8/ 25.1 & 91.5/ 22.7 & 92.1/ 10.5 & 87.8/ 9.7  & 89.9/ 10.4 & 92.0/ 10.6 & 91.9/ 25.0 & 89.4/ 13.0 & 89.6/ 8.7  \\
                       & Neurotoxin & 92.0/ 10.6 & 92.0/ 10.6 & 91.5/ 9.0  & 92.0/ 18.2 & 87.7/ 9.9  & 90.0/ 10.4 & 91.9/ 10.7 & 92.0/ 10.6 & 88.6/ 11.1 & 89.5/ 8.8  \\
                       & CerP       & 92.0/ 10.6 & 92.1/ 11.0 & 92.3/ 42.5 & 91.9/ 9.3  & 87.2/ 9.1  & 90.0/ 10.7 & 92.1/ 9.6 & 91.4/ 9.8  & 88.9/ 22.7 & 89.6/ 9.3 \\
                       & PFedBA     & 91.9/ \textbf{98.1} & 91.8/ \textbf{98.3} & 91.3/ \textbf{97.4} & 92.0/ \textbf{99.1} & 87.7/ \textbf{90.7} & 89.8/ \textbf{99.3} & 91.7/ \textbf{97.4} & 91.8/ \textbf{97.9} & 89.1/ \textbf{91.9} & 89.3/ \textbf{99.9} \\ \hline
\multirow{7}{*}{Trim}  & NoAttack   & 92.1/ 10.5 & 92.2/ 10.5 & 91.5/ 8.8  & 92.2/ 10.5 & 87.7/ 9.6  & 89.8/ 10.4 & 91.8/ 10.6 & 92.1/ 10.6 & 89.3/ 10.2 & 89.8/ 8.7  \\
                       & Sybil      & 91.8/ 24.7 & 92.0/ 24.7 & 91.3/ 38.0 & 92.3/ 15.6 & 87.7/ 11.0 & 89.8/ 12.0 & 91.7/ 18.8 & 92.0/ 24.9 & 89.2/ 16.0 & 89.6/ 29.2 \\
                       & ModelRe    & 92.1/ 10.5 & 92.2/ 10.5 & 92.4/ 10.4 & 92.1/ 10.5 & 87.6/ 9.7  & 89.8/ 10.4 & 91.8/ 10.6 & 92.2/ 10.6 & 89.0/ 10.3 & 89.8/ 8.7  \\
                       & PGD        & 91.8/ 17.5 & 91.8/ 17.4 & 91.3/ 13.0 & 92.2/ 14.2 & 87.9/ 11.3 & 89.9/ 11.8 & 91.8/ 21.5 & 91.7/ 17.4 & 89.3/ 12.4 & 89.6/ 24.7 \\
                       & Neurotoxin & 91.7/ 21.8 & 91.7/ 21.6 & 91.3/ 20.5 & 92.0/ 12.8 & 87.6/ 11.7 & 89.8/ 11.8 & 91.7/ 17.0 & 91.7/ 21.8 & 89.0/ 11.0 & 89.6/ 27.5 \\
                       & CerP       & 91.3/ 32.5 & 91.8/ 32.1 & 92.3/ 48.8 & 91.8/ 26.3 & 87.1/ 11.1 & 89.9/ 14.0 & 92.0/ 27.7 & 91.3/ 32.5 & 89.1/ 18.8 & 89.6/ 39.4 \\
                       & PFedBA     & 91.9/ \textbf{96.9} & 91.9/ \textbf{96.9} & 91.4/ \textbf{97.1} & 92.1/ \textbf{98.6} & 87.7/ \textbf{89.1} & 89.7/ \textbf{98.5} & 91.7/ \textbf{95.6} & 92.0/ \textbf{96.9} & 89.3/ \textbf{88.8} & 89.7/ \textbf{99.5} \\ \hline
\multirow{7}{*}{DnC}   & NoAttack   & 92.1/ 10.5 & 92.1/ 10.5 & 91.6/ 8.8  & 92.4/ 10.5 & 87.8/ 9.7  & 90.0/ 10.4 & 92.0/ 10.5 & 92.1/ 10.5 & 88.7/ 10.3 & 89.6/ 8.7  \\
                       & Sybil      & 92.0/ 26.8 & 91.9/ 26.8 & 91.5/ 39.5 & 92.2/ 17.7 & 87.9/ 11.2 & 90.0/ 11.6 & 91.7/ 20.9 & 92.0/ 26.6 & 88.9/ 20.6 & 89.4/ 33.6 \\
                       & ModelRe    & 92.1/ 10.7 & 91.9/ 10.9 & 92.4/ 10.7 & 92.2/ 10.7 & 87.8/ 10.0 & 89.8/ 10.4 & 91.9/ 10.9 & 92.0/ 10.6 & 89.2/ 10.5 & 89.7/ 9.2  \\
                       & PGD        & 92.0/ 26.2 & 91.8/ 26.7 & 91.5/ 21.9 & 92.1/ 13.8 & 88.2/ 12.9 & 90.0/ 13.8 & 91.8/ 33.4 & 91.8/ 26.5 & 89.0/ 17.0 & 89.7/ 30.6 \\
                       & Neurotoxin & 92.1/ 26.5 & 92.0/ 26.6 & 91.4/ 25.9 & 92.2/ 17.6 & 87.8/ 11.7 & 90.0/ 11.7 & 91.9/ 20.4 & 92.1/ 26.5 & 88.9/ 12.0 & 89.3/ 34.7 \\
                       & CerP       & 91.2/ 42.5 & 91.8/ 45.1 & 92.2/ 57.4 & 92.0/ 25.4 & 87.2/ 11.6 & 89.9/ 13.8 & 92.1/ 34.0 & 91.2/ 42.5 & 89.2/ 35.8 & 89.4/ 47.2 \\
                       & PFedBA     & 92.0/ \textbf{97.8} & 92.0/ \textbf{97.7} & 91.5/ \textbf{97.3} & 92.1/ \textbf{99.0} & 87.8/ \textbf{89.3} & 89.9/ \textbf{97.9} & 91.9/ \textbf{96.7} & 92.1/ \textbf{97.7} & 88.6/ \textbf{91.5} & 89.3/ \textbf{99.8} \\ \hline
\multirow{7}{*}{FLAME} & NoAttack   & 89.8/ 9.8  & 89.9/ 9.8  & 89.2/ 8.6  & 91.7/ 10.4 & 87.5/ 9.7  & 89.5/ 10.4 & 91.6/ 10.6 & 89.7/ 9.7  & 88.2/ 10.5 & 89.4/ 8.6  \\
                       & Sybil      & 90.1/ 9.8  & 89.9/ 9.8  & 89.2/ 8.7  & 91.6/ 10.4 & 87.5/ 9.7  & 89.5/ 10.4 & 91.6/ 10.6 & 89.9/ 9.8  & 88.6/ 22.9 & 89.0/ 18.3 \\
                       & ModelRe    & 89.8/ 9.8  & 89.9/ 9.8  & 90.5/ 10.0 & 91.7/ 10.5 & 87.5/ 9.7  & 89.5/ 10.4 & 91.6/ 10.6 & 89.4/ 9.7  & 87.8/ 10.5 & 89.4/ 8.6  \\
                       & PGD        & 90.1/ 9.9  & 89.9/ 9.9  & 90.1/ 8.7  & 91.6/ 10.4 & 87.5/ 9.7  & 89.5/ 10.4 & 91.3/ 10.6 & 89.7/ 10.0 & 89.2/ 16.2 & 89.4/ 8.9  \\
                       & Neurotoxin & 89.9/ 9.8  & 89.9/ 9.9  & 90.1/ 8.7  & 91.2/ 17.5 & 87.4/ 9.6  & 89.5/ 10.4 & 91.3/ 10.6 & 89.9/ 9.8  & 88.2/ 13.2 & 89.1/ 37.0 \\
                       & CerP       & 89.8/ 9.4 & 90.0/ 10.3 & 91.4/ 32.9 & 91.7/ 9.3 & 86.1/ 9.1 & 89.5/ 10.7 & 91.9/ 9.3 & 89.8/ 9.4 & 87.7/ 15.9 & 89.0/ 19.8 \\
                       &  PFedBA   & 89.7/ \textbf{96.0} & 89.8/ \textbf{96.2} & 89.2/ \textbf{92.1} & 91.5/ \textbf{99.4} & 87.2/ \textbf{73.8} & 89.4/ \textbf{98.8} & 91.5/ \textbf{95.9} & 89.6/ \textbf{95.8} & 88.7/ \textbf{87.0} & 89.3/ \textbf{99.7} \\ \hline
\end{tabular}
}
\end{table*}

%% file: table/table-iot.tex
\begin{table*}[]
\centering
\caption{The ACC/ ASR of various backdoor attacks against PFL with server-end defenses on N-BaIoT.(\%) {Standard Deviation Range: 0.2\%-3\%.}}
\label{tab:iot}
\resizebox{\linewidth}{!}{
\begin{tabular}{|c|c|c|c|c|c|c|c|c|c|c|c|}
\hline
Defenses               & Attacks    & FedAvg-FT   & FedProx-FT  & SCAFFOLD    & FO         & HF         & pFedMe      & Ditto       & FedBN       & FedRep      & FedALA      \\ \hline
\multirow{7}{*}{MKrum} & NoAttack   & 90.7/ 1.9   & 90.4/ 2.0   & 86.4/ 8.6   & 87.3/ 9.4  & 86.6/ 10.5 & 90.0/ 12.8  & 90.9/ 2.0   & 88.5/ 2.7   & 87.4/ 16.1  & 89.8/ 2.9   \\
                       & Sybil      & 90.3/ 10.4  & 90.3/ 14.8  & 86.3/ 67.7  & 87.6/ 9.2  & 86.3/ 10.4 & 90.0/ 13.8  & 90.7/ 25.5  & 88.6/ 6.7   & 87.3/ 24.2  & 89.9/ 4.5   \\
                       & ModelRe    & 90.7/ 1.9   & 90.4/ 2.0   & 86.5/ 8.7   & 87.3/ 9.4  & 86.6/ 10.5 & 90.0/ 13.4  & 90.9/ 2.0   & 88.5/ 2.7   & 87.5/ 16.0  & 89.3/ 3.2   \\
                       & PGD        & 90.4/ 3.0   & 90.4/ 3.3   & 86.4/ 12.6  & 87.6/ 9.2  & 85.6/ 10.4 & 90.0/ 13.2  & 90.9/ 2.1   & 88.5/ 3.9   & 87.1/ 21.7  & 90.0/ 2.8   \\
                       & Neurotoxin & 88.6/ 7.1   & 88.6/ 7.0   & 86.4/ 14.5  & 84.1/ 8.6  & 81.9/ 11.7 & 87.3/ 12.3  & 88.7/ 6.6   & 87.7/ 4.3   & 86.6/ 20.5  & 88.8/ 11.5  \\
                       & CerP & 90.3/ 80.8 & 90.4/ 85.2 & 89.5/ 67.4 & 87.6/ 9.2 & 85.8/ 10.1 & 90.0/ 15.7 & 90.8/ 83.6 & 90.3/ 80.8 & 91.0/ 17.9 & 91.9/ 55.0 \\
                       &  PFedBA   & 90.4/ \textbf{100} & 90.4/ \textbf{100} & 86.4/ \textbf{100} & 87.0/ \textbf{93.8} & 85.9/ \textbf{78.4} & 90.0/ \textbf{98.3}  & 90.9/ \textbf{100} & 88.5/ \textbf{100} & 87.6/ \textbf{100} & 89.8/ \textbf{100} \\ \hline
\multirow{7}{*}{Trim}  & NoAttack   & 90.2/ 2.8   & 90.3/ 2.5   & 86.4/ 8.5   & 87.6/ 10.1 & 85.8/ 11.5 & 89.7/ 12.2  & 90.7/ 2.8   & 88.3/ 2.3   & 87.7/ 14.9  & 89.7/ 1.6   \\
                       & Sybil      & 90.3/ 44.1  & 90.3/ 48.9  & 86.2/ 75.0  & 87.6/ 30.3 & 85.5/ 23.4 & 89.6/ 23.9  & 90.7/ 46.7  & 88.0/ 62.5  & 86.9/ 30.5  & 89.9/ 30.8  \\
                       & ModelRe    & 90.2/ 2.8   & 90.5/ 2.9   & 86.4/ 8.6   & 87.7/ 10.3 & 85.7/ 11.5 & 89.7/ 12.4  & 90.8/ 2.8   & 88.5/ 2.6   & 87.8/ 14.7  & 89.9/ 1.6   \\
                       & PGD        & 90.2/ 25.8  & 90.2/ 25.7  & 86.3/ 15.6  & 87.7/ 21.2 & 86.1/ 20.0 & 89.7/ 22.7  & 90.9/ 58.1  & 88.4/ 41.4  & 87.2/ 32.6  & 89.9/ 33.7  \\
                       & Neurotoxin & 88.3/ 44.0  & 88.0/ 44.5  & 86.3/ 16.6  & 83.7/ 21.0 & 82.0/ 20.2 & 87.1/ 18.4  & 88.3/ 43.4  & 88.3/ 44.0  & 87.0/ 23.9  & 88.8/ 31.7  \\
                       & CerP & 90.2/ 85.2 & 90.3/ 85.4 & 89.5/ 64.7 & 88.0/ 36.9 & 86.2/ 31.1 & 89.6/ 26.9 & 90.2/ 78.6 & 90.2/ 85.2 & 90.4/ 58.1 & 91.8/ 62.8 \\
                       &  PFedBA   & 90.3/ \textbf{100} & 90.2/ \textbf{100} & 86.4/ \textbf{100} & 86.8/ \textbf{89.2} & 86.2/ \textbf{74.1} & 89.6/ \textbf{99.6}  & 90.8/ \textbf{100} & 88.4/ \textbf{100} & 87.2/ \textbf{100} & 88.7/ \textbf{100} \\ \hline
\multirow{7}{*}{DnC}   & NoAttack   & 90.4/ 2.0   & 90.6/ 2.1   & 86.4/ 8.8   & 87.2/ 10.4 & 85.4/ 11.0 & 90.0/ 13.4  & 90.9/ 2.1   & 87.9/ 2.7   & 87.7/ 17.1  & 89.9/ 2.2   \\
                       & Sybil      & 90.6/ 80.8  & 90.6/ 82.0  & 86.2/ 79.5  & 86.5/ 25.2 & 85.9/ 22.3 & 90.0/ 23.6  & 90.8/ 84.8  & 88.4/ 77.4  & 86.1/ 34.4  & 89.8/ 27.9  \\
                       & ModelRe    & 90.3/ 2.1   & 90.6/ 2.1   & 86.5/ 8.6   & 87.2/ 10.3 & 85.4/ 11.1 & 90.0/ 13.4  & 90.9/ 2.1   & 88.6/ 2.7   & 87.5/ 15.4  & 89.8/ 22.7  \\
                       & PGD        & 90.3/ 25.0  & 90.3/ 26.1  & 86.3/ 17.3  & 87.4/ 25.7 & 85.9/ 29.9 & 90.0/ 25.6  & 90.8/ 94.4  & 88.4/ 67.8  & 87.6/ 69.7  & 89.7/ 84.9  \\
                       & Neurotoxin & 88.5/ 76.3  & 88.2/ 79.1  & 86.4/ 19.1  & 84.0/ 17.5 & 81.3/ 16.8 & 87.2/ 25.5  & 88.7/ 76.1  & 88.5/ 76.3  & 87.0/ 27.9  & 89.7/ 46.6  \\
                       & CerP & 90.2/ 92.9 & 90.3/ 91.6 & 90.2/ 98.6 & 87.7/ 33.3 & 86.3/ 28.3 & 90.0/ 24.8 & 91.0/ 94.8 & 90.2/ 92.9 & 89.2/ 26.3 & 91.1/ 82.6 \\
                       &  PFedBA   & 90.6/ \textbf{99.1}  & 90.3/ \textbf{100} & 86.4/ \textbf{100} & 86.6/ \textbf{93.6} & 84.8/ \textbf{82.2} & 90.0/ \textbf{100} & 90.9/ \textbf{100} & 88.6/ \textbf{100} & 87.7/ \textbf{100} & 89.5/ \textbf{100} \\ \hline
\multirow{7}{*}{FLAME} & NoAttack   & 87.7/ 4.9   & 87.6/ 4.6   & 86.4/ 10.3  & 84.3/ 9.1  & 81.1/ 12.4 & 86.8/ 6.0   & 88.4/ 4.5   & 87.6/ 4.8   & 85.7/ 18.0  & 90.0/ 3.3   \\
                       & Sybil      & 87.7/ 4.9   & 87.2/ 4.7   & 86.3/ 59.9  & 84.3/ 9.1  & 81.5/ 12.7 & 87.3/ 6.7   & 88.4/ 4.5   & 87.7/ 4.9   & 86.6/ 20.5  & 90.1/ 5.1   \\
                       & ModelRe    & 87.7/ 4.9   & 87.7/ 4.6   & 86.4/ 10.9  & 84.3/ 9.1  & 81.5/ 12.7 & 87.3/ 6.7   & 88.4/ 4.5   & 87.6/ 4.8   & 87.0/ 18.7  & 90.1/ 3.4   \\
                       & PGD        & 87.7/ 4.9   & 87.1/ 6.2   & 86.5/ 10.5  & 84.3/ 9.1  & 81.5/ 12.7 & 87.3/ 6.7   & 88.4/ 4.5   & 87.7/ 4.9   & 85.9/ 20.2  & 90.1/ 3.6   \\
                       & Neurotoxin & 85.8/ 4.2   & 87.2/ 4.8   & 86.5/ 10.5  & 84.3/ 9.1  & 81.5/ 12.7 & 87.3/ 6.7   & 88.4/ 4.5   & 87.7/ 4.3   & 86.6/ 20.5  & 90.1/ 3.8   \\
                       & CerP & 90.5/ 8.2 & 89.8/ 11.7 & 89.4/ 54.9 & 86.6/ 10.4 & 83.7/ 12.3 & 89.0/ 14.9 & 90.2/ 11.1 & 90.5/ 8.2 & 89.0/ 16.3 & 91.8/ 67.3 \\
                       &  PFedBA   & 87.4/ \textbf{98.9}  & 87.3/ \textbf{98.9}  & 86.4/ \textbf{100} & 83.8/ \textbf{96.4} & 81.4/ \textbf{79.3} & 86.8/\textbf{ 99.1}  & 88.3/ \textbf{100} & 86.8/ \textbf{98.9}  & 86.7/ \textbf{98.6}  & 90.0/ \textbf{100} \\ \hline
\end{tabular}
}
\end{table*}

%% file: table/table-cifar10.tex
\begin{table*}[]
\caption{The ACC/ ASR of various backdoor attacks against PFL with server-end defenses on CIFAR-10.(\%) {Standard Deviation Range: 0.5\%-5\%.}}
\label{tab:cifar}
\resizebox{\linewidth}{!}{
\begin{tabular}{|c|c|c|c|c|c|c|c|c|c|c|c|}
\hline
Defenses               & Attacks    & FedAvg-FT  & FedProx-FT & SCAFFOLD   & FO          & HF         & pFedMe     & Ditto       & FedBN      & FedRep     & FedALA                               \\ \hline
\multirow{7}{*}{MKrum} & NoAttack   & 89.5/ 7.0  & 90.2/ 7.4  & 92.6/ 7.3  & 86.7/ 7.9   & 83.3/ 5.9  & 86.1/ 7.9  & 91.2/ 7.7   & 89.5/ 7.0  & 91.8/ 6.5  & 81.0/ 8.2                            \\
                       & Sybil      & 89.7/ 34.5 & 90.2/ 15.9 & 92.6/ 33.4 & 86.6/ 71.2  & 82.7/ 37.1 & 85.4/ 65.7 & 91.1/ 25.1  & 89.7/ 34.5 & 91.8/ 11.2 & 80.6/ 83.4                           \\
                       & ModelRe    & 90.1/ 7.1  & 90.2/ 7.4  & 92.5/ 8.6  & 86.6/ 7.6   & 83.8/ 5.9  & 86.9/ 7.0  & 91.1/ 7.6   & 90.1/ 7.1  & 91.8/ 6.5  & 81.1/ 6.3                            \\
                       & PGD        & 90.0/ 40.6 & 90.2/ 38.7 & 92.5/ 40.2 & 86.5/ 66.2  & 82.2/ 24.8 & 85.3/ 74.6 & 91.0/ 61.9  & 90.0/ 40.6 & 91.6/ 44.8 & 80.6/ 84.1                           \\
                       & Neurotoxin & 89.7/ 24.0 & 90.3/ 25.1 & 92.1/ 25.1 & 86.9/ 81.5  & 79.3/ 62.7 & 86.0/ 71.8 & 91.3/ 30.0  & 89.6/ 10.3 & 92.0/ 28.6 & 79.3/ 32.3                           \\
                       & CerP & 89.6/ 34.9 & 89.9/ 29.2 & 92.2/ 41.1 & 86.8/ 89.7 & 79.7/ 49.2 & 85.8/ 89.4 & 91.2/ 20.5 & 89.6/ 34.9 & 92.0/ 39.1 & 82.8/ 22.1 \\
                       &  PFedBA   & 90.0/ \textbf{98.0} & 89.9/ \textbf{99.5} & 92.6/ \textbf{99.6} & 86.7/ \textbf{100} & 82.6/ \textbf{91.1} & 84.6/ \textbf{96.3} & 91.1/ \textbf{98.2}  & 90.0/ \textbf{98.0} & 92.0/ \textbf{99.2} & 82.2/ \textbf{95.9}                           \\ \hline
\multirow{7}{*}{Trim}  & NoAttack   & 90.5/ 6.9  & 90.6/ 7.4  & 92.0/ 8.2  & 87.2/ 7.0   & 81.7/ 6.7  & 86.8/ 7.2  & 91.3/ 7.2   & 90.5/ 6.9  & 92.2/ 7.1  & 85.6/ 7.7                            \\
                       & Sybil      & 90.8/ 20.5 & 90.3/ 20.5 & 92.6/ 28.0 & 86.9/ 81.4  & 81.1/ 35.2 & 87.7/ 71.5 & 91.3/ 16.8  & 90.8/ 20.5 & 92.2/ 15.4 & 84.5/ 51.7                           \\
                       & ModelRe    & 90.7/ 7.1  & 90.4/ 8.0  & 92.7/ 8.1  & 87.2/ 7.1   & 81.5/ 7.1  & 87.2/ 7.3  & 91.3/ 7.3   & 90.7/ 7.1  & 92.2/ 7.6  & 85.5/ 8.6                            \\
                       & PGD        & 90.4/ 38.7 & 90.7/ 36.6 & 92.6/ 41.9 & 86.9/ 72.8  & 80.5/ 30.6 & 86.4/ 71.6 & 91.1/ 54.7  & 90.4/ 38.7 & 92.1/ 41.0 & 82.7/ 79.4                           \\
                       & Neurotoxin & 90.6/ 21.6 & 90.5/ 14.6 & 92.4/ 30.0 & 87.5/ 82.7  & 81.0/ 71.4 & 86.2/ 71.7 & 91.4/ 25.6  & 90.9/ 15.4 & 92.0/ 22.1 & 84.6/ 49.8                           \\
                      & CerP & 90.6/ 28.5 & 90.7/ 28.0 & 92.6/ 44.3 & 87.1/ 88.7 & 81.3/ 31.4 & 86.6/ 82.9 & 91.5/ 33.8 & 90.6/ 28.5 & 92.1/ 28.5 & 83.9/ 69.1 \\
                       &  PFedBA   & 90.5/ \textbf{94.3} & 90.2/ \textbf{95.3} & 91.8/ \textbf{98.8} & 87.2/ \textbf{99.9}  & 81.3/ \textbf{75.4} & 87.5/ \textbf{97.2} & 91.1/ \textbf{99.1}  & 90.5/ \textbf{94.3} & 92.1/ \textbf{97.9} & 83.5/ \textbf{99.5} \\ \hline
\multirow{7}{*}{DnC}   & NoAttack   & 90.2/ 6.6  & 89.9/ 7.0  & 91.8/ 6.6  & 86.9/ 7.2   & 80.7/ 6.8  & 86.2/ 7.3  & 91.3/ 6.6   & 90.2/ 6.6  & 92.1/ 6.3  & 85.5/ 5.3                            \\
                       & Sybil      & 90.2/ 30.4 & 89.8/ 28.9 & 91.9/ 28.7 & 86.7/ 86.9  & 80.1/ 83.6 & 85.5/ 72.9 & 91.0/ 41.1  & 90.2/ 30.4 & 92.0/ 35.2 & 82.1/ 90.8                           \\
                       & ModelRe    & 78.1/ 54.0 & 78.3/ 48.8 & 89.1/ 58.3 & 86.4/ 22.6  & 75.2/ 9.8  & 86.0/ 7.8  & 80.6/ 69.7  & 78.1/ 54.0 & 92.0/ 12.5 & 85.1/ 8.7                            \\
                       & PGD        & 90.3/ 46.2 & 90.1/ 33.2 & 92.0/ 47.5 & 86.4/ 65.5  & 79.3/ 20.1 & 86.1/ 72.2 & 91.2/ 55.5  & 90.3/ 46.2 & 92.0/ 49.7 & 81.4/ 97.1                           \\
                       & Neurotoxin & 90.0/ 29.1 & 89.5/ 35.5 & 92.0/ 17.9 & 86.4/ 83.6  & 79.5/ 69.1 & 84.6/ 79.1 & 91.2/ 55.5  & 90.0/ 29.2 & 92.1/ 41.8 & 82.5/ 93.1                           \\
                       & CerP      & 90.4/ 46.9 & 90.1/ 42.8 & 91.7/ 25.8 & 87.1/ 95.6 & 79.9/ 83.8 & 84.7/ 91.7 & 91.4/ 61.8 & 90.4/ 46.9 & 92.1/ 45.7 & 82.6/ 96.0 \\
                       &  PFedBA   & 90.7/ \textbf{98.0} & 90.0/ \textbf{97.8} & 91.5/ \textbf{99.0} & 86.8/ \textbf{100} & 79.7/ \textbf{85.3} & 85.7/ \textbf{98.1} & 91.2/ \textbf{99.3}  & 90.7/ \textbf{98.0} & 92.3/ \textbf{99.3} & 83.4/ \textbf{99.8}                           \\ \hline
\multirow{7}{*}{FLAME} & NoAttack   & 82.6/ 6.9  & 82.4/ 7.7  & 90.6/ 6.8  & 82.0/ 6.7   & 80.4/ 6.4  & 78.3/ 6.7  & 87.2/ 6.7   & 82.6/ 6.9  & 85.9/ 7.0  & 68.0/ 4.8                            \\
                       & Sybil      & 83.3/ 64.5 & 82.9/ 51.8 & 90.3/ 88.5 & 82.2/ 36.5  & 79.8/ 41.5 & 77.7/ 27.3 & 86.8/ 87.2  & 83.3/ 64.5 & 85.6/ 25.7 & 67.7/ 71.8                           \\
                       & ModelRe    & 83.0/ 7.2  & 82.6/ 7.4  & 90.6/ 6.7  & 81.7/ 6.9   & 80.3/ 6.5  & 77.9/ 6.9  & 87.3/ 6.3   & 83.0/ 7.2  & 86.0/ 7.1  & 70.6/ 4.5                            \\
                       & PGD        & 81.9/ 66.9 & 81.6/ 60.6 & 90.3/ 69.2 & 81.4/ 70.2  & 79.4/ 36.8 & 77.6/ 19.3 & 86.4/ 97.4  & 81.9/ 66.9 & 85.8/ 38.4 & 66.7/ 78.2                           \\
                       & Neurotoxin & 82.8/ 57.1 & 83.0/ 51.7 & 90.4/ 47.9 & 81.6/ 84.1  & 79.8/ 70.6 & 77.4/ 52.7 & 86.7/ 87.9  & 82.8/ 57.1 & 86.0/ 17.2 & 67.1/ 70.1                           \\
                       & CerP & 83.1/ 67.6 & 83.1/ 70.5 & 90.6/ 81.4 & 81.8/ 60.6 & 78.0/ 18.8 & 79.4/ 39.3 & 87.0/ 94.6 & 83.1/ 67.6 & 86.0/ 20.3 & 68.5/ 78.0 \\
                       &  PFedBA   & 82.1/ \textbf{94.0} & 82.7/ \textbf{94.6} & 90.8/ \textbf{99.9} & 81.8/ \textbf{99.7}  & 77.7/ \textbf{90.6} & 78.5/ \textbf{94.0} & 86.6/ \textbf{100} & 82.1/ \textbf{94.0} & 86.2/ \textbf{97.9} & 69.2/ \textbf{96.4}                           \\ \hline
\end{tabular}
}
\end{table*}

%% file: table/table-client-defense.tex
\begin{table*}[]
\centering
\caption{The ACC/ ASR of backdoor attacks against NC and NAD on Fashion-MNIST and N-BaIoT.(\%) {Standard Deviation Range: 0.5\%-4.5\%.}}
\label{tab:client-defense}
\resizebox{\linewidth}{!}{
\begin{tabular}{|c|c|c|c|c|c|c|c|c|c|c|c|c|}
\hline
Dataset                                                                   & Defense              & Attack     & FedAvg-FT  & FedProx-FT & SCAFFOLD   & FO         & HF         & pFedMe     & Ditto      & FedBN      & FedRep     & FedALA     \\ \hline
\multirow{14}{*}{\begin{tabular}[c]{@{}c@{}}Fashion\\ MNIST\end{tabular}} & \multirow{6}{*}{NC}  & No Attack  & 89.8/ 10.2 & 89.6/ 10.2 & 90.3/ 8.8  & 90.5/ 10.5 & 86.7/ 10.2 & 86.2/ 10.3 & 89.7/10.2  & 89.8/ 10.3 & 88.9/ 10.1 & 89.7/ 8.8  \\
                                                                          &                      & Sybil      & 90.1/ 20.6 & 90.0/ 20.5 & 90.4/ 23.2 & 90.8/ 20.0 & 86.5/ 11.9 & 86.4/ 12.0 & 89.8/ 19.3 & 90.1/ 20.4 & 89.2/ 30.6 & 89.8/ 21.9 \\
                                                                          &                      & ModelRe    & 89.9/ 10.6 & 89.7/ 10.5 & 89.8/ 10.6 & 90.8/ 10.7 & 86.5/ 10.2 & 86.4/ 10.4 & 89.9/ 10.5 & 90.0/ 10.7 & 89.1/ 10.7 & 89.9/ 9.3  \\
                                                                          &                      & PGD        & 89.8/ 12.1 & 89.6/ 12.2 & 90.3/ 12.6 & 90.8/ 13.2 & 86.5/ 12.9 & 86.7/ 12.1 & 89.9/ 29.6 & 89.8/ 12.3 & 89.3/ 31.2 & 89.8/ 27.8 \\
                                                                          &                      & Neurotoxin & 89.8/ 20.3 & 89.9/ 20.0 & 90.1/ 22.5 & 90.7/ 11.8 & 86.5/ 12.2 & 85.6/ 11.2 & 89.8/ 19.6 & 89.8/ 20.3 & 88.8/ 12.5 & 89.9/ 21.3 \\
                                                                          &                      & CerP & 89.3/ 19.4 & 89.2/ 18.6 & 88.3/ 19.4 & 90.6/ 17.0 & 86.5/ 11.9 & 85.7/ 12.2 & 89.0/ 19.2 & 83.4/ 19.4 & 89.5/ 18.7 & 88.8/ 19.3 \\
                                                                          &                      &  PFedBA   & 89.8/ \textbf{77.0} & 89.9/ \textbf{76.3} & 90.3/ \textbf{77.2} & 91.0/ \textbf{79.1} & 86.5/ \textbf{70.6} & 86.3/ \textbf{72.4} & 89.5/ \textbf{74.7} & 89.6/ \textbf{74.5} & 89.0/ \textbf{77.3} & 90.0/ \textbf{75.7} \\ \cline{2-13} 
                                                                          & \multirow{7}{*}{NAD} & No Attack  & 92.4/ 10.5 & 92.5/ 10.5 & 92.4/ 9.0  & 90.1/ 10.3 & 85.6/ 9.7  & 85.2/ 9.8  & 92.4/ 10.5 & 87.7/ 10.4 & 88.5/ 9.9 & 92.5/ 8.9  \\
                                                                          &                      & Sybil      & 90.9/ 26.9 & 91.5/ 46.3 & 91.2/ 47.1 & 88.2/ 17.1 & 82.5/ 10.6 & 85.3/ 13.9 & 90.9/ 26.9 & 91.5/ 39.7 & 89.0/ 39.7 & 92.1/ 45.8 \\
                                                                          &                      & ModelRe    & 92.5/ 10.9 & 92.4/ 10.9 & 92.3/ 9.2  & 90.1/ 10.5 & 85.7/ 9.7  & 85.2/ 9.9  & 92.5/ 10.9 & 90.8/ 10.7 & 89.5/ 10.6 & 91.9/ 46.9 \\
                                                                          &                      & PGD        & 92.0/ 22.4 & 92.0/ 22.4 & 91.2/ 21.9 & 89.7/ 16.2 & 85.9/ 15.7 & 85.3/ 19.0 & 92.2/ 47.3 & 88.7/ 22.0 & 88.0/ 35.6 & 91.5/ 49.1 \\
                                                                          &                      & Neurotoxin & 90.9/ 40.0 & 86.5/ 36.5 & 91.8/ 42.2 & 87.4/ 14.1 & 82.5/ 14.3 & 83.6/ 11.9 & 89.7/ 38.6 & 90.8/ 39.0 & 88.6/ 19.3 & 92.4/ 44.7 \\
                                                                          &                       & CerP      & 92.4/ 55.6  & 92.1/ 53.9 & 92.5/ 60.6 & 90.4/ 28.9 & 85.1/ 15.1 & 84.8/ 13.5 & 92.3/ 55.4 & 92.4/ 55.6 & 90.0/ 38.7 & 92.4/ 54.7 \\
                                                                          &                      &  PFedBA   & 92.1/ \textbf{88.7} & 92.0/ \textbf{89.1} & 92.3/ \textbf{92.4} & 90.1/ \textbf{88.1} & 85.7/ \textbf{76.7} & 85.0/ \textbf{72.5} & 92.1/ \textbf{88.4} & 88.2/ \textbf{79.2} & 88.2/ \textbf{85.4} & 92.1/ \textbf{92.2} \\ \hline
\multirow{14}{*}{N-BaIoT}                                                 & \multirow{7}{*}{NC}  & No Attack  & 90.5/ 8.1  & 90.5/ 7.6  & 87.5/ 5.6  & 90.6/ 11.2 & 90.1/11.7  & 87.1/ 14.3 & 90.7/ 7.4  & 87.5/ 8.5  & 89.2/ 12.2 & 87.8/ 7.0  \\
                                                                          &                      & Sybil      & 90.5/ 18.2 & 90.5/ 19.3 & 87.5/ 24.0 & 90.4/ 21.1 & 90.1/ 22.7 & 87.5/ 13.8 & 90.4/ 19.0 & 87.4/ 22.6 & 88.7/ 26.1 & 87.3/ 22.8 \\
                                                                          &                      & ModelRe    & 90.4/ 12.9 & 90.5/ 13.8 & 87.5/ 24.6 & 90.5/ 15.0 & 90.1/ 15.1 & 87.2/ 14.5 & 90.5/ 13.1 & 87.7/ 16.3 & 88.8/ 17.0 & 87.5/ 16.6 \\
                                                                          &                      & PGD        & 90.6/ 17.7 & 90.4/ 17.2 & 87.5/ 18.8 & 90.4/ 22.4 & 90.0/ 21.2 & 87.7/ 20.7 & 90.0/ 22.1 & 87.4/ 22.9 & 88.6/ 32.1 & 87.1/ 24.5 \\
                                                                          &                      & Neurotoxin & 87.4/ 21.1 & 87.7/ 21.1 & 87.9/ 22.9 & 88.2/ 17.3 & 87.5/ 16.9 & 84.3/ 15.6 & 87.2/ 23.6 & 87.5/ 22.4 & 88.8/ 20.4 & 87.2/ 25.2 \\
                                                                          &                      & CerP        & 90.2/ 33.6 & 90.4/ 33.1 & 90.8/ 39.9 & 90.4/ 28.7 & 90.1/ 23.4 & 87.6/ 13.4 & 90.6/ 33.6 & 90.3/ 33.6 & 91.7/ 33.0 & 90.4/ 23.4 \\
                                                                          &                      &  PFedBA   & 90.6/ \textbf{90.8} & 90.6/ \textbf{91.6} & 87.7/ \textbf{86.9} & 90.5/ \textbf{76.8} & 90.0/ \textbf{71.8} & 87.1/ \textbf{70.2} & 89.9/ \textbf{90.1} & 87.2/ \textbf{89.2} & 89.1/ \textbf{94.6} & 87.7/ \textbf{94.2} \\ \cline{2-13} 
                                                                          & \multirow{7}{*}{NAD} & No Attack  & 90.3/ 8.7  & 90.6/ 9.0  & 88.9/ 5.6  & 89.2/ 13.1 & 89.5/ 15.9 & 87.6/ 14.6 & 90.3/ 8.7  & 89.0/ 8.0  & 89.1/ 12.3 & 88.9/ 7.7  \\
                                                                          &                      & Sybil      & 90.3/ 59.6 & 90.3/ 59.4 & 89.3/ 70.0 & 89.7/ 46.1 & 89.6/ 30.6 & 87.4/ 25.1 & 90.3/ 59.6 & 88.5/ 55.5 & 89.2/ 45.1 & 88.6/ 52.2 \\
                                                                          &                      & ModelRe    & 90.0/ 39.7 & 90.1/ 43.0 & 89.3/ 21.0 & 89.8/ 24.0 & 89.2/ 25.9 & 87.8/ 16.3 & 90.0/ 39.7 & 88.6/ 32.8 & 89.0/ 32.6 & 88.9/ 24.4 \\
                                                                          &                      & PGD        & 90.3/ 49.3 & 89.7/ 50.1 & 89.0/ 53.3 & 89.9/ 49.4 & 90.0/ 41.0 & 87.7/ 39.9 & 90.4/ 89.6 & 88.7/ 51.4 & 88.9/ 71.2 & 88.8/ 87.1 \\
                                                                          &                      & Neurotoxin & 89.0/ 59.8 & 88.9/ 58.1 & 89.0/ 61.4 & 87.8/ 33.8 & 87.7/ 21.5 & 85.8/ 26.1 & 89.0/ 59.8 & 89.0/ 59.8 & 89.0/ 27.0 & 88.6/ 55.5 \\
                                                                          &                      & CerP      & 90.4/ 51.3 & 90.4/ 55.3 & 90.6/ 62.0 & 90.1/ 47.5 & 89.9/ 35.4 & 87.8/ 17.6 & 90.4/ 55.4 & 90.4/ 51.3 & 91.7/ 60.1 & 90.2/ 55.2 \\
                                                                          &                      &  PFedBA   & 90.3/ \textbf{96.3} & 90.2/ \textbf{95.4} & 89.1/ \textbf{98.6} & 89.3/ \textbf{81.6} & 89.5/ \textbf{73.2} & 87.4/ \textbf{71.4} & 90.3/ \textbf{95.1} & 88.6/ \textbf{96.8} & 89.0/ \textbf{93.7} & 88.3/ \textbf{98.3} \\ \hline
\end{tabular}
}
\end{table*}

%% file: table/table-cifar100.tex
\begin{table*}[]
\centering
\caption{The ACC/ ASR of various backdoor attacks against PFL with server-end defenses on CIFAR-100.(\%) {Standard Deviation Range: 0.6\%-4.6\%.}}
\label{tab:cifar100}
\resizebox{\linewidth}{!}{
\begin{tabular}{|c|c|c|c|c|c|c|c|c|c|c|c|}
\hline
Defenses               & Attacks    & FedAvg-FT  & FedProx-FT & SCAFFOLD    & FO          & HF          & pFedMe     & Ditto      & FedBN      & FedRep      & FedALA      \\ \hline
\multirow{7}{*}{MKrum} & NoAttack   & 74.1/ 1.2  & 75.1/ 1.1  & 78.0/ 1.4   & 63.8/ 1.3   & 65.7/ 3.0   & 62.6/ 1.5  & 76.4/ 1.3  & 74.1/ 1.2  & 76.4/ 1.5   & 75.4/ 1.3   \\
                       & Sybil      & 74.5/ 62.0 & 75.7/ 9.7  & 77.5/ 74.5  & 63.6/ 62.3  & 64.2/ 93.4  & 61.9/ 51.4 & 76.3/ 70.4 & 74.5/ 62.0 & 76.2/ 41.4  & 75.5/ 26.5  \\
                       & ModelRe    & 74.3/ 1.2  & 75.1/ 1.1  & 78.0/ 1.1   & 63.8/ 1.4   & 65.5/ 2.8   & 62.6/ 1.5  & 76.6/ 1.2  & 74.3/ 1.2  & 76.4/ 1.5   & 74.9/ 1.7   \\
                       & PGD        & 73.2/ 84.7 & 73.9/ 86.3 & 77.5/ 80.2  & 64.0/ 80.0  & 63.0/ 56.7  & 63.1/ 69.5 & 75.5/ 98.9 & 73.2/ 84.7 & 76.3/ 91.9  & 71.9/ 95.0  \\
                       & Neurotoxin & 74.5/ 39.4 & 74.7/ 31.0 & 78.5/ 78.8  & 64.6/ 72.2  & 64.3/ 69.1  & 61.7/ 63.7 & 75.8/ 43.4 & 74.5/ 39.4 & 76.1/ 35.1  & 75.0/ 45.5  \\
                       & CerP & 75.5/ 13.5 & 75.4/ 28.6 & 77.7/ 91.1 & 64.6/ 87.5 & 66.7/ 98.1 & 61.7/ 67.2 & 76.4/ 39.4 & 75.5/ 13.5 & 76.1/ 46.4 & 75.0/ 28.4 \\
                       &  PFedBA   & 74.1/\textbf{ 99.3 }& 75.0/ \textbf{99.8} & 77.9/ \textbf{99.9}  & 64.2/ \textbf{99.9}  & 64.8/\textbf{ 99.5}  & 62.9/\textbf{ 97.5} & 76.4/\textbf{ 99.9} & 74.1/ \textbf{99.3 }& 76.4/ \textbf{99.8 } & 74.6/\textbf{ 100 }\\ \hline
\multirow{7}{*}{Trim}  & NoAttack   & 76.8/ 1.0  & 76.4/ 0.8  & 78.6/ 1.2   & 63.9/ 1.4   & 65.5/ 0.9   & 62.3/ 1.5  & 77.8/ 1.0  & 76.8/ 1.0  & 78.5/ 1.6   & 71.6/ 0.8   \\
                       & Sybil      & 76.9/ 61.3 & 77.0/ 61.2 & 77.8/ 84.1  & 63.9/ 61.7  & 64.2/ \textbf{81.9}  & 62.8/ 57.1 & 77.7/ 64.5 & 76.9/ 61.3 & 78.3/ 72.4  & 72.1/ 64.0  \\
                       & ModelRe    & 77.0/ 1.0  & 77.0/ 1.0  & 78.5/ 1.5   & 64.0/ 1.4   & 64.9/ 1.4   & 62.8/ 1.7  & 77.9/ 1.1  & 77.0/ 1.0  & 78.4/ 1.9   & 71.9/ 0.9   \\
                       & PGD        & 76.1/ 85.8 & 76.1/ 87.2 & 78.1/ 86.9  & 63.7/ 78.1  & 63.1/ 62.2  & 64.2/ 65.7 & 77.2/ 98.1 & 76.1/ 85.8 & 77.9/ 92.3  & 69.0/ 96.5  \\
                       & Neurotoxin & 77.2/ 64.9 & 76.9/ 56.1 & 78.5/ 74.7  & 64.3/ 63.7  & 64.0/ 65.4  & 62.7/ 57.5 & 78.0/ 68.5 & 77.2/ 64.9 & 77.7/ 69.6  & 72.7/ 64.7  \\
                        & CerP & 77.3/ 75.1 & 76.9/ 71.8 & 78.1/ 87.5 & 63.8/ 65.9 & 66.1/ 80.4 & 62.6/ 60.2 & 77.4/ 79.5 & 77.3/ 62.0 & 78.2/ 77.8 & 72.4/ 65.7 \\
                       &  PFedBA   & 76.9/ \textbf{99.6 }& 76.7/ \textbf{99.7} & 78.2/ \textbf{100 }& 63.9/ \textbf{100} & 64.1/ 76.5  & 63.7/ \textbf{98.7} & 77.8/ \textbf{99.9 }& 76.9/\textbf{ 99.6} & 78.2/ \textbf{100} & 72.1/ \textbf{100} \\ \hline
\multirow{7}{*}{DnC}   & NoAttack   & 76.5/ 1.0  & 76.0/ 1.6  & 78.7/ 1.1   & 64.7/ 1.4   & 66.0/ 1.9   & 63.7/ 0.6  & 78.1/ 1.0  & 76.5/ 1.0  & 77.5/ 1.7   & 77.3/ 1.1   \\
                       & Sybil      & 75.9/ 82.1 & 75.3/ 83.7 & 78.0/ 76.4  & 64.3/ 91.5  & 64.9/ 92.3  & 61.5/ 67.6 & 77.3/ 88.5 & 75.9/ 82.1 & 76.9/ 85.2  & 76.1/ 92.4  \\
                       & ModelRe    & 69.3/ 24.0 & 68.7/ 24.9 & 78.2/ 12.0  & 64.8/ 24.0  & 66.3/ 6.9   & 63.3/ 0.5  & 73.8/ 27.9 & 69.3/ 24.0 & 77.4/ 3.4   & 42.4/ 1.1   \\
                       & PGD        & 76.1/ 77.6 & 75.7/ 78.5 & 78.0/ 81.6  & 63.8/ 83.1  & 58.1/ 33.9  & 63.1/ 77.8 & 76.5/ 97.5 & 76.1/ 77.6 & 77.1/ 81.2  & 70.5/ 98.3  \\
                       & Neurotoxin & 75.4/ 78.0 & 75.6/ 79.4 & 77.8/ 73.5  & 64.8/ 83.6  & 63.0/ 57.7  & 62.2/ 68.9 & 77.3/ 86.2 & 75.4/ 78.0 & 77.2/ 80.8  & 75.8/ 93.6  \\
                       & CerP & 74.8/ 82.4 & 75.7/ 85.9 & 77.6/ 80.5 & 64.5/ 94.4 & 63.8/ 97.1 & 62.5/ 64.3 & 77.2/ 92.0 & 75.5/ 85.0 & 77.5/ 86.3 & 76.0/ 96.5 \\
                       &  PFedBA   & 76.1/ \textbf{99.8} & 75.4/ \textbf{99.9} & 77.6/ \textbf{100} & 64.4/ \textbf{100} & 65.0/ \textbf{100} & 63.1/ \textbf{99.4} & 77.6/ \textbf{99.9} & 76.1/ \textbf{99.8} & 77.1/ \textbf{100} & 76.2/ \textbf{99.9}  \\ \hline
\multirow{7}{*}{FLAME} & NoAttack   & 61.5/ 0.9  & 59.5/ 1.2  & 61.5/ 1.6   & 54.1/ 1.5   & 56.7/ 1.4   & 53.0/ 1.3  & 66.6/ 1.9  & 59.0/ 1.3  & 62.1/ 0.7   & 50.4/ 0.9   \\
                       & Sybil      & 58.2/ 66.0 & 58.9/ 73.4 & 60.3/ 79.7  & 54.2/ 83.4  & 55.6/ 89.8  & 49.7/ 60.9 & 65.8/ 89.6 & 58.3/ 72.5 & 62.6/ 78.1  & 49.6/ 76.6  \\
                       & ModelRe    & 61.1/ 1.3  & 57.1/ 0.9  & 78.9/ 2.6   & 54.6/ 1.9   & 57.5/ 1.0   & 51.0/ 0.8  & 66.3/ 1.7  & 61.1/ 0.8  & 62.5/ 1.5   & 47.5/ 1.3   \\
                       & PGD        & 57.8/ 72.8 & 57.7/ 70.9 & 75.7/ 90.3  & 52.8/ 77.3  & 55.0/ 56.6  & 47.8/ 24.7 & 65.4/ 92.9 & 60.0/ 73.5 & 61.9/ 85.3  & 41.1/ 85.1  \\
                       & Neurotoxin & 58.8/ 70.9 & 59.3/ 69.5 & 59.0/ 53.7  & 53.3/ 74.0  & 55.2/ 66.5  & 52.0/ 62.8 & 65.9/ 90.2 & 58.5/ 75.0 & 62.9/ 79.1  & 48.5/ 59.7  \\
                       & CerP & 57.6/ 53.1 & 57.8/ 72.7 & 76.1/ 93.3 & 54.2/ 83.3 & 56.8/ 39.3 & 50.9/ 63.7 & 65.9/ 94.0 & 57.3/ 62.4 & 65.3/ 88.5 & 43.4/ 68.8 \\
                       &  PFedBA   & 59.7/ \textbf{96.5} & 59.0/ \textbf{95.3} & 71.3/ \textbf{99.9}  & 54.4/ \textbf{99.8}  & 56.3/ \textbf{93.2}  & 52.4/ \textbf{98.1} & 67.6/ \textbf{99.9} & 58.5/ \textbf{96.6} & 63.1/ \textbf{99.3}  & 48.9/ \textbf{88.2}  \\ \hline
\end{tabular}
}
\end{table*}

%% file: table/paramters.tex
\begin{table}[t]
\centering
\caption{{The parameters of PFL systems.}}
\label{tab:parameters}
\resizebox{\linewidth}{!}{
\begin{tabular}{|c|c|c|c|c|c|c|}
\hline
Datasets      & $N$   & $N_p$ & \multicolumn{1}{l|}{Malicious clients} & Batch Size & FL Training Iteration & Dirichlet $\alpha$ \\ \hline
FMNIST        & 100 & 10   & 10                                     & 64         & 100                       & 0.1               \\ \hline
CIFAR-10      & 100 & 10   & 10                                     & 64         & 200                       & 0.5            \\ \hline
CIFAR-100     & 100 & 10   & 10                                     & 64         & 300                       & 0.5              \\ \hline
N-BaIoT       & 100 & 10   &10                                      & 128        & 150                       & 0.1           \\ \hline
\end{tabular}
}
\end{table}

%% file: table/table-fltrust.tex
\begin{table*}[]
\centering
\caption{{The ACC/ ASR of backdoor attacks against FLTrust on Fashion-MNIST, CIFAR-10, and N-BaIoT.(\%)}}
\label{tab:fltrust}
\resizebox{\linewidth}{!}{
\begin{tabular}{|c|c|c|c|c|c|c|c|c|c|c|}
 \hline
Dataset                                                 & FedAvg-FT  & FedProx-FT & SCAFFOLD   & FO         & HF         & pFedMe     & Ditto       & FedBN      & FedRep     & FedALA     \\ \hline
FMNIST & 95.5/ 96.3 & 90.1/ 97.7 & 91.6/ 98.0 & 92.1/ 99.2 & 89.0/ 93.3 & 90.3/ 99.6 & 91.0/ 95.2  & 90.3/ 95.6 & 88.5/ 86.7 & 85.9/ 99.6 \\ 
CIFAR10                                                 & 80.5/ 92.6 & 85.0/ 97.6 & 84.0/ 94.3 & 80.7/ 99.8 & 72.1/ 83.9 & 81.2/ 71.5 & 88.3/ 92.6  & 85.1/ 92.2 & 85.4/ 98.0 & 87.9/ 95.9 \\ 
N-BaIoT                                                 & 88.6/ 99.9 & 91.9/ 98.9 & 92.1/ 98.9 & 89.7/ 97.8 & 84.9/ 99.2 & 89.2/ 99.9 & 92.6/ 100.0 & 86.5/ 99.9 & 89.3/ 99.5 & 89.9/ 99.9  \\\hline
\end{tabular}
}
\end{table*}

%% file: table/table-fedrecover.tex
\begin{table*}[]
\centering
\caption{{The ACC/ ASR of backdoor attacks against FedRecover on Fashion-MNIST, CIFAR-10, and N-BaIoT.(\%)}}
\label{tab:fedrecover}
\resizebox{\linewidth}{!}{
\begin{tabular}{|c|c|c|c|c|c|c|c|c|c|c|}
 \hline
Dataset                                                 & FedAvg-FT  & FedProx-FT & SCAFFOLD   & FO         & HF         & pFedMe     & Ditto      & FedBN      & FedRep      & FedALA      \\\hline
FMNIST & 90.7/ 95.9 & 90.6/ 95.9 & 90.7/ 96.8 & 89.4/ 91.7 & 84.7/ 81.5  & 85.6/ 72.5 & 89.3/ 90.9 & 90.6/ 95.8 & 89.3/ 87.4  & 86.4/ 98.8  \\
CIFAR10                                                 & 80.5/ 94.7 & 83.1/ 91.4 & 81.2/ 95.8 & 74.4/ 77.3 & 71.8/ 90.6     & 75.4/ 91.5 & 79.5/ 96.5 & 79.8/ 84.2 & 80.3/ 96.8  & 74.6/ 97.4  \\
B-NaIoT                                                 & 88.0/ 99.9 & 88.1/ 99.9 & 88.1/ 99.9 & 83.7/ 95.6 & 81.2/ 72.0 & 85.6/ 95.8 & 88.7/ 99.9 & 88.1/ 99.9 & 84.8/ 100.0 & 87.61/ 99.9\\\hline
\end{tabular}
}
\end{table*}

%% file: table/l2-distance.tex
\begin{table}[t]
\centering
\caption{The Euclidean distance between the backdoor local model generated by different backdoor attacks and the backdoor-free global model obtained at the previous iteration on Fashion-MNIST.}
\label{tab:l2-distance}
\resizebox{\linewidth}{!}{
\begin{tabular}{c|cccccc}
               & Sybil & PGD  & ModelRe & Neurotoxin &CerP &  PFedBA      \\ \hline
FedAvg-FT      & 1.06  & 0.75 & 21.22   & 1.07      &0.93     & \textbf{0.51} \\
FedProx-FT     & 1.06  & 0.75 & 21.28   & 1.07       &1.17    & \textbf{0.51} \\
Per-Fedavg(FO) & 0.73  & 0.73 & 14.50   & 0.40     &0.63      & \textbf{0.37} \\
Per-Fedavg(HF) & 0.29  & 0.44 & 5.71    & 0.28     &0.28      & \textbf{0.28} \\
pFedMe         & 0.37  & 0.73 & 7.47    & 0.39      &0.31     & \textbf{0.26} \\
Ditto          & 1.06  & 0.75 & 21.22   & 1.07      &0.93     & \textbf{0.51} \\
SCAFFOLD       & 0.94  & 0.59 & 31.92   & 0.96     &1.15      & \textbf{0.51} \\
FedALA         & 0.95  & 0.60 & 18.78   & 0.97      &0.95     & \textbf{0.53} 
\end{tabular}
}
\end{table}

%% file: table/ablation-fmnist-onlyl2.tex
\begin{table*}[]
\centering
\caption{The ACC/ ASR of different variants of  PFedBA on Fashion-MNIST.(\%)}
\label{tab:ablation-fmnist}
\resizebox{\linewidth}{!}{
\begin{tabular}{|ccccccccccc|}
\hline
\multicolumn{1}{|c|}{Defense}    & \multicolumn{1}{c|}{FedAvg-FT}              & \multicolumn{1}{c|}{FedProx-FT}             & \multicolumn{1}{c|}{SCAFFOLD}                & \multicolumn{1}{c|}{FO}                     & \multicolumn{1}{c|}{HF}                      & \multicolumn{1}{c|}{pFedMe}                 & \multicolumn{1}{c|}{Ditto}                  & \multicolumn{1}{c|}{FedBN}                  & \multicolumn{1}{c|}{FedRep}                  & FedALA                  \\ \hline\hline
\multicolumn{11}{|c|}{No-Gradient}                                                                                                                                                                                                                                                                                                                                                                                                                                                          \\ \hline
\multicolumn{1}{|c|}{No defense} & \multicolumn{1}{c|}{92.1/ 97.3}             & \multicolumn{1}{c|}{92.1/ 97.3}             & \multicolumn{1}{c|}{92.4/ 94.9$\downarrow$}  & \multicolumn{1}{c|}{92.1/ 99.3}             & \multicolumn{1}{c|}{87.5/ 92.6}              & \multicolumn{1}{c|}{89.7/ 98.6}             & \multicolumn{1}{c|}{91.7/ 94.7$\downarrow$} & \multicolumn{1}{c|}{92.2/ 93.3$\downarrow$} & \multicolumn{1}{c|}{90.4/ 88.8 $\downarrow$} & 92.4/ 94.9 $\downarrow$ \\
\multicolumn{1}{|c|}{MKrum} & \multicolumn{1}{c|}{91.8/ 98.2}             & \multicolumn{1}{c|}{91.9/ 98.1}             & \multicolumn{1}{c|}{92.2/ 95.8 $\downarrow$} & \multicolumn{1}{c|}{92.0/ 99.2}             & \multicolumn{1}{c|}{87.5/ 90.1}              & \multicolumn{1}{c|}{89.9/ 99.4}             & \multicolumn{1}{c|}{91.7/ 97.2}             & \multicolumn{1}{c|}{91.9/ 94.0$\downarrow$} & \multicolumn{1}{c|}{89.3/ 22.6$\downarrow$}  & 92.2/ 95.8$\downarrow$  \\
\multicolumn{1}{|c|}{Trim}       & \multicolumn{1}{c|}{91.9/ 95.5}             & \multicolumn{1}{c|}{91.9/ 95.4}             & \multicolumn{1}{c|}{91.8/ 92.4 $\downarrow$} & \multicolumn{1}{c|}{91.6/ 95.8$\downarrow$} & \multicolumn{1}{c|}{87.3/ 84.8$\downarrow$}  & \multicolumn{1}{c|}{89.4/ 96.7}             & \multicolumn{1}{c|}{91.3/ 93.9$\downarrow$} & \multicolumn{1}{c|}{91.8/ 88.2$\downarrow$} & \multicolumn{1}{c|}{88.9/ 76.5$\downarrow$}  & 91.8/ 92.4$\downarrow$  \\
\multicolumn{1}{|c|}{DnC}        & \multicolumn{1}{c|}{92.1/ 97.8}             & \multicolumn{1}{c|}{92.1/ 97.7}             & \multicolumn{1}{c|}{92.0/ 91.2 $\downarrow$} & \multicolumn{1}{c|}{92.1/ 98.7}             & \multicolumn{1}{c|}{87.2/ 79.9 $\downarrow$} & \multicolumn{1}{c|}{89.8/ 98.0}             & \multicolumn{1}{c|}{91.7/ 97.1}             & \multicolumn{1}{c|}{92.0/ 95.5}             & \multicolumn{1}{c|}{89.4/ 26.1 $\downarrow$} & 92.0/ 91.2              \\
\multicolumn{1}{|c|}{FLAME}      & \multicolumn{1}{c|}{89.8/ 90.7$\downarrow$} & \multicolumn{1}{c|}{89.6/ 92.8$\downarrow$} & \multicolumn{1}{c|}{91.2/ 96.7}              & \multicolumn{1}{c|}{91.5/ 16.8$\downarrow$} & \multicolumn{1}{c|}{87.5/ 9.8$\downarrow$}   & \multicolumn{1}{c|}{89.5/ 10.6$\downarrow$} & \multicolumn{1}{c|}{91.2/ 86.5$\downarrow$} & \multicolumn{1}{c|}{89.8/ 90.7$\downarrow$} & \multicolumn{1}{c|}{87.8/ 10.5$\downarrow$}  & 91.2/ 96.7              \\ \hline\hline
\multicolumn{11}{|c|}{No-Loss}                                                                                                                                                                                                                                                                                                                                                                                                                                                              \\ \hline 
\multicolumn{1}{|c|}{No defense} & \multicolumn{1}{c|}{92.2/ 93.6$\downarrow$} & \multicolumn{1}{c|}{92.1/ 93.4$\downarrow$} & \multicolumn{1}{c|}{92.5/ 98.3}              & \multicolumn{1}{c|}{92.1/ 92.5$\downarrow$} & \multicolumn{1}{c|}{87.5/ 16.6$\downarrow$}  & \multicolumn{1}{c|}{89.9/ 23.9$\downarrow$} & \multicolumn{1}{c|}{91.8/ 88.4$\downarrow$} & \multicolumn{1}{c|}{92.1/ 97.3}             & \multicolumn{1}{c|}{90.1/ 95.7}              & 92.5/ 98.3              \\
\multicolumn{1}{|c|}{MKrum} & \multicolumn{1}{c|}{91.8/ 93.9$\downarrow$} & \multicolumn{1}{c|}{92.0/ 93.7$\downarrow$} & \multicolumn{1}{c|}{92.3/ 98.4}              & \multicolumn{1}{c|}{92.0/ 91.6$\downarrow$} & \multicolumn{1}{c|}{87.8/ 10.1$\downarrow$}  & \multicolumn{1}{c|}{90.0/ 10.7$\downarrow$} & \multicolumn{1}{c|}{91.9/ 89.9$\downarrow$} & \multicolumn{1}{c|}{91.8/ 98.2}             & \multicolumn{1}{c|}{89.0/ 91.3}              & 92.3/ 98.4              \\
\multicolumn{1}{|c|}{Trim}       & \multicolumn{1}{c|}{91.8/ 87.9$\downarrow$} & \multicolumn{1}{c|}{91.8/ 87.7$\downarrow$} & \multicolumn{1}{c|}{92.0/ 96.2}              & \multicolumn{1}{c|}{91.6/ 71.2$\downarrow$} & \multicolumn{1}{c|}{87.5/ 11.2$\downarrow$}  & \multicolumn{1}{c|}{89.3/ 12.7$\downarrow$} & \multicolumn{1}{c|}{91.4/ 83.8$\downarrow$} & \multicolumn{1}{c|}{91.9/ 95.5}             & \multicolumn{1}{c|}{87.0/ 92.3}              & 92.0/ 96.2              \\
\multicolumn{1}{|c|}{DnC}        & \multicolumn{1}{c|}{91.9/ 95.5}             & \multicolumn{1}{c|}{92.0/ 95.4$\downarrow$} & \multicolumn{1}{c|}{92.3/ 98.5}              & \multicolumn{1}{c|}{92.0/ 92.7$\downarrow$} & \multicolumn{1}{c|}{87.2/ 10.4$\downarrow$}  & \multicolumn{1}{c|}{90.1/ 14.0$\downarrow$} & \multicolumn{1}{c|}{91.7/ 92.8$\downarrow$} & \multicolumn{1}{c|}{92.1/ 97.8}             & \multicolumn{1}{c|}{89.5/ 91.1}              & 92.3/ 98.5              \\
\multicolumn{1}{|c|}{FLAME}      & \multicolumn{1}{c|}{89.5/ 96.0}             & \multicolumn{1}{c|}{90.5/ 96.7}             & \multicolumn{1}{c|}{91.3/ 99.0}              & \multicolumn{1}{c|}{91.4/ 99.5}             & \multicolumn{1}{c|}{87.0/ 77.5}              & \multicolumn{1}{c|}{89.5/ 98.9}             & \multicolumn{1}{c|}{91.4/ 96.6}             & \multicolumn{1}{c|}{89.5/ 96.0}             & \multicolumn{1}{c|}{88.3/ 87.4}              & 91.3/ 99.0              \\ \hline
\end{tabular}
}
\end{table*}

%% file: table/ablation-cifar-onlyl2.tex

\begin{table*}[]
\centering
\caption{The ACC/ ASR of different variants of  PFedBA  on CIFAR-10.(\%) }
\label{tab:ablation-cifar}
\resizebox{\linewidth}{!}{
\begin{tabular}{|ccccccccccc|}
\hline
\multicolumn{1}{|c|}{Defense}    & \multicolumn{1}{c|}{FedAvg-FT}              & \multicolumn{1}{c|}{FedProx-FT}             & \multicolumn{1}{c|}{SCAFFOLD}               & \multicolumn{1}{c|}{FO}                     & \multicolumn{1}{c|}{HF}                     & \multicolumn{1}{c|}{pFedMe}                 & \multicolumn{1}{c|}{Ditto}                  & \multicolumn{1}{c|}{FedBN}                  & \multicolumn{1}{c|}{FedRep}                 & FedALA                 \\ \hline\hline
\multicolumn{11}{|c|}{No-Gradient}                                                                                                                                                                                                                                                                                                                                                                                                                                                      \\ \hline
\multicolumn{1}{|c|}{No defense} & \multicolumn{1}{c|}{91.0/ 91.8$\downarrow$} & \multicolumn{1}{c|}{91.2/ 94.1$\downarrow$} & \multicolumn{1}{c|}{92.7/ 94.9$\downarrow$} & \multicolumn{1}{c|}{86.9/ 93.9$\downarrow$} & \multicolumn{1}{c|}{79.8/ 70.0$\downarrow$} & \multicolumn{1}{c|}{87.0/ 81.4$\downarrow$} & \multicolumn{1}{c|}{91.4/ 95.9$\downarrow$} & \multicolumn{1}{c|}{91.7/ 94.2$\downarrow$} & \multicolumn{1}{c|}{92.1/ 95.5$\downarrow$} & 85.9/ 96.9$\downarrow$ \\
\multicolumn{1}{|c|}{MKrum} & \multicolumn{1}{c|}{90.5/ 79.3$\downarrow$} & \multicolumn{1}{c|}{90.5/ 66.3$\downarrow$} & \multicolumn{1}{c|}{93.2/ 93.3$\downarrow$} & \multicolumn{1}{c|}{86.4/ 86.2$\downarrow$} & \multicolumn{1}{c|}{83.2/ 63.5$\downarrow$} & \multicolumn{1}{c|}{86.0/ 68.6$\downarrow$} & \multicolumn{1}{c|}{91.2/ 66.4$\downarrow$} & \multicolumn{1}{c|}{91.0/ 84.9$\downarrow$} & \multicolumn{1}{c|}{91.9/ 84.6$\downarrow$} & 85.5/ 60.4$\downarrow$ \\
\multicolumn{1}{|c|}{Trim}       & \multicolumn{1}{c|}{91.7/ 76.2$\downarrow$} & \multicolumn{1}{c|}{91.6/ 84.0$\downarrow$} & \multicolumn{1}{c|}{93.1/ 94.7$\downarrow$} & \multicolumn{1}{c|}{87.4/ 100.0} & \multicolumn{1}{c|}{82.1/ 36.6$\downarrow$} & \multicolumn{1}{c|}{86.6/ 74.9$\downarrow$} & \multicolumn{1}{c|}{91.6/ 77.5$\downarrow$} & \multicolumn{1}{c|}{91.8/ 86.8$\downarrow$} & \multicolumn{1}{c|}{92.1/ 78.3$\downarrow$} & 87.6/ 71.5$\downarrow$ \\
\multicolumn{1}{|c|}{DnC}        & \multicolumn{1}{c|}{90.8/ 82.4$\downarrow$} & \multicolumn{1}{c|}{90.7/ 85.8$\downarrow$} & \multicolumn{1}{c|}{93.1/ 94.2$\downarrow$} & \multicolumn{1}{c|}{86.5/ 85.9$\downarrow$} & \multicolumn{1}{c|}{71.6/ 5.9$\downarrow$}  & \multicolumn{1}{c|}{83.6/ 85.9$\downarrow$} & \multicolumn{1}{c|}{91.7/ 93.8$\downarrow$} & \multicolumn{1}{c|}{90.7/ 84.4$\downarrow$} & \multicolumn{1}{c|}{92.4/ 84.9$\downarrow$} & 82.8/ 97.3             \\
\multicolumn{1}{|c|}{FLAME}      & \multicolumn{1}{c|}{83.6/ 48.2$\downarrow$} & \multicolumn{1}{c|}{83.0/ 48.5$\downarrow$} & \multicolumn{1}{c|}{74.2/ 8.9$\downarrow$}  & \multicolumn{1}{c|}{82.4/ 58.7$\downarrow$} & \multicolumn{1}{c|}{80.1/ 39.9$\downarrow$} & \multicolumn{1}{c|}{79.7/ 39.1$\downarrow$} & \multicolumn{1}{c|}{87.0/ 100.0}            & \multicolumn{1}{c|}{82.5/ 55.8$\downarrow$} & \multicolumn{1}{c|}{86.3/ 80.7$\downarrow$} & 66.9/ 83.3$\downarrow$ \\ \hline\hline
\multicolumn{11}{|c|}{No-Loss}                                                                                                                                                                                                                                                                                                                                                                                                                                                          \\ \hline
\multicolumn{1}{|c|}{No defense} & \multicolumn{1}{c|}{91.0/ 91.8$\downarrow$} & \multicolumn{1}{c|}{90.4/ 89.8$\downarrow$} & \multicolumn{1}{c|}{93.0/ 89.9$\downarrow$} & \multicolumn{1}{c|}{86.7/ 99.8}             & \multicolumn{1}{c|}{73.1/ 15.3$\downarrow$} & \multicolumn{1}{c|}{86.9/ 77.6$\downarrow$} & \multicolumn{1}{c|}{91.6/ 94.9$\downarrow$} & \multicolumn{1}{c|}{91.2/ 91.5$\downarrow$} & \multicolumn{1}{c|}{92.2/ 94.8$\downarrow$} & 85.1/ 94.7$\downarrow$ \\
\multicolumn{1}{|c|}{MKrum} & \multicolumn{1}{c|}{91.2/ 71.5$\downarrow$} & \multicolumn{1}{c|}{90.1/ 69.9$\downarrow$} & \multicolumn{1}{c|}{93.2/ 91.3$\downarrow$} & \multicolumn{1}{c|}{86.6/ 99.7}             & \multicolumn{1}{c|}{80.0/ 66.1$\downarrow$} & \multicolumn{1}{c|}{84.4/ 61.4$\downarrow$} & \multicolumn{1}{c|}{91.5/ 77.4$\downarrow$} & \multicolumn{1}{c|}{90.4/ 80.2$\downarrow$} & \multicolumn{1}{c|}{91.9/ 84.6$\downarrow$} & 84.9/ 36.0$\downarrow$ \\
\multicolumn{1}{|c|}{Trim}       & \multicolumn{1}{c|}{91.9/ 71.8$\downarrow$} & \multicolumn{1}{c|}{91.5/ 77.9$\downarrow$} & \multicolumn{1}{c|}{92.9/ 92.8$\downarrow$} & \multicolumn{1}{c|}{83.1/ 99.0}             & \multicolumn{1}{c|}{81.9/ 22.3$\downarrow$} & \multicolumn{1}{c|}{87.1/ 70.9$\downarrow$} & \multicolumn{1}{c|}{91.5/ 74.4$\downarrow$} & \multicolumn{1}{c|}{91.6/ 77.7$\downarrow$} & \multicolumn{1}{c|}{92.3/ 74.1$\downarrow$} & 87.1/ 67.3$\downarrow$ \\
\multicolumn{1}{|c|}{DnC}        & \multicolumn{1}{c|}{91.0/ 80.1$\downarrow$} & \multicolumn{1}{c|}{90.8/ 84.8$\downarrow$} & \multicolumn{1}{c|}{93.1/ 91.1$\downarrow$} & \multicolumn{1}{c|}{86.5/ 24.8$\downarrow$} & \multicolumn{1}{c|}{71.6/ 5.9$\downarrow$}  & \multicolumn{1}{c|}{84.9/ 78.9$\downarrow$} & \multicolumn{1}{c|}{91.4/ 85.2$\downarrow$} & \multicolumn{1}{c|}{90.8/ 79.8$\downarrow$} & \multicolumn{1}{c|}{92.3/ 84.9$\downarrow$} & 82.4/ 93.5$\downarrow$ \\
\multicolumn{1}{|c|}{FLAME}      & \multicolumn{1}{c|}{82.5/ 93.0}             & \multicolumn{1}{c|}{83.2/ 94.2}             & \multicolumn{1}{c|}{90.5/ 99.8}             & \multicolumn{1}{c|}{82.5/ 99.0}             & \multicolumn{1}{c|}{80.4/ 82.5$\downarrow$} & \multicolumn{1}{c|}{78.5/ 93.3}             & \multicolumn{1}{c|}{87.0/ 100.0}            & \multicolumn{1}{c|}{82.1/ 94.3}             & \multicolumn{1}{c|}{86.3/ 80.7$\downarrow$} & 67.3/ 89.4$\downarrow$ \\ \hline
\end{tabular}
}
\end{table*}

%% file: arvix-main.bbl
\begin{thebibliography}{10}

\bibitem{ainsworth2023git}
Samuel~K. Ainsworth, Jonathan Hayase, and Siddhartha~S. Srinivasa.
\newblock Git re-basin: Merging models modulo permutation symmetries.
\newblock In {\em ICLR}, 2023.

\bibitem{howtobackdoor}
Eugene Bagdasaryan, Andreas Veit, Yiqing Hua, Deborah Estrin, and Vitaly Shmatikov.
\newblock How to backdoor federated learning.
\newblock In {\em AISTATS}, 2020.

\bibitem{alittle}
Gilad Baruch, Moran Baruch, and Yoav Goldberg.
\newblock A little is enough: Circumventing defenses for distributed learning.
\newblock In {\em NeurIPS}, 2019.

\bibitem{mult-krum}
Peva Blanchard, El~Mahdi~El Mhamdi, Rachid Guerraoui, and Julien Stainer.
\newblock Machine learning with adversaries: Byzantine tolerant gradient descent.
\newblock In {\em NIPS}, 2017.

\bibitem{HDBSCAN}
Ricardo J. G.~B. Campello, Davoud Moulavi, and J{\"{o}}rg Sander.
\newblock Density-based clustering based on hierarchical density estimates.
\newblock In {\em PAKDD}, 2013.

\bibitem{DBLP:conf/ndss/CaoF0G21}
Xiaoyu Cao, Minghong Fang, Jia Liu, and Neil~Zhenqiang Gong.
\newblock Fltrust: Byzantine-robust federated learning via trust bootstrapping.
\newblock In {\em NDSS}, 2021.

\bibitem{DBLP:conf/sp/CaoJZG23}
Xiaoyu Cao, Jinyuan Jia, Zaixi Zhang, and Neil~Zhenqiang Gong.
\newblock Fedrecover: Recovering from poisoning attacks in federated learning using historical information.
\newblock In {\em SP}, 2023.

\bibitem{DBLP:journals/tifs/CaoZJG22}
Xiaoyu Cao, Zaixi Zhang, Jinyuan Jia, and Neil~Zhenqiang Gong.
\newblock Flcert: Provably secure federated learning against poisoning attacks.
\newblock {\em {IEEE} Trans. Inf. Forensics Secur.}, 17:3691--3705, 2022.

\bibitem{backdoormeta}
Chien{-}Lun Chen, Sara Babakniya, Marco Paolieri, and Leana Golubchik.
\newblock Defending against poisoning backdoor attacks on federated meta-learning.
\newblock {\em {ACM} Trans. Intell. Syst. Technol.}, 13(5):76:1--76:25, 2022.

\bibitem{fedrep}
Liam Collins, Hamed Hassani, Aryan Mokhtari, and Sanjay Shakkottai.
\newblock Exploiting shared representations for personalized federated learning.
\newblock In {\em ICML}, 2021.

\bibitem{pfedme}
Canh~T. Dinh, Nguyen~Hoang Tran, and Tuan~Dung Nguyen.
\newblock Personalized federated learning with moreau envelopes.
\newblock In {\em NeurIPS}, 2020.

\bibitem{doan2020februus}
Bao~Gia Doan, Ehsan Abbasnejad, and Damith~C Ranasinghe.
\newblock Februus: Input purification defense against trojan attacks on deep neural network systems.
\newblock In {\em Annual Computer Security Applications Conference}, pages 897--912, 2020.

\bibitem{Domingos2020EveryML}
Pedro~M. Domingos.
\newblock Every model learned by gradient descent is approximately a kernel machine.
\newblock {\em ArXiv}, abs/2012.00152, 2020.

\bibitem{per-fedavg}
Alireza Fallah, Aryan Mokhtari, and Asuman~E. Ozdaglar.
\newblock Personalized federated learning with theoretical guarantees: {A} model-agnostic meta-learning approach.
\newblock In {\em NeurIPS}, 2020.

\bibitem{finn2017model}
Chelsea Finn, Pieter Abbeel, and Sergey Levine.
\newblock Model-agnostic meta-learning for fast adaptation of deep networks.
\newblock In {\em ICML}, 2017.

\bibitem{french1999catastrophic}
Robert~M French.
\newblock Catastrophic forgetting in connectionist networks.
\newblock {\em Trends in cognitive sciences}, 3(4):128--135, 1999.

\bibitem{foolsgold}
Clement Fung, Chris J.~M. Yoon, and Ivan Beschastnikh.
\newblock Mitigating sybils in federated learning poisoning.
\newblock {\em CoRR}, abs/1808.04866, 2018.

\bibitem{resnet}
Kaiming He, Xiangyu Zhang, Shaoqing Ren, and Jian Sun.
\newblock Deep residual learning for image recognition.
\newblock In {\em CVPR 2016}.

\bibitem{scaffold}
Sai~Praneeth Karimireddy, Satyen Kale, Mehryar Mohri, Sashank~J. Reddi, Sebastian~U. Stich, and Ananda~Theertha Suresh.
\newblock {SCAFFOLD:} stochastic controlled averaging for federated learning.
\newblock In {\em ICML}, 2020.

\bibitem{cifar10}
Alex Krizhevsky, Geoffrey Hinton, et~al.
\newblock Learning multiple layers of features from tiny images.
\newblock 2009.

\bibitem{DBLP:journals/corr/abs-2201-07063}
Phung Lai, NhatHai Phan, Abdallah Khreishah, Issa Khalil, and Xintao Wu.
\newblock Model transferring attacks to backdoor hypernetwork in personalized federated learning.
\newblock {\em CoRR}, abs/2201.07063, 2022.

\bibitem{lenet}
Yann LeCun, L{\'e}on Bottou, Yoshua Bengio, and Patrick Haffner.
\newblock Gradient-based learning applied to document recognition.
\newblock {\em Proceedings of the IEEE}, 86(11):2278--2324, 1998.

\bibitem{ditto}
Tian Li, Shengyuan Hu, Ahmad Beirami, and Virginia Smith.
\newblock Ditto: Fair and robust federated learning through personalization.
\newblock In {\em ICML}, 2021.

\bibitem{fedprox}
Tian Li, Anit~Kumar Sahu, Manzil Zaheer, Maziar Sanjabi, Ameet Talwalkar, and Virginia Smith.
\newblock Federated optimization in heterogeneous networks.
\newblock In {\em MLSys}, 2020.

\bibitem{fedbn}
Xiaoxiao Li, Meirui Jiang, Xiaofei Zhang, Michael Kamp, and Qi~Dou.
\newblock Fedbn: Federated learning on non-iid features via local batch normalization.
\newblock In {\em ICLR}, 2021.

\bibitem{li2021anti}
Yige Li, Xixiang Lyu, Nodens Koren, Lingjuan Lyu, Bo~Li, and Xingjun Ma.
\newblock Anti-backdoor learning: Training clean models on poisoned data.
\newblock In {\em NeurIPS}, 2021.

\bibitem{NAD}
Yige Li, Xixiang Lyu, Nodens Koren, Lingjuan Lyu, Bo~Li, and Xingjun Ma.
\newblock Neural attention distillation: Erasing backdoor triggers from deep neural networks.
\newblock In {\em ICLR}, 2021.

\bibitem{apple}
Jun Luo and Shandong Wu.
\newblock Adapt to adaptation: Learning personalization for cross-silo federated learning.
\newblock In {\em IJCAI}, 2022.

\bibitem{LyuHWLWL023}
Xiaoting Lyu, Yufei Han, Wei Wang, Jingkai Liu, Bin Wang, Jiqiang Liu, and Xiangliang Zhang.
\newblock Poisoning with cerberus: Stealthy and colluded backdoor attack against federated learning.
\newblock In {\em AAAI}, 2023.

\bibitem{ma2018teacher}
Yuzhe Ma, Robert Nowak, Philippe Rigollet, Xuezhou Zhang, and Xiaojin Zhu.
\newblock Teacher improves learning by selecting a training subset.
\newblock In {\em International Conference on Artificial Intelligence and Statistics}, pages 1366--1375. PMLR, 2018.

\bibitem{iot}
Yair Meidan, Michael Bohadana, Yael Mathov, Yisroel Mirsky, Asaf Shabtai, Dominik Breitenbacher, and Yuval Elovici.
\newblock N-baiot: Network-based detection of iot botnet attacks using deep autoencoders.
\newblock {\em IEEE Pervasive Computing}, 17(3):12--22, 2018.

\bibitem{flame}
Thien~Duc Nguyen, Phillip Rieger, Huili Chen, Hossein Yalame, Helen M{\"{o}}llering, Hossein Fereidooni, Samuel Marchal, Markus Miettinen, Azalia Mirhoseini, Shaza Zeitouni, Farinaz Koushanfar, Ahmad{-}Reza Sadeghi, and Thomas Schneider.
\newblock {FLAME:} taming backdoors in federated learning.
\newblock In {\em USENIX Security}, 2022.

\bibitem{kddpflbackdoor}
Zeyu Qin, Liuyi Yao, Daoyuan Chen, Yaliang Li, Bolin Ding, and Minhao Cheng.
\newblock Revisiting personalized federated learning: Robustness against backdoor attacks.
\newblock In {\em KDD}, 2023.

\bibitem{ndss2021}
Virat Shejwalkar and Amir Houmansadr.
\newblock Manipulating the byzantine: Optimizing model poisoning attacks and defenses for federated learning.
\newblock In {\em NDSS}, 2021.

\bibitem{towardsPFL}
Alysa~Ziying Tan, Han Yu, Lizhen Cui, and Qiang Yang.
\newblock Towards personalized federated learning.
\newblock {\em IEEE Transactions on Neural Networks and Learning Systems}, 2022.

\bibitem{wang2019neural}
Bolun Wang, Yuanshun Yao, Shawn Shan, Huiying Li, Bimal Viswanath, Haitao Zheng, and Ben~Y. Zhao.
\newblock Neural cleanse: Identifying and mitigating backdoor attacks in neural networks.
\newblock In {\em SP}, 2019.

\bibitem{tails}
Hongyi Wang, Kartik Sreenivasan, Shashank Rajput, Harit Vishwakarma, Saurabh Agarwal, Jy{-}yong Sohn, Kangwook Lee, and Dimitris~S. Papailiopoulos.
\newblock Attack of the tails: Yes, you really can backdoor federated learning.
\newblock In {\em NeurIPS}, 2020.

\bibitem{localfinetuning-1}
Kangkang Wang, Rajiv Mathews, Chlo{\'{e}} Kiddon, Hubert Eichner, Fran{\c{c}}oise Beaufays, and Daniel Ramage.
\newblock Federated evaluation of on-device personalization.
\newblock {\em CoRR}, abs/1910.10252, 2019.

\bibitem{wu2022backdoorbench}
Baoyuan Wu, Hongrui Chen, Mingda Zhang, Zihao Zhu, Shaokui Wei, Danni Yuan, and Chao Shen.
\newblock Backdoorbench: {A} comprehensive benchmark of backdoor learning.
\newblock In {\em NeurIPS}, 2022.

\bibitem{fmnist}
Han Xiao, Kashif Rasul, and Roland Vollgraf.
\newblock Fashion-mnist: a novel image dataset for benchmarking machine learning algorithms.
\newblock {\em CoRR}, abs/1708.07747, 2017.

\bibitem{DBA}
Chulin Xie, Keli Huang, Pin{-}Yu Chen, and Bo~Li.
\newblock {DBA:} distributed backdoor attacks against federated learning.
\newblock In {\em ICLR}, 2020.

\bibitem{localfinetuning-2}
Chen Yang, Yingchao Wang, Shulin Lan, Lihui Wang, Weiming Shen, and George~Q. Huang.
\newblock Cloud-edge-device collaboration mechanisms of deep learning models for smart robots in mass personalization.
\newblock {\em Robotics Comput. Integr. Manuf.}, 77:102351, 2022.

\bibitem{DBLP:journals/corr/abs-2307-15971}
Tiandi Ye, Cen Chen, Yinggui Wang, Xiang Li, and Ming Gao.
\newblock You can backdoor personalized federated learning.
\newblock {\em CoRR}, abs/2307.15971, 2023.

\bibitem{trimmed_mean}
Dong Yin, Yudong Chen, Kannan Ramchandran, and Peter~L. Bartlett.
\newblock Byzantine-robust distributed learning: Towards optimal statistical rates.
\newblock In {\em ICML}, 2018.

\bibitem{fedala}
Jianqing Zhang, Yang Hua, Hao Wang, Tao Song, Zhengui Xue, Ruhui Ma, and Haibing Guan.
\newblock Fedala: Adaptive local aggregation for personalized federated learning.
\newblock In {\em AAAI}, 2023.

\bibitem{durable}
Zhengming Zhang, Ashwinee Panda, Linyue Song, Yaoqing Yang, Michael~W. Mahoney, Prateek Mittal, Kannan Ramchandran, and Joseph Gonzalez.
\newblock Neurotoxin: Durable backdoors in federated learning.
\newblock In {\em ICML}, 2022.

\end{thebibliography}
